\documentclass{article} % For LaTeX2e
\usepackage{iclr2024_conference,times}

\usepackage{hyperref}
\usepackage{url}

\usepackage{xpatch}

\usepackage{calc,amsfonts,amssymb,amsmath,bm,color,theorem,graphicx,cite,epstopdf,nicefrac,stmaryrd}
\usepackage{psfrag,subfigure,float,epstopdf,bbold}
\usepackage[algoruled,linesnumbered]{algorithm2e}
\usepackage{multicol}
\usepackage{multirow,comment}
\usepackage{siunitx}
\usepackage{algorithmic}
\usepackage[normalem]{ulem}
\usepackage{mdframed}
\usepackage{footnote}
\makesavenoteenv{tabular}
\usepackage{booktabs}
\usepackage{xcolor}
\definecolor{orange}{RGB}{255,107,0}
\definecolor{green}{RGB}{50,170,50}
\definecolor{purple}{RGB}{255,0,255}

\def\green{\color{green}}
\usepackage{bibunits}

\usepackage{tikz}
\usetikzlibrary{bayesnet}
\usetikzlibrary{arrows}
\usepackage{color}
\usepackage{graphicx}
\usepackage{caption}
\usepackage{wrapfig}
\usetikzlibrary{backgrounds}

\newtheorem{Fact}{Fact}

\newtheorem{Assumption}{Assumption}

\newtheorem{Alemma}{Lemma}

\usepackage[flushleft]{threeparttable}

\providecommand{\qedsymbol}{$\\square$}
\newcommand{\mathqed}{\quad\hbox{\qedsymbol}}
\DeclareRobustCommand{\qed}{%
    \ifmmode \mathqed \else \leavevmode\unskip\penalty9999 \hbox{}\\nobreak\hfill \quad\hbox{\qedsymbol}%
\fi}

\newenvironment{proof}{\par\noindent{\bfseries Proof.}}{\qed\par}

\title{Towards Identifiable Unsupervised Domain Translation: A Diversified Distribution Matching Approach}
\author{Sagar Shrestha \& Xiao Fu \thanks{Source code is available at \href{https://github.com/XiaoFuLab/Identifiable-UDT.git}{https://github.com/XiaoFuLab/Identifiable-UDT.git}} \\
School of Electrical Engineering and Computer Science\\
Oregon State University\\
Corvallis, OR 97331, USA \\
\texttt{\{shressag,xiao.fu\}@oregonstate.edu} \\
}

\usepackage{steinmetz}
\renewcommand{\a}{\boldsymbol{a}}
\renewcommand{\b}{\boldsymbol{b}}

\newcommand{\f}{\boldsymbol{f}}

\newcommand{\g}{\boldsymbol{g}}
\newcommand{\h}{\boldsymbol{h}}
\newcommand{\m}{\boldsymbol{m}}

\renewcommand{\d}{\boldsymbol{d}}

\renewcommand{\u}{\boldsymbol{u}}
\newcommand{\w}{\boldsymbol{w}}
\newcommand{\y}{\boldsymbol{y}}
\newcommand{\x}{\boldsymbol{x}}
\newcommand{\z}{\boldsymbol{z}}

\newcommand{\bP}{\mathbb{P}}

\newcommand{\T}{{\!\top\!}}
\newcommand{\cA}{\mathcal{A}}
\newcommand{\cB}{\mathcal{B}}
\newcommand{\cC}{\mathcal{C}}
\newcommand{\cD}{\mathcal{D}}
\newcommand{\cE}{\mathcal{E}}

\newcommand{\cL}{\mathcal{L}}

\newcommand{\cN}{\mathcal{N}}

\newcommand{\cR}{\mathcal{R}}

\newcommand{\cT}{\mathcal{T}}

\newcommand{\cV}{\mathcal{V}}
\newcommand{\cW}{\mathcal{W}}
\newcommand{\cX}{\mathcal{X}}
\newcommand{\cY}{\mathcal{Y}}
\newcommand{\cZ}{\mathcal{Z}}

\newcommand{\bbR}{\mathbb{R}}
\newcommand{\bbE}{\mathbb{E}}

\usepackage{extarrows}
\newcommand{\deq}{\xlongequal{({\sf d})}}

\newtheorem{theorem}{Theorem}
\newtheorem{lemma}{Lemma}
\newtheorem{proposition}{Proposition}

\newtheorem{definition}{Definition}

\DeclareMathOperator*{\minimize}{\textrm{minimize}}

\usepackage{xcolor}
\usepackage{psfrag,framed}
\usepackage{lipsum}
\usepackage{chapterbib}
\PassOptionsToPackage{normalem}{ulem}
\usepackage{ulem}
\definecolor{shadecolor}{RGB}{220,220,220}
\iclrfinalcopy % Uncomment for camera-ready version, but NOT for submission.
\begin{document}
% \begin{bibunit}[iclr2024_conferene]   

\maketitle

\begin{abstract}

Unsupervised domain translation (UDT) aims to find functions that convert samples from one domain (e.g., sketches) to another domain (e.g., photos) without changing the high-level semantic meaning (also referred to as ``content''). The translation functions are often sought by probability distribution matching of the transformed source domain and target domain. CycleGAN stands as arguably the most representative approach among this line of work. However, it was noticed in the literature that CycleGAN and variants could fail to identify the desired translation functions and produce content-misaligned translations.
This limitation arises due to the presence of multiple translation functions---referred to as ``measure-preserving automorphism" (MPA)---in the solution space of the learning criteria. Despite awareness of such identifiability issues, solutions have remained elusive. This study delves into the core identifiability inquiry and introduces an MPA elimination theory. Our analysis shows that MPA is unlikely to exist, if multiple pairs of diverse cross-domain conditional distributions are matched by the learning function.
Our theory leads to a UDT learner using distribution matching over auxiliary variable-induced subsets of the domains---other than over the entire data domains as in the classical approaches.  The proposed framework is the first to rigorously establish translation identifiability under reasonable UDT settings, to our best knowledge.
Experiments corroborate with our theoretical claims.

\end{abstract}

\section{Introduction}
Domain translation (DT) aims to convert data samples from one feature domain to another, while keeping the key content information.
DT naturally arises in many applications, e.g., transfer learning \citep{zhuang2020comprehensive}, domain adaptation \citep{ganin2016domain, courty2017joint}, and cross-domain retrieval \citep{huang2015cross}.
Among them, a premier application is image-to-image (I2I) translation (e.g., profile photo to cartonized emoji and satellite images to street map plots \citep{isola2017image}).
{\it Supervised} domain translation (SDT) relies on paired data from the source and target domains. There, the translation functions are learned via matching the sample pairs. 

Nonetheless, paired data are not always available. In {\it unsupervised domain translation} (UDT), the arguably most widely adopted idea is to find neural transformation functions that perform probability distribution matching of the domains. 
The idea emerged in the literature in early works, e.g., \citep{liu2016coupled, taigman2016unsupervised, kim2017learning}. High-resolution image translation using distribution matching was later realized by the seminal work, namely, CycleGAN \citep{zhu2017unpaired}. 
CycleGAN learns a pair of transformations that are inverse of each other. One of transformations maps the source domain to match the distribution of the target domain, and the other transformation does the opposite. The distribution matching part is realized by the generative adversarial network (GAN) \citep{goodfellow2014generative}.
Using GAN-based distribution matching for UDT has attracted much attention---many follow-up works emerged; see the survey \citep{pang2021image}.

{\bf Challenge - Lack of Translation Identifiability.}
While UDT approaches have demonstrated significant empirical success, the theoretical question of translation identifiability has received relatively limited attention. Recent works \citep{galanti2017role, galanti2018generalization, moriakov2020kernel, galanti2021risk} pointed out failure cases of CycleGAN (e.g., content-misaligned translations like those in Fig. \ref{fig:failure_shoes}) largely attribute to the lack of translation identifiability.
That is, translation functions in the solution space of CycleGAN (or any distribution matching-based learners) is non-unique, due to the existence of {\it measure-preserving automorphism} (MPA) \citep{moriakov2020kernel} (the same concept was called {\it density-preserving mappings} in \citep{galanti2017role, galanti2018generalization}). MPA can ``swap'' the cross-domain sample correspondences without changing the data distribution---which is likely the main source of producing content misaligned samples after translation as seen in Fig.~\ref{fig:failure_shoes}. 
Many efforts were made to empirically enhance the performance of UDT, via implicitly or explicitly promoting solution uniqueness of their loss functions \citep{liu2017unit, courty2017joint, xu2022maximum, yang2023gp}. 
{A number of notable works approached the identifiability/uniqueness challenge by assuming that 
the desired translation functions have simple (e.g., linear \citep{gulrajani2022identifiability}) or specific structures \citep{de2021cyclegan}.
However, {translation identifiability} without using such restrictive structural assumptions have remained elusive.  }

{\bf Contributions.} 
In this work, we revisit distribution matching-based UDT.
Our contribution lies in both identifiability theory and implementation:

\begin{wrapfigure}{R}{0.4\textwidth}
     \vspace{-0.5cm}

        \centering
        \includegraphics[width=\linewidth]{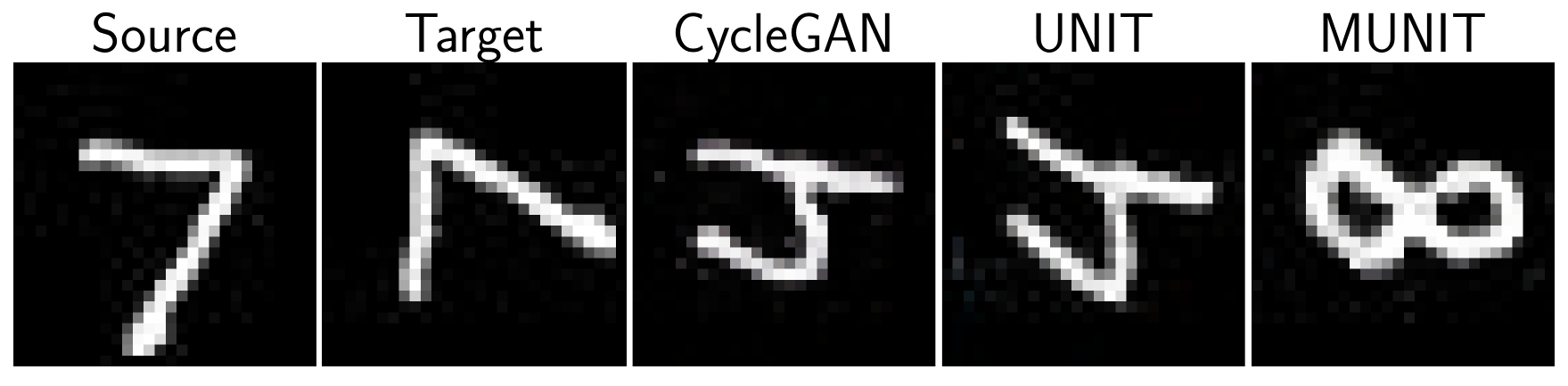}
        \caption{Lack of translation identifiability often leads to
        {\it content misalignment} in distribution matching based UDT methods, e.g.,
        \texttt{CycleGAN} \citep{zhu2017unpaired}, \texttt{MUNIT}\citep{huang2018multimodal}, and \texttt{UNIT} \citep{liu2017unit}. Source domain: MNIST Digits. Target Domain: Rotated Display of MNIST. }
        \label{fig:failure_shoes}
     \vspace{-0.35cm}
\end{wrapfigure}

$\bullet$ {\bf Theory Development: Establishing Translation Identifiability.} We delve into the core theoretical challenge regarding identifiability of the translation functions. As mentioned, the solution space of existing distribution matching criteria could be easily affected by MPA.
However, our analysis shows that the chance of having MPA decreases quickly when the translation function aligns more than one pair of diverse distributions.
This insight allows us to come up a sufficient condition, namely, the {\it sufficiently diverse condition} (SDC), to
establish translation identifiability of UDT.
To our best knowledge, our result stands as the first UDT identifiability theory without using simplified structural assumptions.

$\bullet$ {\bf Simple Implementation via Auxiliary Variables.} 
Our theoretical revelation naturally gives rise to a novel UDT learning criterion. This criterion aligns multiple pairs of conditional distributions across the source and target domains.
We define these conditional distributions over (overlapping) { sub-domaions} of the source/target domains using auxiliary variables. We demonstrate that in practical applications such as unpaired I2I translation, obtaining these { sub-domains} can be a straightforward task, e.g., through available side information or querying the foundation models like CLIP \citep{radford2021learning}.
Consequently, our { identification} theory can be readily put into practice.

{\bf Notation.} The full list of notations is in the supplementary material. 
Notably, we use $\bP_{\x}$  and $\bP_{\x|u}$ to denote the \textit{probability measures} of $\x$ and $\x$ conditioned on $u$, respectively. We denote the corresponding \textit{probability density function} (PDF) of $\x$ by $p(\x)$. For a measurable function $\f: \cX \to \cY$ and a distribution $\bP_{\x}$ defined over space $\cX$, the notation $\f_{\# \bP_{\x}}$ denotes the {\it push-forward measure}; that is, for any measurable set {$\cA \subseteq \cY$},
$ { \f_{\# \bP_{\x}} [\cA]} = \bP_{\x} [\f^{\rm preimg}(\cA)], \text{ where } \f^{\rm preimg}(\cA) = \{\x \in \cX ~|~ \f(\x) \in \cA\}. $ 
Simply speaking, $\f_{\# \bP_{\x}}$ denotes the distribution of $\f(\x)$ where $\x\sim \bP_{\x}$. The notation $\f_{\#\bP_{\x}} = \bP_{\y}$ means that the PDFs of $\f(\x)$ and $\y$ are identical {\it almost everywhere} (a.e.).

\section{Preliminaries}\label{sec:background}
Considers two data domains (e.g., photos and sketches). The samples from the two domains are represented by ${\bm x} \in {\cal X} \subseteq \mathbb{R}^{D_x}$ and ${\bm y} \in {\cal Y} \subseteq \mathbb{R}^{D_y}$. 
We make the following assumption:
\begin{Assumption}\label{assumption:correspondance}
For every $\x\in {\cal X}$, it has a corresponding $\y\in{\cal Y}$, and vice versa. In addition, there exist deterministic continuous functions $\bm f^{\star}: {\cal Y} \rightarrow {\cal X}$ and $\bm g^\star: {\cal X} \rightarrow {\cal Y}$ that link the corresponding pairs; i.e.,  
\begin{align}\label{eq:f_g_link}
    \bm f^\star(\y) =\x,\quad \g^\star(\x) = \y,~~\forall~\text{corresponding pair $(\x,\y)$}.
\end{align}
\end{Assumption}
In the context of domain translation, a linked $(\x,\y)$ pair can be regarded as cross-domain data samples that represent the same ``content'', and the translation functions $(\bm f^\star,\g^\star)$ are responsible for changing their ``appearances/styles''.
The term ``content'' refers to the semantic information to be kept across domains after translation. 
In Fig.~\ref{fig:failure_shoes}, the content is the { identity of the} digit ({ other than writing style or the rotation}); in 
Fig.~\ref{fig:content_correspondence.} of Sec.~\ref{sec:proposed},
the content { can be understood as the shared} characteristics of the person { in both the cartoon and the photo domains, which can collectively identify the person}.

Note that in the above setting, the goal is to find {\it two} ground-truth translation functions where one function's source is the other's target. Hence, both ${\cal X}$ and ${\cal Y}$ can serve as the source/target domains.
In addition, the above also implies $\f^\star =(\g^\star)^{-1}$, i.e., the ground-truth translation functions are invertible. Under this setting, if one can identify $\g^\star$ and $\f^\star$, then the samples in one domain can be translated to the other domain---while not changing the content. 
Note that Assumption \ref{assumption:correspondance} means that there is one-to-one correspondence between samples in the two domains, which can be a somewhat stringent condition in some cases. However, as we will explain in detail later, many UDT works, e.g., CycleGAN \citep{zhu2017unpaired} and variants \citep{liu2017unit, kim2017learning, choi2018stargan, park2020contrastive}, essentially used the model in Assumption \ref{assumption:correspondance} to attain quite interesting empirical results. This makes it a useful model and intrigues us to understand its underlying properties.

{\bf Supervised Domain Translation (SDT).}
In SDT, the corresponding pairs $(\x,\y)$ are assumed to be aligned {\it a priori}. 
Then, learning a translation function is essentially a regression problem---e.g., via finding $\bm g$ (or $\f$) such that $D(\bm g(\x) || \bm y)$ (or $D(\f(\y) || \x )$) is minimized over all given pairs, where $D(\cdot||\cdot)$ is { a certain ``distance'' measure}; see, e.g., \citep{isola2017image, wang2018high}.

{\bf Unsupervised Domain Translation (UDT).} In UDT, samples from the two domains are acquired separately without alignment.
Hence, sample-level matching as often done in SDT is not viable. Instead, UDT is often formulated as a probability distribution matching problem (see, e.g., \citep{zhu2017unpaired, taigman2016unsupervised, kim2019ugatit, park2020contrastive})---as distribution matching can be attained without using sample-level correspondences.
Assume that $\x$ and $\y$ are the random vectors that represent the data from the ${\cal X}$-domain and the ${\cal Y}$-domain, respectively.
Then, the desired $\f^\star$ and $\g^\star$ are sought via finding $\f$ and $\g$ such that 
\begin{align}\label{eq:prob_eq}
    \bP_{\y} = \g_{\# \bP_{\x}} \text{~~~and~~~} \bP_{\x} = \f_{\# \bP_{\y}}.
\end{align}
The hope is that distribution matching can work as a surrogate of sample-level matching as in SDT.
The arguably most representative work in UDT is CycleGAN \citep{zhu2017unpaired}. The CycleGAN loss function is as follows:
\begin{align}\label{eq:cyclegan}
    \min_{\f, \g} \max_{\d_{x}, \d_{y}} ~ \cL_{\rm GAN}(\g, \d_{y}, \x, \y) + \cL_{\rm GAN}(\f, \d_{x}, \x, \y) + \lambda \cL_{\rm cyc}(\g, \f),
\end{align}
where $\d_x$ and $\d_y$ represent two discriminators in domains $\cX$ and $\cY$, respectively,
\begin{align}\label{eq:theGAN}
    \cL_{\rm GAN}(\g, \d_{y}, \x, \y) = \bbE_{\y \sim \bP_{\y}} [\log \d_y(\y)] + \bbE_{\x \sim \bP_{\x}} [ \log ( 1 - \d_{y} (\g(\x)))],
\end{align}
$\cL_{\rm GAN}(\f, \d_{x}, \x, \y)$ is defined in the same way,
and the cycle-consistency term is defined as
\begin{align}\label{eq:cycleconsistenyc}
      \cL_{\rm cyc}(\g, \f) = \bbE_{\x \sim \bP_{\x}} \left[ \|\f(\g(\x)) - \x \|_1 \right] + \bbE_{\y \sim \bP_{\y}} \left[ \|\g(\f(\y)) - \y \|_1 \right].
\end{align}
The minimax optimization of the $\cL_{\rm GAN}$ terms enforces $\g_{\# \bP_{\x}} = \bP_{\y}$ and $\f_{ \# \bP_{\y}} = \bP_{\x}$. The $\cL_{\rm cyc}$ term encourages $\f=\g^{-1}$. 
CycleGAN showed the power of distribution matching in UDT and has triggered a lot of interests in I2I translation.
Many variants of CycleGAN were also proposed to improve the performance;
see the survey \citep{pang2021image}.

{\bf Lack of Translation Identifiability, MPA and Content Misalignment.}
Many works have noticed that distribution matching-type learning criterion may suffer from the lack of translation identifiability \citep{liu2017unit, moriakov2020kernel, galanti2017role, galanti2021risk, xu2022maximum}; i.e., the solution space of these criteria could have multiple solutions, and thus lack the ability to recover the ground-truth $\g^\star$ and $\f^\star$.
The lack of identifiability often leads to issues such as content misalignment as we saw in Fig.~\ref{fig:failure_shoes}.
To understand the identifiability challenge, let us formally define identifiability of any bi-directional UDT learning criterion:

\begin{definition}(Identifiability) 
Under the setting of Assumption~\ref{assumption:correspondance}, assume that $(\widehat{\f},\widehat{\g})$ is any {\it optimal solution} of a UDT learning criterion. Then, identifiability of $(\f^\star,\g^\star)$ holds under the UDT learning criterion if and only if $\widehat{\f}=\f^\star$ and $\widehat{\g}=\g^\star$ a.e. 
\end{definition}
Notice that we used the {\it optimal solution} in the definition.
This is because identifiability is a characterization of the ``kernel space'' (which contains all the zero-loss solutions) of a learning criterion \citep{moriakov2020kernel,fu2019nonnegative}.
In other words, when a UDT criterion admits translation identifiability, it indicates that the criterion provides a valid objective for the learning task---but identifiability is not related to the optimization procedure. 
We will also use the following:
\begin{definition}(MPA)
    A measure-preserving automorphism (MPA) of $\bP_{\x}$ is a continuous function $\h: \cX \to \cX$ such that
 $\bP_{\x} = \h_{ \# \bP_{\x}}$. 
\end{definition}
Simply speaking, MPA defined in this work is the continuous transformation $\bm h(\x)$ whose output has the same PDF as $p(\x)$.
Take the one-dimensional Gaussian distribution $x\sim {\cal N}(\mu,\sigma^2)$ as an example. The MPA of ${\cal N}(\mu,\sigma^2)$ is { $h(x)= - x + 2\mu$}.
A recent work \citep{moriakov2020kernel} suggested that non-identifiability of the desired translation functions by CycleGAN is caused by the existence of MPA. Their finding can be summarized in the following Fact:
\begin{Fact}
    If MPA of $\bP_{\x}$ or $\bP_{\y}$ exists, then CycleGAN and any criterion using distribution matching in \eqref{eq:prob_eq} do not have identifiability of $\f^\star$ and $\g^\star$. 
\end{Fact}
{\it Proof}: It is straightforward to see that  $\bP_{\y} = \g^\star_{\# \bP_{\x}}$ and $\bP_{\x} = \f^\star_{\# \bP_{\y}}$.
In addition, $\f^\star$ and $\g^\star$ are invertible.
Hence, the ground truth $(\f^\star, \g^\star)$ is an optimal solution of CycleGAN that makes the loss in \eqref{eq:cyclegan} equal to zero. However, due to the existence MPA, one can see that $\widehat{\f} = \bm h\circ \f^\star$ can also attain $\bP_{\x} = \widehat{\f}_{\# \bP_{\y}}$. This is because we have $\widehat{\f}_{\# \bP_{\y}}= \h\circ \f^\star_{\# \bP_{\y}} = \h_{\# \bP_{\x}}= \bP_{\x}$.
\begin{wrapfigure}{r}{2.55in}
\centering
    \includegraphics[width=\linewidth]{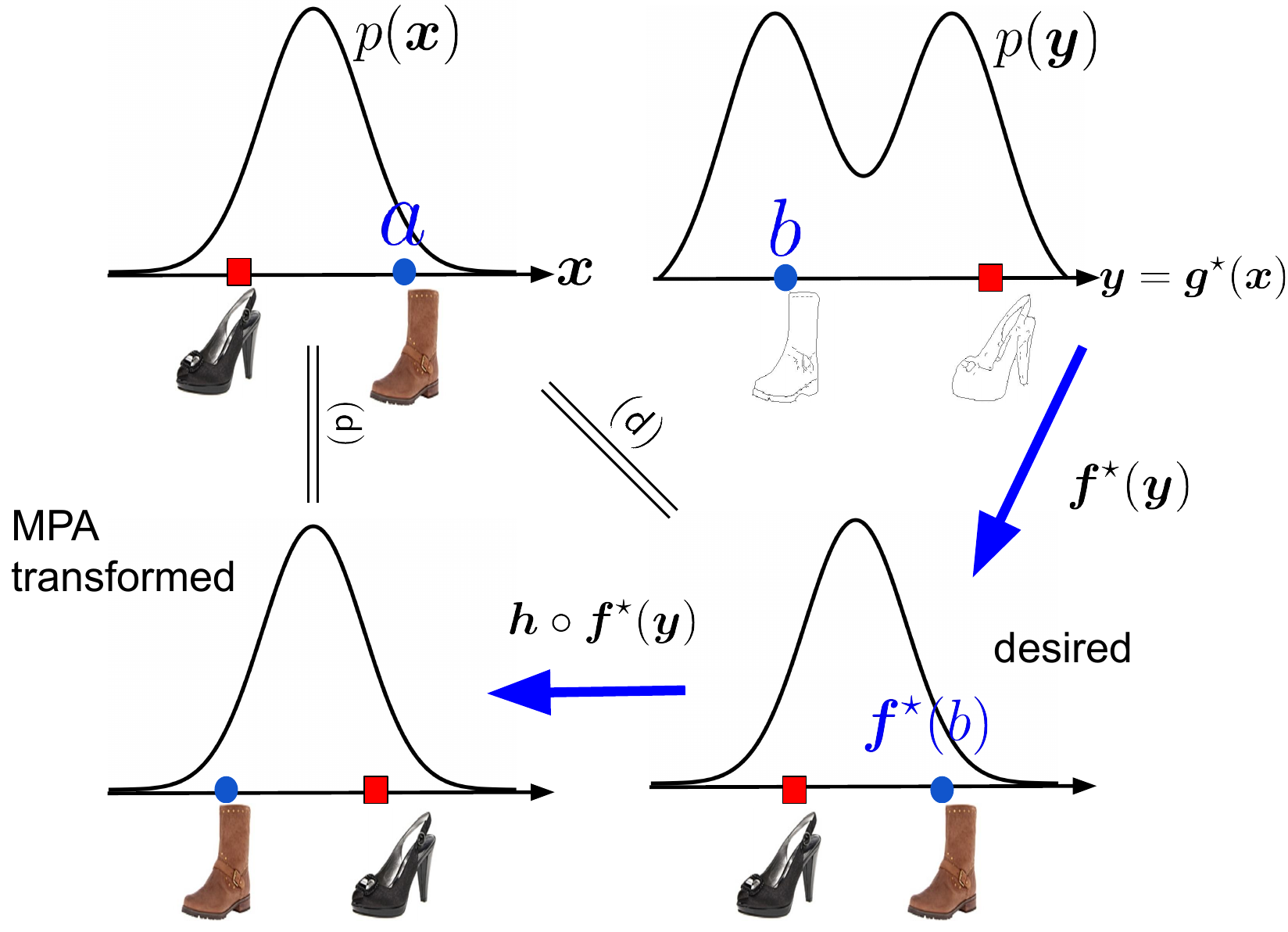}
    \caption{Illustration of of the lack of identifiability and MPA-induced content misalignment; ``$\deq$'' means distribution matching.}\label{fig:1d_gaussian_mpa}
    \includegraphics[width=\linewidth]{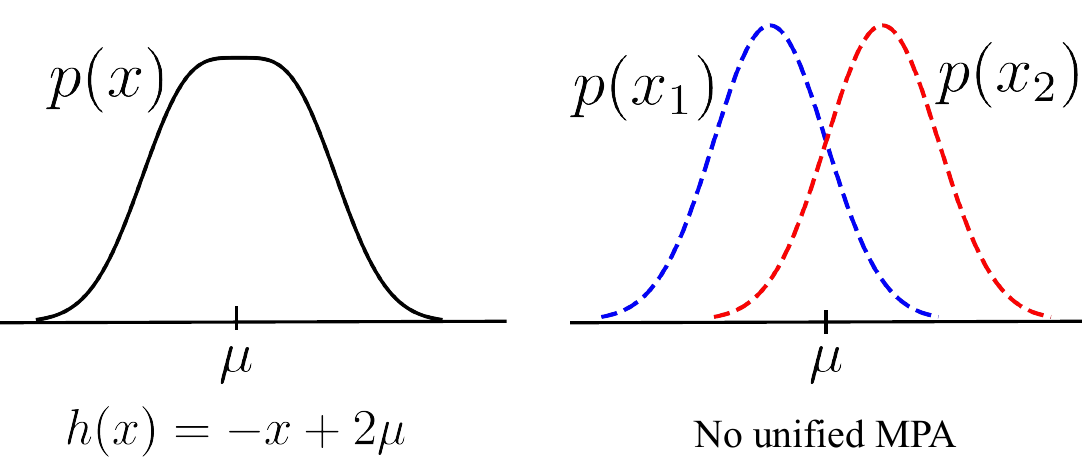}
    \caption{A unified MPA is harder to exist for a group of distributions.}
    \label{fig:multiple_distr_helps}
      \vspace{-1cm}
\end{wrapfigure}
Plus, as $\h\circ \f^\star$ is still invertible, $\widehat{\f}$ still makes the cycle-consistency loss zero.
Hence, the solution of CycleGAN is not unique and this loses identifiability of the ground truth translation functions.
\hfill $\square$

The existence of MPA in the solution space of the UDT learning losses may be detrimental in terms of avoiding content misalignment. 
To see this, consider the example
in Fig. \ref{fig:1d_gaussian_mpa}.
There, $\bP_{x}=\cN(\mu, \sigma^2)$ and { $\h(x) = - x + 2\mu$} is an MPA of $\bP_{x}$, as mentioned. 
Note that $\widehat{\f} =\h \circ \f^\star$ can be an optimal solution found by CycleGAN.
However, such an $\widehat{\f}$ can cause misalignment.
To explain,
assume $x=a$ and $y=b$ are associated with the same entity, which means that $a = \f^\star(b)$ represents the ground-truth alignment and translation.
However, as
${ p(-a+2\mu)=p(\h(a))}  = p( \h\circ \f^\star(b)) =p( \widehat{\f}(b))$,
the learned function $\widehat{\f}$ wrongly translates $y=b$ to { $x=-a+2\mu$}. 

Our Gaussian example seems to be special as it has symmetry about its mean. 
However, the existence of MPA is not unusual. To see this, we show the following result:
\begin{proposition}\label{prop:mpa_existence}
    Suppose that $\bP_{\x}$ admits a continuous PDF, $p(\x)$ and $p(\x)>0, \forall \x \in \cX$. Assume that $\cX$ is simply connected. Then, there exists a continuous non-trivial (non-identity) $\h(\cdot)$ such that $\h_{\# \bP_{\x}} = \bP_{\x}$. 
\end{proposition}
Note that there are similar results in \citep{moriakov2020kernel} regarding the existence of MPA, but more assumptions were  made in their proof.
The universal existence of MPA attests to the challenging nature of establishing translation identfiability in UDT. 

\section{Identifiable UDT via Diversified Distribution Matching}\label{sec:proposed}

{\bf Intuition - Exploiting Diversity of Distributions.} Our idea starts with the following observation: If two distributions have different PDFs, a shared MPA is unlikely to exist.
Fig. \ref{fig:multiple_distr_helps} illustrates the intuition.
Consider two Gaussian distributions $x_1\sim {\cal N}(\mu_1,1)$ and $x_2 \sim {\cal N}(\mu_2,1)$ with $\mu_1\neq \mu_2$. For each of them, { $h(x)=-x+2\mu_i$} for $i=1,2$ is {an} MPA.
However, there is not a function that
can serve as a unified MPA to attain 
$h_{\# \bP_{x_1}} = \bP_{x_1} ~\&~ h_{ \# \bP_{x_2}} = \bP_{x_2} $ simultaneously.
Intuitively, the diversity of the PDFs of $x_1$ and $x_2$ has made finding a unified MPA $h(\cdot)$ difficult. 
This suggests that instead of matching the distributions of $\x$ and $\f(\y)$ and those of $\y$ and $\g(\x)$, it may be beneficial to match the distributions of more variable pairs whose probability measures are diverse.

\begin{wrapfigure}{r}{0.4\textwidth}
   \vspace{-.25cm}
    \centering
    \includegraphics[width=\linewidth]{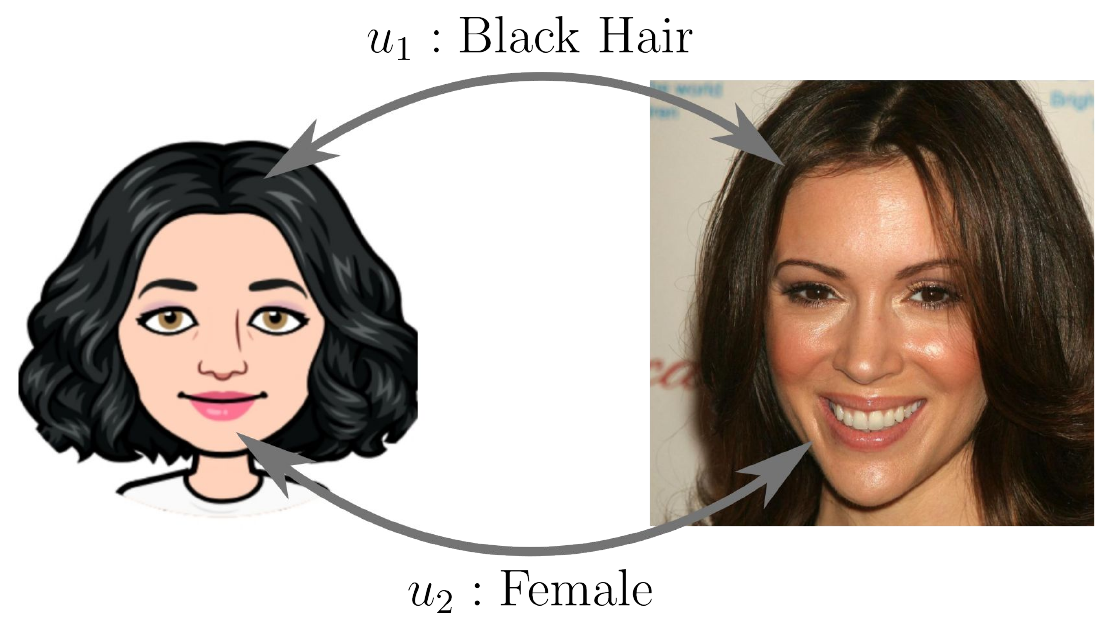}
    \caption{Examples of $u_i$. }
    \label{fig:content_correspondence.}
 \vspace{-.3cm}
\end{wrapfigure}

{\bf Auxiliary Variable-Assisted Distribution Diversification.}
{In applications, the corresponding samples $\x,\y$ often share some aspects/traits. 
For example, in Fig.~\ref{fig:content_correspondence.}, the corresponding $\x$ and $\y$ both have dark hair or the same gender.
If we model a collection of such traits as different realizations of discrete random variable $u$, the alphabet of $u$, denoted as $\{u_1,\ldots,u_I\}$ represents these traits.
We should emphasize that the traits is a result of the desired content invariance across domains, but need not to represent the whole content.

To proceed, we observe that the conditional distributions $\bP_{\x|u=u_i}$ and $\bP_{\y|u=u_i}$ satisfy
$ \bP_{\x|{u=u_i} } = ~\f^\star_{\#\bP_{\y|u=u_i}},~ \bP_{\y|u=u_i} = \g^\star_{ \# \bP_{\x|u=u_i}},~\forall i. $
The above holds since $\x$ and $\y$ have a deterministic relation and because the trait $u_i$ is shared by the content-aligned pairs $(\x,\y)$.

In practice, $u$ can take various forms. In I2I translation, one may use image categories or labels, if available, to serve as $u$. Note that knowing the image categories does {\it not} mean the samples from the two domains are aligned, as each category could contain a large amount of samples.
In addition, one can use sample attributes (such as hair color, gender as in Fig.~\ref{fig:content_correspondence.}) to serve as $u$, if these attributes are not meant to be changed in the considered translation tasks. 
If not immediately available,
these attributes can be annotated by open-sourced AI models, e.g., CLIP \citep{radford2021learning}; see detailed implementation in the supplementary material.
{A similar idea of using CLIP to acquire auxiliary information was explored in \citep{gabbay2021image}}.}

By Proposition~\ref{prop:mpa_existence}, it is almost certain that $\bP_{\x|u=u_i}$ has an MPA $\h_i$ for all $i\in[I]$. 
However, it is likely that $\h_i\neq \h_j$ if $\bP_{\x|u=u_i}$ and $\bP_{\x|u=u_j} $ are sufficiently different.
As a consequence, similar to what we saw in Fig.~\ref{fig:multiple_distr_helps},
if one looks for $\f$ that does simultaneous matching of
\begin{align}\label{eq:cond_match}
          \bP_{\x|u=u_i} =  \f_{\# \bP_{\y|u=u_i}},~\forall i\in [I], 
\end{align}
it is more possible that $\f=\f^\star$ instead of having other solutions---this leads to identfiiability of $\f^\star$.

\textbf{Proposed Loss Function. } 
We propose to match multiple distribution pairs $(\bP_{\x|u_i}, \f_{\# \bP_{\y|u_i}})$ (as well as $(\bP_{\y|u_i},\g_{\# \bP_{\x|u_i}})$) for $i=1,\ldots,I$.  
For each pair, we use discriminator $\d_x^{(i)}: \cX \to [0,1]$ (and $\d_y^{(i)}: \cY \to [0,1]$ in reverse direction). Then, our loss function is as follows:
\begin{align}\label{eq:split_cyclegan}
    \min_{\f, \g} \max_{\{\d_{x}^{(i)}, \d_{y}^{(i)}\}} ~ \sum_{i=1}^I \left(\cL_{\rm GAN}(\g, \d_{y}^{(i)}, \x, \y) + \cL_{\rm GAN}(\f, \d_{x}^{(i)}, \x, \y)\right) + \lambda \cL_{\rm cyc}(\g, \f),
\end{align}
where we have 
\begin{align}\label{eq:dsganloss}
 &\cL_{\rm GAN}\left(\g, \d_{y}^{(i)}, \x, \y \right) =  {{\sf Pr}(u = u_i)} \left(\bbE_{\y \sim \bP_{\y|u=u_i}} \left[\log \d_y^{(i)}(\y)\right] + \bbE_{\x \sim \bP_{\x|u=u_i}} \left[ \log\left( 1 - \d_y^{(i)} (\g(\x)) \right)\right] \right). \nonumber
\end{align}
Note that $\x\sim \bP_{\x|u_i}$ represents samples that share the same characteristic defined by $u_i$ (e.g., hair color, eye color, gender). This means that the loss function matches a suite of distributions defined over (potentially overlapping) subdomains over the entire domain ${\cal X}$ and ${\cal Y}$.
We should emphasize that the auxiliary variable is only needed in the training stage, but not the testing stage.

We call the proposed method {\it \uline{d}iversified d\uline{i}stribution \uline{m}atching for unsup\uline{e}rvised domai\uline{n} tran\uline{s}lat\uline{ion}} (\texttt{DIMENSION}) \footnote{{Note that we still use the term ``unsupervised'' despite the need of auxiliary information---as no paired samples are required. We avoided using ``semi-supervised'' or ``weakly supervised'' as these are often reserved for methods using some paired samples; see, e.g.,  \citep{wang2020semi, mustafa2020transformation}.  } }.
The following lemma shows that \texttt{DIMENSION} exactly realizes our idea in \eqref{eq:cond_match}:
\begin{lemma}\label{lemma:fact_sln}
Assume that an optimal solution of \eqref{eq:split_cyclegan} is $(\widehat{\bm f},\widehat{\bm g},\{ \widehat{\bm d}^{(i)}_x,\widehat{\bm d}^{(i)}_y\})$. Then, under Assumption~\ref{assumption:correspondance}, we have
$\bP_{\x |u=u_i} = \widehat{\f}_{\# \bP_{\y|u=u_i}}$,  $\bP_{\y |u=u_i}=\widehat{\g}_{\# \bP_{\x|u=u_i}}, ~\forall i \in [I]$, and $\widehat{\f} = \widehat{\g}^{-1}$, a.e. 
\end{lemma}

{\bf Identfiiability Characterization.}
Lemma~\ref{lemma:fact_sln} means that solving the \texttt{DIMENSION} loss leads to conditional distribution matching as we hoped {for in} \eqref{eq:cond_match}.
Hower, it does not guarantee that $(\widehat{\f},\widehat{\g})$ found by \texttt{DIMENSION} satisfies $\widehat{\f}=\f^\star$ and $\widehat{\g}=\g^\star$. Towards establishing {\it identifiability} of the ground-truth translation functions via \texttt{DIMENSION},
we will use the following definition:
\begin{definition}[Admissible MPA]
    Given auxiliary variable $u$, the function $\h(\cdot)$ is said to be an admissible MPA of $ \{\bP_{\x|u=u_i}\}_{i=1}^I$ if and only if 
    $\bP_{\x|u=u_i} = \h_{\# \bP_{\x|u=u_i}}, \forall i \in [I].$
\end{definition}
Now, due to the deterministic relationship between the pair $\x$ and $\y$, we have the following fact:
\begin{Fact}\label{fact:x_y_symmetry}
Suppose that Assumption~\ref{assumption:correspondance} holds.
Then, there exists an admissible MPA of $\{\bP_{\x|u=u_i}\}_{i=1}^I$ if and only if there exists an admissible MPA of $\{\bP_{\y|u=u_i}\}_{i=1}^I$.
\end{Fact}
The above means that if we establish {that} there is no admissible MPA of the $\{\bP_{\x|u=u_i}\}_{i=1}^I$, it suffices to conclude that
there is no admissible MPA of $\{ \bP_{\y|u=u_i}\}_{i=1}^I \}$.

\begin{wrapfigure}{r}{2in}
    \vspace{-0.25cm}
    \centering
    \includegraphics[width=1\linewidth]{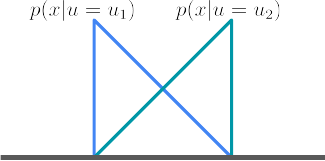}
    \caption{Conditional PDFs $p(x|u = u_1)$ and $p(x|u = u_2)$ that satisfy the SDC.  }
    \label{fig:distinct_distr}
    \centering
    \includegraphics[width=1\linewidth]{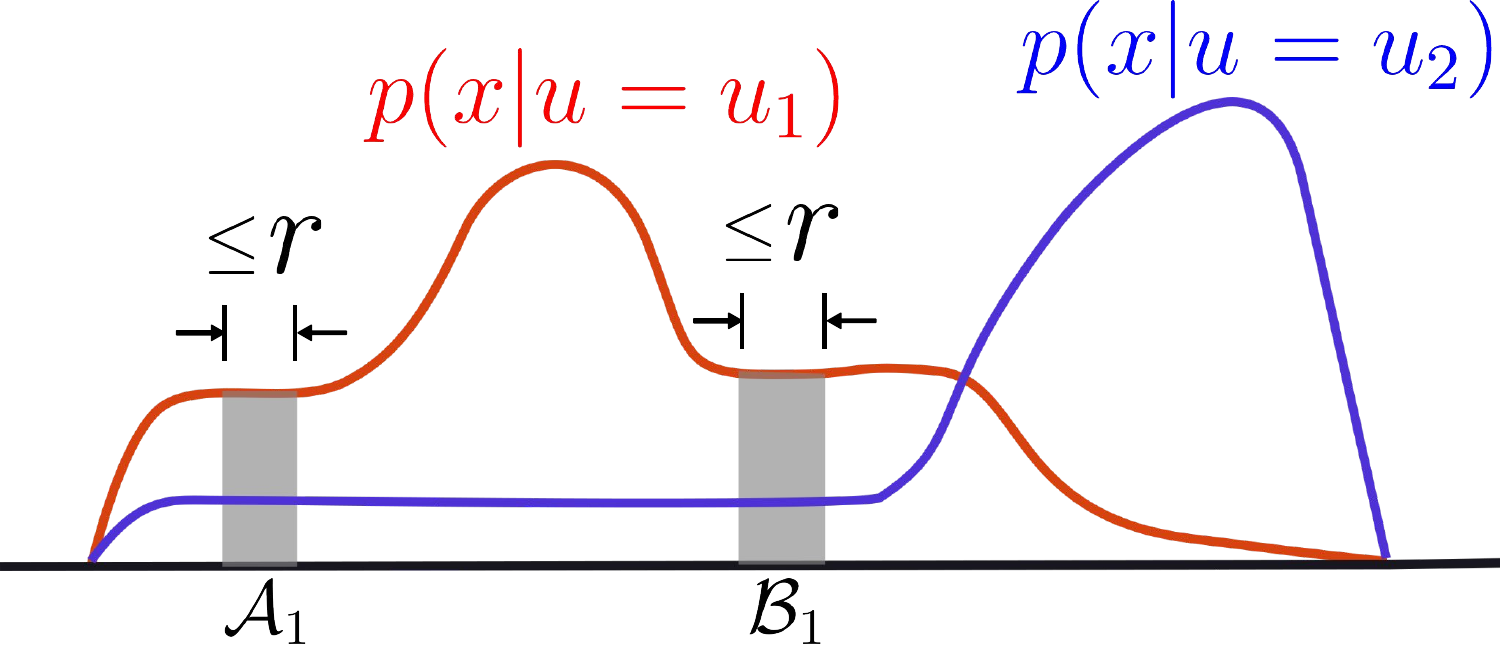}
    \caption{Illustration of relaxed SDC ($r$-SDC).}
    \label{fig:robust_sdc_illus}
    \vspace{-.85cm}
\end{wrapfigure}

As described before, to ensure identifiability of the translation functions via solving the \texttt{DIMENSION} loss, we hope {the} conditional distributions $\bP_{\x|u=u_i}$ and $\bP_{\y|u=u_i}$ to be sufficiently different. We formalize this requirement in the following definition:
\begin{definition}[Sufficiently Diverse Condition (SDC)]\label{assump:distinct_distr}
For any two disjoint sets $\cA, \cB \subset \cX $, where ${\cal A}$ and ${\cal B}$ are connected, open, and non-empty, there exists a ${u}_{(\cA,\cB)} \in  \{u_1, \dots, u_I\}$ such that 
$\bP_{\x|u={u}_{(\cA,\cB)}} [\cA] \neq \bP_{\x|u={u}_{(\cA,\cB)}} [\cB].$
Then, the set of conditional distributions $\{\bP_{\x|u=u_i}\}_{i=1}^I$ is called {\it sufficiently diverse}.
\end{definition}
Definition \ref{assump:distinct_distr} puts the desired ``diversity'' into context. 
It is important to note that the SDC only requires the {\it existence} of a certain ${u}_{(\cA,\cB)}\in\{u_1,\ldots,u_I\}$ for a given disjoint set pair $({\cal A},{\cal B})$. It does not require a unified ${u}$ for all pairs; i.e., $u_{(\cA,\cB)}$ needs not to be the same as $u_{({\cA',\cB'})}$ for $(\cA,\cB)\neq (\cA',\cB')$.
Fig. \ref{fig:distinct_distr} shows a simple example where the two conditional distributions satisfy the SDC.
In more general cases, this implies that if the PDFs of the conditional {distributions} exhibit different ``shapes'' over their supports, SDC is likely to hold.
Using SDC, we show the following translation identifiability result:
\begin{theorem}[Identifiability]\label{thm:identifiability}
Suppose that Assumption~\ref{assumption:correspondance} holds. 
Let ${\sf E}_{i,j}$ denote the event that the pair $(\bP_{\x|u=u_i},\bP_{\x|u=u_j})$ does not satisfy the SDC. Assume that ${\rm Pr}[{\sf E}_{i,j}] \leq \rho$ for any $i \neq j$, where $i,j \in [I]$.
Let $(\widehat{\f},\widehat{\g})$ be from an optimal solution of the \texttt{DIMENSION} loss \eqref{eq:split_cyclegan}. Then, there is no admissible MPA of $\{\bP_{\x|u=u_i}\}_{i=1}^I$ of the solution, i.e., 
{$
\widehat{\f} = \f^\star,~a.e.$ and $\widehat{\g} = \g^\star,~a.e.
$}
with a probability of at least $1-\rho^{I\choose 2}$.    
\end{theorem}

Theorem \ref{thm:identifiability} shows that if the conditional distributions are sufficiently diverse, solving \eqref{eq:split_cyclegan} can correctly identify the ground-truth translation functions. Theorem~\ref{thm:identifiability} also spells out the importance of having more $u_i$'s (which means more auxiliary information). The increase of $I$ improves the probability of success quickly.

{\bf Towards More Robust Identifiability.} Theorem \ref{thm:identifiability} uses the fact that the SDC holds with high probability for every pair of $(\bP_{\x|u_i},\bP_{\x|u_j})$ (cf. ${\sf Pr}[{\sf E}_{i,j}]\leq \rho$). It is also of interest to see if the method is robust to violation of the SDC. To this end, consider the following condition:
\begin{definition}[Relaxed Condition: $r$-SDC]\label{assump:violation}
    Let ${\rm dia}(\cA) = {\rm sup}_{\w,\z \in \cA} \|\w - \z\|_2$ and
  $  
    \cV_{i,j} = \big\{(\cA, \cB) ~|~\bP_{\x|{u}_i} [\cA] = \bP_{\x|{u}_{i}} [\cB]~\&~\bP_{\x|{u}_j} [\cA] = \bP_{\x|{u}_{j}} [\cB], \cA \cap \cB = \phi \big\},
$  
    where $
    \cA, \cB$ are non-empty, open and connected.
    Denote
    $M_{i,j} = \max_{(\cA, \cB) \in \cV_{i,j}} {\rm max}\{{\rm dia}(\cA), {\rm dia}(\cB)\}$. Then, $(\bP_{\x|{u}_i},\bP_{\x|{u}_j})$ satisfies the $r$-SDC if $M_{i,j} \leq r$ for $r \geq 0$.
\end{definition}
Note that the $r$-SDC becomes the SDC when $r=0$.
Unlike SDC in Definition~\ref{assump:distinct_distr}, the relaxed SDC condition allows the violation of SDC over regions ${\cal V}_{i,j}$. Our next theorem shows that the translation identifiability still approximately holds, as long as the largest region in ${\cal V}_{i,j}$ is not substantial:

\begin{theorem}[Robust Identifiability]\label{thm:lipschitz}
    Suppose that Assumption \ref{assumption:correspondance} holds with $\bm g^\star$ being $L$-Lipschitz continuous, and that any pair of $(\bP_{\x|u_i},\bP_{\x|u_j})$  satisfies the $r$-SDC (cf. Definition~\ref{assump:violation}) with probability at least $1-\gamma$, i.e., ${\rm Pr}[M_{i,j} \geq r] \leq \gamma$ for any $i \neq j$, where $(i, j) \in [I]\times [J]$.
    Let $\widehat{\g}$ be from any optimal solution of the \texttt{DIMENSION} loss in \eqref{eq:split_cyclegan}. Then, we have
    $\|\widehat{\g}(\x) - \g^\star(\x) \|_2 \leq 2rL, \quad \forall \x \in \cX,$
    with a probability of at least $1 - \gamma^{I \choose 2}$.
    The same holds for $\widehat{\f}$.
\end{theorem}
 Theorem \ref{thm:lipschitz} asserts that the estimation error of $\widehat{\g}$ scales linearly with the ``degree'' of violation of the SDC (measured by $r$). 
The result is encouraging: It shows that even if the SDC is  violated, the performance of \texttt{DIMENSION} will not decline drastically.
{The Lipschitz continuity assumption in Theorem \ref{thm:lipschitz} is mild. Note that translation functions are often represented by neural networks in practice, and neural networks with bounded weights are Lipschitz continuous functions \citep{bartlett2017spectrally}. Hence, the numerical successes of many neural UDT models (e.g., CycleGAN) suggest that assuming that Lipschitz continuous ground-truth translation functions exist is reasonable.}

\section{Related Works}
Prior to CycleGAN \citep{zhu2017unpaired}, the early works \citep{liu2016coupled, taigman2016unsupervised, kim2017learning} started using GAN-based neural structures for distribution matching in the context of I2I translation.
Similar ideas appeared in UDT problems in NLP (e.g., machine translation) \citep{conneau2017word, lample2017unsupervised}.
In the literature, it was noticed that distribution matching modules lack solution uniqueness, and many works proposed remedies (see, e.g, \citep{liu2017unit, xu2022maximum, xie2022unsupervised, park2020contrastive}).
These approaches have worked to various extents empirically, but the translation identifiability question was unanswered.
The term ``content'' was used in the vision literature (in the context of I2I translation) to refer to domain-invariant attributes (e.g., pose and orientation \citep{kim2019ugatit, amodio2019travelgan, wu2019transgaga, yang2023gp}). 
This is a narrower interpretation of content relative to ours---as content in our case can be high-level or latent semantic meaning that is not represented by specific attributes. Our definition of content is closer to that in multimodal and self-supervised learning \citep{von2021self,lyu2021understanding,daunhawer2023identifiability}.
Before our work, auxiliary information was also considered in UDT. For example, semi-supervised UDT (see, e.g., \citep{wang2020semi, mustafa2020transformation}) uses a small set of paired data samples, but our method does {not} use any sample-level pairing information. Attribute-guided I2I translation (see, e.g., \citep{li2019attribute, choi2018stargan, choi2020starganv2}) specifies the desired attributes in the target domain to ``guide'' the translation. These are different from our auxiliary variables that can be both sample attributes or high-level concepts (which is closer to the ``auxiliary variables'' in nonlinear independent component analysis works, e.g., \citep{hyvarinen2019nonlinear}). Again, translation identifiability was not considered for semi-supervised or attribute-guided UDT.
There has been efforts towards understanding the translation identifiability of CycleGAN.
The works of \citet{galanti2017role, galanti2018generalization} recognized that the success of UDT may attribute to the existence of a small number of MPAs. 
\citet{moriakov2020kernel} showed that MPA exists in the solution space of CycleGAN, and used it to explain the ill-posedness of CycleGAN. \citet{chakrabarty2022translation} studied the finite sample complexity of CycleGAN in terms of distribution matching and cycle consistency. 
\citet{gulrajani2022identifiability} and \citet{de2021cyclegan} argued that if the target translation functions have known structures (e.g., linear or optimal transport structures), then translation identifiability can be established.
However, these conditions can be restrictive.
Translation identifiability without using such structural assumptions had remained unclear before our work.

\section{Numerical Validation}

{\bf Constructing Challenging Translation Tasks.}
We construct challenging translation tasks to validate our theorems and to illustrate the importance of translation identifiability. To this end, we make three datasets.
The first two are ``MNIST v.s. Rotated MNIST'' (MrM) and ``Edges v.s. Rotated Shoes'' (ErS). 
In both datasets, the rotated domains consist of samples from the ``MNIST'' and ``Shoes'' with a 90 degree rotation, respectively. We intentionally make this rotation, as rotation is a large geometric change across domains. This type of large geometric change poses a challenging translation task \citep{kim2019ugatit, wu2019transgaga, amodio2019travelgan, yang2023gp}.  
In addition, we construct a task ``CelebA-HQ \citep{karras2017progressive} v.s. Bitmoji \citep{bitmojifaces}'' (CB). 
In this task, profile photos of celebrities are translated to cartoonized bitmoji figures, and vice versa.
We intentionally choose these two domains to make the translation challenging:
The profile photos have rich details and are diverse in terms of face orientation, expression, hair style, etc., but the Bitmoji pictures have a relatively small set of choices of these attributes (e.g., they are always front-facing).
{More details of the datasets are in Sec. \ref{sec:app_dataset_details} in the supplementary material.} 

\textbf{Baselines.} 
The baselines include some representative UDT methods and some recent developments, i.e., \texttt{GP-UNIT} \citep{yang2023gp}, \texttt{Hneg-SRC} \citep{jung2022exploring}, {\texttt{OverLORD} \citep{gabbay2021scaling}, \texttt{ZeroDIM} \citep{gabbay2021image}}, \texttt{StarGAN-v2} \citep{choi2020starganv2}, \texttt{U-GAT-IT} \citep{kim2019ugatit}, \texttt{MUNIT} \citep{huang2018multimodal}, \texttt{UNIT} \citep{liu2017unit}, and \texttt{CycleGAN} \citep{zhu2017unpaired}. 
{In particular, two versions of CycleGAN are used. ``\texttt{CycleGAN Loss}'' refers to the plan-vanilla CycleGAN objective in \eqref{eq:cyclegan} and \texttt{CycleGAN+Id} refers to the ``identity-regularized'' version in \citep{zhu2017unpaired}.} { \texttt{ZeroDIM} uses the same auxiliary information as that used by the proposed method.}

\begin{wrapfigure}{r}{3.1in}
    \vspace{-0.5cm}
    \centering
    \includegraphics[width=\linewidth]{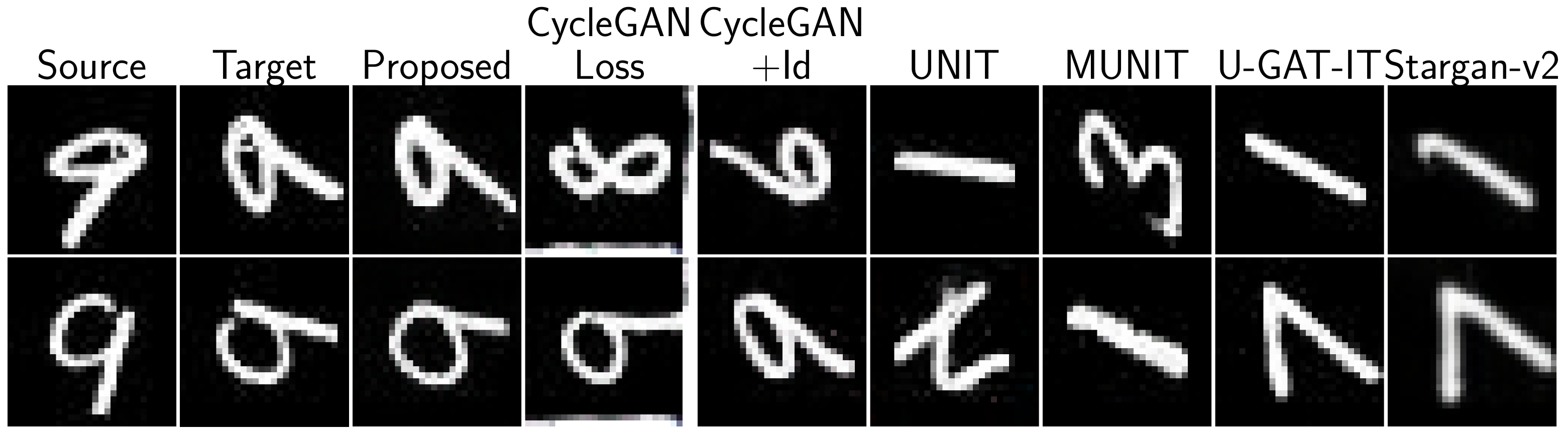}
    \caption{Translation from MNIST to rotated MNIST.}
    \label{fig:mnist_shoes_result}
    \vspace{-0.25cm}
% \end{figure}
\end{wrapfigure}

\textbf{MNIST to Rotated MNIST.}
% \begin{figure}
Fig.~\ref{fig:mnist_shoes_result} shows the results. In this case, we use $u\in\{1,\ldots,10\}$, i.e., the labels of the identity of digits, as the alphabet of the auxiliary variable. Note that knowing such labels does not mean that the cross-domain pairs $(\x,\y)$ are known. { Alternatively, one can also use digit shapes as the alphabets (see Sec. \ref{app:additional_results}).} One can see that \texttt{DIMENSION} learns to translate the digits to their corresponding rotated versions. But the baselines sometimes misalign the samples. The results are consistent with our analysis ({see Sec. \ref{app:additional_results}  for more results}).

\begin{figure}[t]
    \centering
    \includegraphics[width=0.95\linewidth]{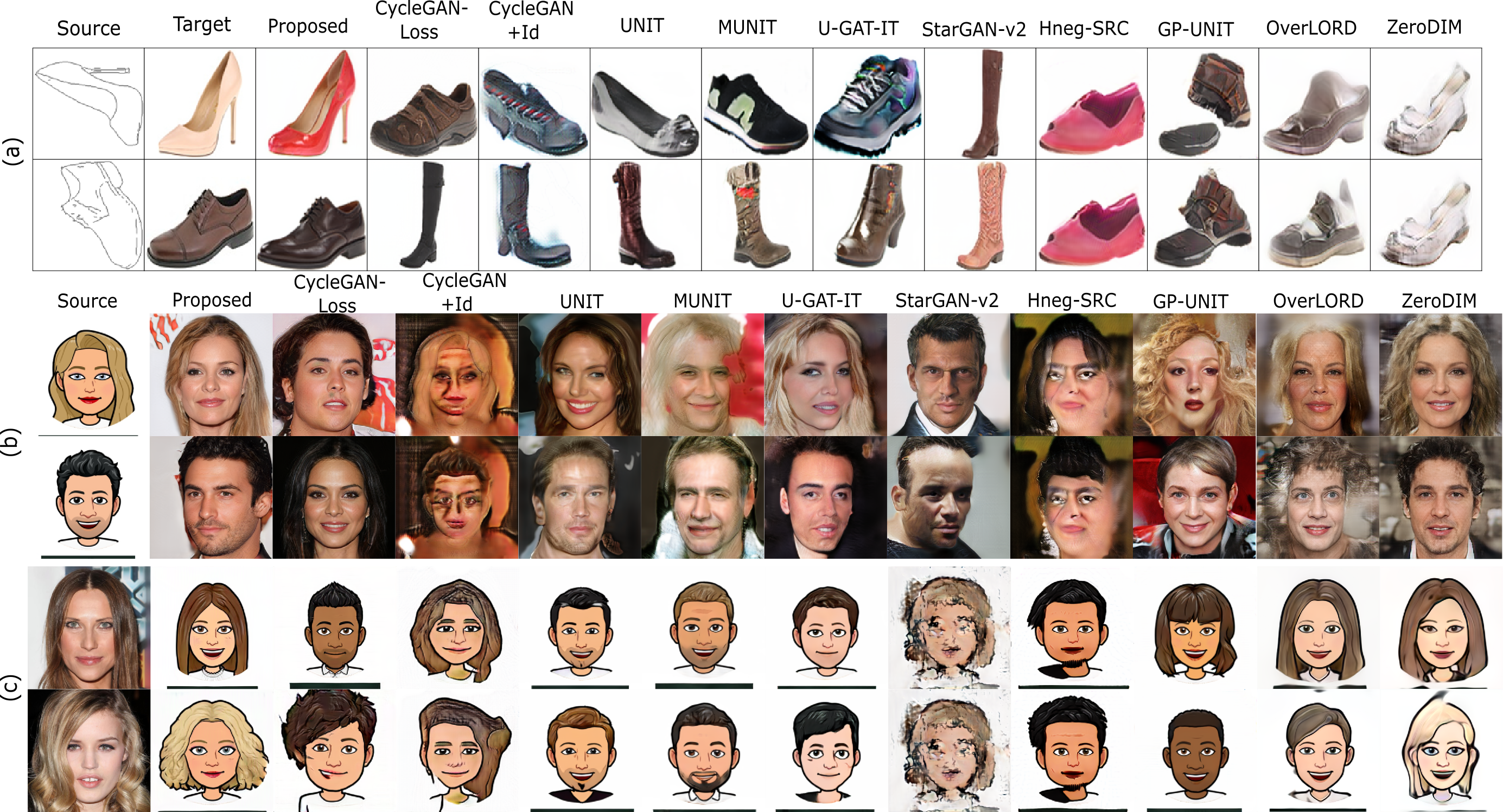}
    \caption{Qualitative results on (a) Edges to Rotated Shoes, (b) Bitmoji Faces to CelebA-HQ, and (c) CelebA-HQ to Bitmoji Faces tasks. More comprehensive illustrations are in the appendix.}
   \vspace{-.6cm}
    \label{fig:real_data_glimpse}
\end{figure}

\textbf{Edges to Rotated Shoes.} From Fig.~\ref{fig:real_data_glimpse} (a), one can see that
the baselines all misalign the edges with wrong shoes. 
Instead, the proposed \texttt{DIMENSION}, using the shoe types (shoes, boots, sandals, and slippers) as the alphabet of $u$, does not encounter this issue. More experiments including the reverse translation (i.e., shoes to edges) are in Sec. \ref{app:additional_results} in the supplementary material.

\textbf{CelebA-HQ and Bitmoji.}
Figs.~\ref{fig:real_data_glimpse} (b)-(c) show the results.
The proposed method uses $u\in$ \{`\texttt{male}',`\texttt{female}',`\texttt{black hair}',`\texttt{non-black hair}'\}. 
{To obtain the auxilliary information for each sample, we use CLIP to automatically annotate the images.}
A remark is that translating from the Bitmoji domain to the CelebA-HQ domain [see. Fig.~\ref{fig:real_data_glimpse} (b)] is particularly hard. This is because the learned translation function needs to ``fill in'' a lot of details to make the generated profiles photorealistic. Our method clearly outperforms the baselines in both directions of translation; see more in Sec. \ref{app:additional_results} in the supplemenary material.

\textbf{Metrics and Quantative Evaluation.} We employ two widely adopted metrics in UDT.
The first is the {\it learned perceptual image patch similarity} (\texttt{LPIPS}) \citep{zhang2018unreasonable}, which leverages the known ground-truth correspondence between $(\x,\y)$. \texttt{LPIPS} measures the ``perceptual distance'' between the translated images and the ground-truth target images.
In addition, we also use the {\it Fr\'echet inception distance} (\texttt{FID}) score \citep{heusel2017gans} in all tasks. \texttt{FID} measures the visual quality of the learned translation using a distribution divergence between the translated images and the target domain. 
In short, \texttt{LPIPS} and \texttt{FID} correspond to the content alignment performance and the target domain-attaining ability, respectively; see details of the metrics Sec. \ref{sec:app_dataset_details}.

Table \ref{tab:lpips} shows the \texttt{LPIPS} scores over the first two datasets where the ground-truth pairs are known. One can see that \texttt{DIMENSION} significantly outperforms the baselines---which is a result of good content alignment. 
The \texttt{FID} scores in the same table show that our method produces translated images that have similar characteristics of the target domains.
The \texttt{FID} scores output by our method are either the lowest or the second lowest.

{{\bf Detailed Settings and More Experiments.} See Sec.~\ref{app:synthetic}-\ref{app:robustness_auxiliary} for settings and more results.}

\begin{table}[t]
    \centering
    
    \caption{\texttt{LPIPS} scores for the ErS and MrM tasks and \texttt{FID} scores for all tasks. E: Edges, rS: rotated Shoes, M: MNIST, rM: rotated MNIST, C: CelebA-HQ, B: Bitmoji faces. }
    \label{tab:lpips}
    \resizebox{.8\linewidth}{!}{
    \bgroup
    \def\arraystretch{1.5}
    \begin{threeparttable}
        \begin{tabular}{r|cccc|rrrrrr}
        
            \toprule
            \multirow{2}{*}{\textbf{Method}} & \multicolumn{4}{c|}{\textbf{\texttt{LPIPS}} ($\downarrow$)} & \multicolumn{6}{c}{\textbf{\texttt{FID}} ($\downarrow$)} \\ \cmidrule{2-11}
             &\textbf{ E $\to$ rS }& \textbf{rS $\to$ E} & \textbf{M $\to$ rM} & \textbf{rM $\to$ M} & \multicolumn{1}{c}{\textbf{E}} & \multicolumn{1}{c}{\textbf{rS}} & \multicolumn{1}{c}{\textbf{M}} & \multicolumn{1}{c}{\textbf{rM}} & \multicolumn{1}{c}{\textbf{C}} & \multicolumn{1}{c}{\textbf{B}} \\ \midrule
            Proposed      & \textbf{0.29 $\pm$ 0.06} & \textbf{0.35 $\pm$ 0.10} & \textbf{0.11  $\pm$  0.08} & \textbf{0.09 $\pm$  0.04} & \textbf{21.47} & \textbf{40.14 } & 13.95           &   16.07          &\textbf{ 32.03}   &   \textbf{20.50} \\
            CycleGAN-Loss & $0.43 \pm 0.06$ & $ 0.50 \pm 0.07$                  & $0.34  \pm  0.07$ & $ 0.33 \pm  0.09 $                 & 35.83          & 55.42           & 16.09           &   16.11          & 36.71            &   28.02          \\
            CycleGAN      & $0.65 \pm 0.03$ & $ 0.54 \pm 0.07$                  & $0.27  \pm  0.09$ & $ 0.28 \pm  0.09 $                 & 259.31         & 130.84          & 46.05           &   34.01          & 196.52           &   85.05          \\
            U-GAT-IT      & $0.56 \pm 0.05$ & $ 0.48 \pm 0.07$                  & $0.25  \pm  0.09$ & $ 0.25 \pm  0.09 $                 & 288.03         & 58.20           & \textbf{11.78}  &   \textbf{11.67} & 50.28            &   39.09          \\
            UNIT          & $0.49 \pm 0.03$ & $ 0.58 \pm 0.03$                  & $0.25  \pm  0.06$ & $ 0.25 \pm  0.08 $                 & 33.95          & 96.28           & 20.44           &   19.15          & 53.63            &   33.56          \\ 
            MUNIT         & $0.50 \pm 0.03$ & $ 0.58 \pm 0.04$.                 & $0.28  \pm  0.09$ & $ 0.28 \pm  0.09 $                 & 43.83          & 86.68           & 14.89           &   15.96          & 62.49            &   27.59          \\
            StarGAN-v2    & $0.39 \pm 0.05$ & $ 0.52 \pm 0.11$                  & $0.28  \pm  0.09$ & $ 0.29 \pm  0.10 $                 & 75.46          & 138.34          & 30.07           &   32.20          & 35.44            &   282.98         \\
            Hneg-SRC      & $0.45 \pm 0.06$ & $ 0.50 \pm 0.07$                  & --\tnote{~}        &   --                                 & 210.27         & 198.77          & --              &   --             & 129.34           &   66.36          \\
            GP-UNIT       & $0.49 \pm 0.08$ & $ 0.44 \pm 0.05$                  & --                 &   --                                 & 231.31         & 96.32           & --              &   --             & 32.40            &   30.30          \\ 
            {OverLORD}     & {$0.43 \pm 0.06$} & {$ 0.42 \pm 0.05$}                  & --                 &   --                                 &  {101.14}       &  {124.02}          & --              &   --             & {76.10}            & {31.08}          \\ 
            {ZeroDIM}      & {$0.38 \pm 0.06$} & {$ 0.41 \pm 0.07$}                  & --                 &   --                                 & {85.56}         & {187.45}           & --              &   --             & {88.36}            & {36.21}          \\ \bottomrule
            \bottomrule
        \end{tabular}
        \begin{tablenotes}
           \item [~] ``--'' means that method is not applicable to the dataset due to small resolution.
         \end{tablenotes}
      \end{threeparttable}
    \egroup
    }
    \vspace{-.5cm}
    \end{table}

\section{Conclusion}
In this work, we revisited the UDT and took a deep look at a core theoretical challenge, namely, the translation identifiability issue. Existing UDT approaches (such as CycleGAN) often lack translation identifiability and may produce content-misaligned translations. This issue largely attributes to the presence of MPA in the solution space of their distribution matching modules.
Our approach leverages the existence of domain-invariant auxiliary variables to establish translation identifiability, using a novel diversified distribution matching criterion.
To our best knowledge,
the identifiability result stands as the first of its kind, without using restrictive conditions on the structure of the desired translation functions.
We also analyzed the robustness of proposed method when the key sufficient condition for identifiability is violated.
Our identifiability theory leads to an easy-to-implement UDT system. Synthetic and real-data experiments corroborated with our theoretical findings.

{\bf Limitations.} Our work considers a model where the ground-truth translation functions are deterministic and bijective. This setting has been (implicitly or explicitly) adopted by a large number of existing works, with the most notable representative being CycleGAN.
However, there can be multiple ``correct'' translation functions in UDT, as the same ``content'' can be combined with various ``styles''.
Such cases may be modeled using probabilistic translation mechanisms {\citep{huang2018multimodal, choi2020starganv2, yang2023gp}}, yet the current analytical framework needs a significant revision to accommodate the probabilistic setting.
In addition, our method makes use of auxiliary variables that may be nontrivial to acquire in certain cases. We have shown that open-sourced foundation models such as CLIP can help acquire such auxiliary variables and that the method is robust to noisy/wrong auxiliary variables (see Sec.~\ref{app:robustness_auxiliary}).
However, it is still of great interest to develop provable UDT translation schemes without using auxiliary variables.

\clearpage

\textbf{Acknowledgement.} This work is supported  in part by the Army Research Office (ARO) under Project ARO W911NF-21-1-0227, and in part by the National Science Foundation (NSF) CAREER Award ECCS-2144889.

\bibliographystyle{iclr2024_conference}
\bibliography{main}

\begin{thebibliography}{64}
\providecommand{\natexlab}[1]{#1}
\providecommand{\url}[1]{\texttt{#1}}
\expandafter\ifx\csname urlstyle\endcsname\relax
  \providecommand{\doi}[1]{doi: #1}\else
  \providecommand{\doi}{doi: \begingroup \urlstyle{rm}\Url}\fi

\bibitem[Amodio \& Krishnaswamy(2019)Amodio and Krishnaswamy]{amodio2019travelgan}
Matthew Amodio and Smita Krishnaswamy.
\newblock {TravelGAN}: Image-to-image translation by transformation vector learning.
\newblock In \emph{Proceedings of IEEE/CVF Computer Vision and Pattern Recognition (CVPR)}, pp.\  8983--8992, 2019.

\bibitem[Ba et~al.(2016)Ba, Kiros, and Hinton]{ba2016layer}
Jimmy~Lei Ba, Jamie~Ryan Kiros, and Geoffrey~E Hinton.
\newblock Layer normalization.
\newblock \emph{arXiv preprint arXiv:1607.06450}, 2016.

\bibitem[Bartlett et~al.(2017)Bartlett, Foster, and Telgarsky]{bartlett2017spectrally}
Peter~L Bartlett, Dylan~J Foster, and Matus~J Telgarsky.
\newblock Spectrally-normalized margin bounds for neural networks.
\newblock \emph{Advances in Neural Information Processing Systems (NeurIPS)}, 30, 2017.

\bibitem[Carothers(2000)]{carothers2000real}
Neal~L Carothers.
\newblock \emph{Real analysis}.
\newblock Cambridge University Press, 2000.

\bibitem[Chakrabarty \& Das(2022)Chakrabarty and Das]{chakrabarty2022translation}
Anish Chakrabarty and Swagatam Das.
\newblock On translation and reconstruction guarantees of the cycle-consistent generative adversarial networks.
\newblock \emph{Advances in Neural Information Processing Systems (NeurIPS)}, 35:\penalty0 23607--23620, 2022.

\bibitem[Choi et~al.(2018)Choi, Choi, Kim, Ha, Kim, and Choo]{choi2018stargan}
Yunjey Choi, Minje Choi, Munyoung Kim, Jung-Woo Ha, Sunghun Kim, and Jaegul Choo.
\newblock {StarGAN}: Unified generative adversarial networks for multi-domain image-to-image translation.
\newblock In \emph{Proceedings of IEEE/CVF Computer Vision and Pattern Recognition (CVPR)}, pp.\  8789--8797, 2018.

\bibitem[Choi et~al.(2020)Choi, Uh, Yoo, and Ha]{choi2020starganv2}
Yunjey Choi, Youngjung Uh, Jaejun Yoo, and Jung-Woo Ha.
\newblock {StarGAN v2}: Diverse image synthesis for multiple domains.
\newblock In \emph{Proceedings of IEEE/CVF Computer Vision and Pattern Recognition (CVPR)}, pp.\  8188--8197, 2020.

\bibitem[Conneau et~al.(2017)Conneau, Lample, Ranzato, Denoyer, and J{\'e}gou]{conneau2017word}
Alexis Conneau, Guillaume Lample, Marc'Aurelio Ranzato, Ludovic Denoyer, and Herv{\'e} J{\'e}gou.
\newblock Word translation without parallel data.
\newblock \emph{arXiv preprint arXiv:1710.04087}, 2017.

\bibitem[Courty et~al.(2017)Courty, Flamary, Habrard, and Rakotomamonjy]{courty2017joint}
Nicolas Courty, R{\'e}mi Flamary, Amaury Habrard, and Alain Rakotomamonjy.
\newblock Joint distribution optimal transportation for domain adaptation.
\newblock \emph{Advances in Neural Information Processing Systems (NeurIPS)}, 30, 2017.

\bibitem[Darmois(1951)]{darmois1951analyse}
George Darmois.
\newblock Analyse des liaisons de probabilit{\'e}.
\newblock In \emph{Proceedings of International Statistic Conferences}, pp.\  231, 1951.

\bibitem[Daunhawer et~al.(2023)Daunhawer, Bizeul, Palumbo, Marx, and Vogt]{daunhawer2023identifiability}
Imant Daunhawer, Alice Bizeul, Emanuele Palumbo, Alexander Marx, and Julia~E Vogt.
\newblock Identifiability results for multimodal contrastive learning.
\newblock \emph{arXiv preprint arXiv:2303.09166}, 2023.

\bibitem[de~B{\'e}zenac et~al.(2021)de~B{\'e}zenac, Ayed, and Gallinari]{de2021cyclegan}
Emmanuel de~B{\'e}zenac, Ibrahim Ayed, and Patrick Gallinari.
\newblock {CycleGAN} through the lens of (dynamical) optimal transport.
\newblock In \emph{Proceedings of Joint European Conference on Machine Learning and Knowledge Discovery in Databases (ECML)}, pp.\  132--147. Springer, 2021.

\bibitem[Fu et~al.(2019)Fu, Huang, Sidiropoulos, and Ma]{fu2019nonnegative}
Xiao Fu, Kejun Huang, Nicholas~D Sidiropoulos, and Wing-Kin Ma.
\newblock Nonnegative matrix factorization for signal and data analytics: Identifiability, algorithms, and applications.
\newblock \emph{IEEE Signal Processing Magazine}, 36\penalty0 (2):\penalty0 59--80, 2019.

\bibitem[Gabbay \& Hoshen(2021)Gabbay and Hoshen]{gabbay2021scaling}
Aviv Gabbay and Yedid Hoshen.
\newblock Scaling-up disentanglement for image translation.
\newblock In \emph{Proceedings of the IEEE/CVF International Conference on Computer Vision (CVPR)}, pp.\  6783--6792, 2021.

\bibitem[Gabbay et~al.(2021)Gabbay, Cohen, and Hoshen]{gabbay2021image}
Aviv Gabbay, Niv Cohen, and Yedid Hoshen.
\newblock An image is worth more than a thousand words: Towards disentanglement in the wild.
\newblock \emph{Advances in Neural Information Processing Systems (NeurIPS)}, 34:\penalty0 9216--9228, 2021.

\bibitem[Galanti et~al.(2018{\natexlab{a}})Galanti, Benaim, and Wolf]{galanti2018generalization}
Tomer Galanti, Sagie Benaim, and Lior Wolf.
\newblock Generalization bounds for unsupervised cross-domain mapping with {WGAN}s.
\newblock \emph{arXiv preprint arXiv:1807.08501}, 2018{\natexlab{a}}.

\bibitem[Galanti et~al.(2018{\natexlab{b}})Galanti, Wolf, and Benaim]{galanti2017role}
Tomer Galanti, Lior Wolf, and Sagie Benaim.
\newblock The role of minimal complexity functions in unsupervised learning of semantic mappings.
\newblock In \emph{Proceedings of International Conference on Learning Representations (ICLR)}, 2018{\natexlab{b}}.

\bibitem[Galanti et~al.(2021)Galanti, Benaim, and Wolf]{galanti2021risk}
Tomer Galanti, Sagie Benaim, and Lior Wolf.
\newblock Risk bounds for unsupervised cross-domain mapping with ipms.
\newblock \emph{The Journal of Machine Learning Research}, 22\penalty0 (1):\penalty0 4019--4060, 2021.

\bibitem[Ganin et~al.(2016)Ganin, Ustinova, Ajakan, Germain, Larochelle, Laviolette, Marchand, and Lempitsky]{ganin2016domain}
Yaroslav Ganin, Evgeniya Ustinova, Hana Ajakan, Pascal Germain, Hugo Larochelle, Fran{\c{c}}ois Laviolette, Mario Marchand, and Victor Lempitsky.
\newblock Domain-adversarial training of neural networks.
\newblock \emph{Journal of Machine Learning Research (JMLR)}, 17:\penalty0 2096--2030, 2016.

\bibitem[Goodfellow et~al.(2014)Goodfellow, Pouget-Abadie, Mirza, Xu, Warde-Farley, Ozair, Courville, and Bengio]{goodfellow2014generative}
Ian Goodfellow, Jean Pouget-Abadie, Mehdi Mirza, Bing Xu, David Warde-Farley, Sherjil Ozair, Aaron Courville, and Yoshua Bengio.
\newblock Generative adversarial networks.
\newblock In \emph{Advances in Neural Information Processing Systems (NeurIPS)}, 2014.

\bibitem[Gulrajani \& Hashimoto(2022)Gulrajani and Hashimoto]{gulrajani2022identifiability}
Ishaan Gulrajani and Tatsunori Hashimoto.
\newblock Identifiability conditions for domain adaptation.
\newblock In \emph{Proceedings of International Conference on Machine Learning (ICML)}, pp.\  7982--7997, 2022.

\bibitem[Heusel et~al.(2017)Heusel, Ramsauer, Unterthiner, Nessler, and Hochreiter]{heusel2017gans}
Martin Heusel, Hubert Ramsauer, Thomas Unterthiner, Bernhard Nessler, and Sepp Hochreiter.
\newblock {GAN}s trained by a two time-scale update rule converge to a local {Nash} equilibrium.
\newblock \emph{Advances in Neural Information Processing Systems (NeurIPS)}, 30, 2017.

\bibitem[Huang et~al.(2015)Huang, Feris, Chen, and Yan]{huang2015cross}
Junshi Huang, Rogerio~S Feris, Qiang Chen, and Shuicheng Yan.
\newblock Cross-domain image retrieval with a dual attribute-aware ranking network.
\newblock In \emph{Proceedings of the IEEE International Conference on Computer Vision (ICCV)}, pp.\  1062--1070, 2015.

\bibitem[Huang et~al.(2018)Huang, Liu, Belongie, and Kautz]{huang2018multimodal}
Xun Huang, Ming-Yu Liu, Serge Belongie, and Jan Kautz.
\newblock Multimodal unsupervised image-to-image translation.
\newblock In \emph{Proceedings of European Conference on Computer Vision (ECCV)}, pp.\  172--189, 2018.

\bibitem[Hyv{\"a}rinen \& Pajunen(1999)Hyv{\"a}rinen and Pajunen]{hyvarinen1999nonlinear}
Aapo Hyv{\"a}rinen and Petteri Pajunen.
\newblock Nonlinear independent component analysis: Existence and uniqueness results.
\newblock \emph{Neural networks}, 12\penalty0 (3):\penalty0 429--439, 1999.

\bibitem[Hyvarinen et~al.(2019)Hyvarinen, Sasaki, and Turner]{hyvarinen2019nonlinear}
Aapo Hyvarinen, Hiroaki Sasaki, and Richard Turner.
\newblock Nonlinear ica using auxiliary variables and generalized contrastive learning.
\newblock In \emph{Proceedings of International Conference on Artificial Intelligence and Statistics (AISTATS)}, pp.\  859--868. PMLR, 2019.

\bibitem[Isola et~al.(2017)Isola, Zhu, Zhou, and Efros]{isola2017image}
Phillip Isola, Jun-Yan Zhu, Tinghui Zhou, and Alexei~A Efros.
\newblock Image-to-image translation with conditional adversarial networks.
\newblock In \emph{Proceedings of IEEE/CVF Computer Vision and Pattern Recognition (CVPR)}, pp.\  1125--1134, 2017.

\bibitem[Jung et~al.(2022)Jung, Kwon, and Ye]{jung2022exploring}
Chanyong Jung, Gihyun Kwon, and Jong~Chul Ye.
\newblock Exploring patch-wise semantic relation for contrastive learning in image-to-image translation tasks.
\newblock In \emph{Proceedings of the IEEE/CVF Conference on Computer Vision and Pattern Recognition (CVPR)}, pp.\  18260--18269, 2022.

\bibitem[Karras et~al.(2017)Karras, Aila, Laine, and Lehtinen]{karras2017progressive}
Tero Karras, Timo Aila, Samuli Laine, and Jaakko Lehtinen.
\newblock Progressive growing of {GAN}s for improved quality, stability, and variation.
\newblock \emph{arXiv preprint arXiv:1710.10196}, 2017.

\bibitem[Kim et~al.(2020)Kim, Kim, Kang, and Lee]{kim2019ugatit}
Junho Kim, Minjae Kim, Hyeonwoo Kang, and Kwanghee Lee.
\newblock {U-GAT-IT}: Unsupervised generative attentional networks with adaptive layer-instance normalization for image-to-image translation.
\newblock In \emph{Proceedings of International Conference on Learning Representations (ICLR)}, 2020.

\bibitem[Kim et~al.(2017)Kim, Cha, Kim, Lee, and Kim]{kim2017learning}
Taeksoo Kim, Moonsu Cha, Hyunsoo Kim, Jung~Kwon Lee, and Jiwon Kim.
\newblock Learning to discover cross-domain relations with generative adversarial networks.
\newblock In \emph{Proceedings of International Conference on Machine Learning (ICML)}, pp.\  1857--1865, 2017.

\bibitem[Kingma \& Ba(2015)Kingma and Ba]{kingma2015adam}
Diederik~P. Kingma and Jimmy Ba.
\newblock Adam: {A} method for stochastic optimization.
\newblock In \emph{Proceedings of International Conference on Learning Representations (ICLR)}, 2015.

\bibitem[Krizhevsky et~al.(2012)Krizhevsky, Sutskever, and Hinton]{krizhevsky2012imagenet}
Alex Krizhevsky, Ilya Sutskever, and Geoffrey~E Hinton.
\newblock Imagenet classification with deep convolutional neural networks.
\newblock \emph{Advances in Neural Information Processing Systems (NeurIPS)}, 25, 2012.

\bibitem[Lample et~al.(2017)Lample, Conneau, Denoyer, and Ranzato]{lample2017unsupervised}
Guillaume Lample, Alexis Conneau, Ludovic Denoyer, and Marc'Aurelio Ranzato.
\newblock Unsupervised machine translation using monolingual corpora only.
\newblock \emph{arXiv preprint arXiv:1711.00043}, 2017.

\bibitem[LeCun et~al.(2010)LeCun, Cortes, and Burges]{lecun2010mnist}
Yann LeCun, Corinna Cortes, and CJ~Burges.
\newblock {MNIST} handwritten digit database.
\newblock \emph{ATT Labs [Online]. Available: http://yann.lecun.com/exdb/mnist}, 2, 2010.

\bibitem[Li et~al.(2019)Li, Hu, Zhang, Hong, Ye, Wu, and Ji]{li2019attribute}
Xinyang Li, Jie Hu, Shengchuan Zhang, Xiaopeng Hong, Qixiang Ye, Chenglin Wu, and Rongrong Ji.
\newblock Attribute guided unpaired image-to-image translation with semi-supervised learning.
\newblock \emph{arXiv preprint arXiv:1904.12428}, 2019.

\bibitem[Liu \& Tuzel(2016)Liu and Tuzel]{liu2016coupled}
Ming-Yu Liu and Oncel Tuzel.
\newblock Coupled generative adversarial networks.
\newblock \emph{Advances in Neural Information Processing Systems (NeurIPS)}, 29, 2016.

\bibitem[Liu et~al.(2017)Liu, Breuel, and Kautz]{liu2017unit}
Ming-Yu Liu, Thomas Breuel, and Jan Kautz.
\newblock Unsupervised image-to-image translation networks.
\newblock In \emph{Advances in Neural Information Processing Systems (NeurIPS)}, volume~30, 2017.

\bibitem[Liu et~al.(2019)Liu, Huang, Mallya, Karras, Aila, Lehtinen, and Kautz]{liu2019few}
Ming-Yu Liu, Xun Huang, Arun Mallya, Tero Karras, Timo Aila, Jaakko Lehtinen, and Jan Kautz.
\newblock Few-shot unsupervised image-to-image translation.
\newblock In \emph{Proceedings of the IEEE/CVF International Conference on Computer Vision (CVPR)}, pp.\  10551--10560, 2019.

\bibitem[Lyu et~al.(2022)Lyu, Fu, Wang, and Lu]{lyu2021understanding}
Qi~Lyu, Xiao Fu, Weiran Wang, and Songtao Lu.
\newblock Understanding latent correlation-based multiview learning and self-supervision: An identifiability perspective.
\newblock In \emph{Proceedings of International Conference on Learning Representations (ICLR)}, 2022.

\bibitem[Maas et~al.(2013)Maas, Hannun, and Ng]{maas2013rectifier}
Andrew~L Maas, Awni~Y Hannun, and Andrew~Y Ng.
\newblock Rectifier nonlinearities improve neural network acoustic models.
\newblock In \emph{Proceedings of International Conference on Machine Learning (ICML)}, volume~30, pp.\ ~3, 2013.

\bibitem[Mao et~al.(2017)Mao, Li, Xie, Lau, Wang, and Paul~Smolley]{mao2017least}
Xudong Mao, Qing Li, Haoran Xie, Raymond~YK Lau, Zhen Wang, and Stephen Paul~Smolley.
\newblock Least squares generative adversarial networks.
\newblock In \emph{Proceedings of International Conference on Computer Vision (ICCV)}, pp.\  2794--2802, 2017.

\bibitem[Mescheder et~al.(2018)Mescheder, Geiger, and Nowozin]{mescheder2018training}
Lars Mescheder, Andreas Geiger, and Sebastian Nowozin.
\newblock Which training methods for {GAN}s do actually converge?
\newblock In \emph{Proceedings of International Conference on Machine Learning (ICML)}, pp.\  3481--3490. PMLR, 2018.

\bibitem[Moriakov et~al.(2020)Moriakov, Adler, and Teuwen]{moriakov2020kernel}
Nikita Moriakov, Jonas Adler, and Jonas Teuwen.
\newblock Kernel of {CycleGAN} as a principle homogeneous space.
\newblock In \emph{Proceedings of International Conference on Learning Representations (ICLR)}, 2020.

\bibitem[Mozafari(2020)]{bitmojifaces}
Mostafa Mozafari.
\newblock Bitmoji faces.
\newblock \url{https://www.kaggle.com/datasets/mostafamozafari/bitmoji-faces}, 2020.
\newblock Accessed on September 20th, 2023.

\bibitem[Mustafa \& Mantiuk(2020)Mustafa and Mantiuk]{mustafa2020transformation}
Aamir Mustafa and Rafa{\l}~K Mantiuk.
\newblock Transformation consistency regularization: {A} semi-supervised paradigm for image-to-image translation.
\newblock In \emph{Proceedings of the IEEE/CVF Conference on Computer Vision and Pattern Recognition (CVPR)}, pp.\  599--615, 2020.

\bibitem[Pang et~al.(2021)Pang, Lin, Qin, and Chen]{pang2021image}
Yingxue Pang, Jianxin Lin, Tao Qin, and Zhibo Chen.
\newblock Image-to-image translation: Methods and applications.
\newblock \emph{IEEE Transactions on Multimedia}, 24:\penalty0 3859--3881, 2021.

\bibitem[Park et~al.(2020)Park, Efros, Zhang, and Zhu]{park2020contrastive}
Taesung Park, Alexei~A Efros, Richard Zhang, and Jun-Yan Zhu.
\newblock Contrastive learning for unpaired image-to-image translation.
\newblock In \emph{Proceedings of European Conference on Computer Vision (ECCV)}, pp.\  319--345, 2020.

\bibitem[Radford et~al.(2021)Radford, Kim, Hallacy, Ramesh, Goh, Agarwal, Sastry, Askell, Mishkin, Clark, et~al.]{radford2021learning}
Alec Radford, Jong~Wook Kim, Chris Hallacy, Aditya Ramesh, Gabriel Goh, Sandhini Agarwal, Girish Sastry, Amanda Askell, Pamela Mishkin, Jack Clark, et~al.
\newblock Learning transferable visual models from natural language supervision.
\newblock In \emph{Proceedings of International Conference on Machine Learning (ICML)}, pp.\  8748--8763, 2021.

\bibitem[Rudin(1976)]{rudin1976principles}
Walter Rudin.
\newblock \emph{Principles of mathematical analysis}, volume~3.
\newblock McGraw-hill New York, 1976.

\bibitem[Szegedy et~al.(2016)Szegedy, Vanhoucke, Ioffe, Shlens, and Wojna]{szegedy2016rethinking}
Christian Szegedy, Vincent Vanhoucke, Sergey Ioffe, Jon Shlens, and Zbigniew Wojna.
\newblock Rethinking the inception architecture for computer vision.
\newblock In \emph{Proceedings of the IEEE/CVF Conference on Computer Vision and Pattern Recognition (CVPR)}, pp.\  2818--2826, 2016.

\bibitem[Taigman et~al.(2017)Taigman, Polyak, and Wolf]{taigman2016unsupervised}
Yaniv Taigman, Adam Polyak, and Lior Wolf.
\newblock Unsupervised cross-domain image generation.
\newblock In \emph{Proceedings of International Conference on Learning Representations (ICLR)}, 2017.

\bibitem[Von~K{\"u}gelgen et~al.(2021)Von~K{\"u}gelgen, Sharma, Gresele, Brendel, Sch{\"o}lkopf, Besserve, and Locatello]{von2021self}
Julius Von~K{\"u}gelgen, Yash Sharma, Luigi Gresele, Wieland Brendel, Bernhard Sch{\"o}lkopf, Michel Besserve, and Francesco Locatello.
\newblock Self-supervised learning with data augmentations provably isolates content from style.
\newblock In \emph{Advances in Neural Information Processing Systems (NeurIPS)}, volume~34, pp.\  16451--16467, 2021.

\bibitem[Wang et~al.(2018)Wang, Liu, Zhu, Tao, Kautz, and Catanzaro]{wang2018high}
Ting-Chun Wang, Ming-Yu Liu, Jun-Yan Zhu, Andrew Tao, Jan Kautz, and Bryan Catanzaro.
\newblock High-resolution image synthesis and semantic manipulation with conditional {GAN}s.
\newblock In \emph{Proceedings of IEEE/CVF Computer Vision and Pattern Recognition (CVPR)}, pp.\  8798--8807, 2018.

\bibitem[Wang et~al.(2020)Wang, Khan, Gonzalez-Garcia, Weijer, and Khan]{wang2020semi}
Yaxing Wang, Salman Khan, Abel Gonzalez-Garcia, Joost van~de Weijer, and Fahad~Shahbaz Khan.
\newblock Semi-supervised learning for few-shot image-to-image translation.
\newblock In \emph{Proceedings of the IEEE/CVF Conference on Computer Vision and Pattern Recognition (CVPR)}, pp.\  4453--4462, 2020.

\bibitem[Wu et~al.(2019)Wu, Cao, Li, Qian, and Loy]{wu2019transgaga}
Wayne Wu, Kaidi Cao, Cheng Li, Chen Qian, and Chen~Change Loy.
\newblock {TransGaGa}: Geometry-aware unsupervised image-to-image translation.
\newblock In \emph{Proceedings of IEEE/CVF Computer Vision and Pattern Recognition (CVPR)}, pp.\  8012--8021, 2019.

\bibitem[Xie et~al.(2022)Xie, Ho, and Zhang]{xie2022unsupervised}
Shaoan Xie, Qirong Ho, and Kun Zhang.
\newblock Unsupervised image-to-image translation with density changing regularization.
\newblock \emph{Advances in Neural Information Processing Systems (NeurIPS)}, 35:\penalty0 28545--28558, 2022.

\bibitem[Xu et~al.(2022)Xu, Xie, Wu, Zhang, Gong, and Batmanghelich]{xu2022maximum}
Yanwu Xu, Shaoan Xie, Wenhao Wu, Kun Zhang, Mingming Gong, and Kayhan Batmanghelich.
\newblock Maximum spatial perturbation consistency for unpaired image-to-image translation.
\newblock In \emph{Proceedings of IEEE/CVF Computer Vision and Pattern Recognition (CVPR)}, pp.\  18311--18320, 2022.

\bibitem[Yang et~al.(2023)Yang, Jiang, Liu, and Loy]{yang2023gp}
Shuai Yang, Liming Jiang, Ziwei Liu, and Chen~Change Loy.
\newblock Gp-unit: Generative prior for versatile unsupervised image-to-image translation.
\newblock \emph{IEEE Transactions on Pattern Analysis and Machine Intelligence}, 2023.

\bibitem[Yaz et~al.(2018)Yaz, Foo, Winkler, Yap, Piliouras, Chandrasekhar, et~al.]{yaz2018unusual}
Yasin Yaz, Chuan-Sheng Foo, Stefan Winkler, Kim-Hui Yap, Georgios Piliouras, Vijay Chandrasekhar, et~al.
\newblock The unusual effectiveness of averaging in gan training.
\newblock In \emph{International Conference on Learning Representations}, 2018.

\bibitem[Zhang et~al.(2018)Zhang, Isola, Efros, Shechtman, and Wang]{zhang2018unreasonable}
Richard Zhang, Phillip Isola, Alexei~A Efros, Eli Shechtman, and Oliver Wang.
\newblock The unreasonable effectiveness of deep features as a perceptual metric.
\newblock In \emph{Proceedings of the IEEE/CVF Conference on Computer Vision and Pattern Recognition (CVPR)}, pp.\  586--595, 2018.

\bibitem[Zhu et~al.(2017)Zhu, Park, Isola, and Efros]{zhu2017unpaired}
Jun-Yan Zhu, Taesung Park, Phillip Isola, and Alexei~A Efros.
\newblock Unpaired image-to-image translation using cycle-consistent adversarial networks.
\newblock In \emph{Proceedings of IEEE/CVF Computer Vision and Pattern Recognition (CVPR)}, pp.\  2223--2232, 2017.

\bibitem[Zhuang et~al.(2020)Zhuang, Qi, Duan, Xi, Zhu, Zhu, Xiong, and He]{zhuang2020comprehensive}
Fuzhen Zhuang, Zhiyuan Qi, Keyu Duan, Dongbo Xi, Yongchun Zhu, Hengshu Zhu, Hui Xiong, and Qing He.
\newblock A comprehensive survey on transfer learning.
\newblock \emph{Proceedings of the IEEE}, 109\penalty0 (1):\penalty0 43--76, 2020.

\bibitem[Zimmermann et~al.(2021)Zimmermann, Sharma, Schneider, Bethge, and Brendel]{zimmermann2021contrastive}
Roland~S Zimmermann, Yash Sharma, Steffen Schneider, Matthias Bethge, and Wieland Brendel.
\newblock Contrastive learning inverts the data generating process.
\newblock In \emph{Proceedings of International Conference on Machine Learning (ICML)}, pp.\  12979--12990, 2021.

\end{thebibliography}

\clearpage
\begin{center}
    {\large {\bf Supplementary Material of ``Towards Identifiable Unsupervised Domain Translation: A Diversified Distribution Matching Approach''}}
\end{center}

\appendix
\section{Preliminaries}\label{app:definitions}

\subsection{Notation} 

\begin{itemize}
    \item $x$, $\x$, $\cX$ denote a scalar, vector, and a set, respectively.
    \item $p(\x)$  and $p(\x|u)$ denote the marginal \textit{probability density function} (PDF) of $\x$ and conditional PDF of $\x$ conditioned on $u$, respectively. 
    \item $\|\x \|_2$ denotes the $\ell_2$-norm of $\x$.
    \item ${\rm dia}(\cA) = \sup_{\a, \b \in \cA} \|\a - \b\|_2$.
    \item $\mathbb{I} : \cX \to \cX$ denotes the identity function such that $\mathbb{I}(\x) = \x, \forall \x \in \cX$.
    \item $\cA^{\rm c}$, ${\rm cl}(\cA)$, ${\rm bd}(\cA)$ and ${\rm int}(\cA)$ denote the complement, closure, boundary, and the interior of set $\cA$. 
    \item A set $\cA$ is said to have strictly positive measure under $p(\x)$ if and only if $\bP_{\x}[\cA] > 0$. 
    \item For a (random) vector $\x$, $x(i)$ and $[\x]_i$ denote the $i$th element of $\x$, and $\x(i:j)$ denotes $[x(i), x(i+1), \dots, x(j)]$. 
    \item Distance between two sets is defined as
    $${\rm dist}(\cA, \cB) = \inf_{ \a \in \cA, \b \in \cB} \|\a - \b\|_2.$$
    \item Distance between a set and a point is defined as
    $${\rm dist}(\a, \cB) = \inf_{\b \in \cB} \|\a - \b\|_2.$$
    \item  $\cN_{\epsilon}(\z)$ denotes the $\epsilon$-neighborhood of $\z \in \bbR^N$ defined as 
    $$ \cN_{\epsilon}(\z) = \{ \widehat{\z} \in \bbR^N | \|\z - \widehat{\z}\|_2 < \epsilon\}.$$
    \item ${\rm conn}(\cA)$ denotes the set of connected components of $\cA$ (see definition of connected components in Appendix \ref{app:defn}).
    \item For any function $\m: \cW \to \cZ$, and set $\cA \subseteq \cW$, $\m(\cA) = \{\m(\w) \in \cZ~|~ \w \in \cA\}$
\end{itemize}

\subsection{Definitions}\label{app:defn}
We will employ standard notions from real analysis.
We refer the readers to \citep{carothers2000real, rudin1976principles} for precise definitions and more details. Here we provide working definition with illustration.

\begin{figure}[h]
    \centering
    \includegraphics[width=0.7\linewidth]{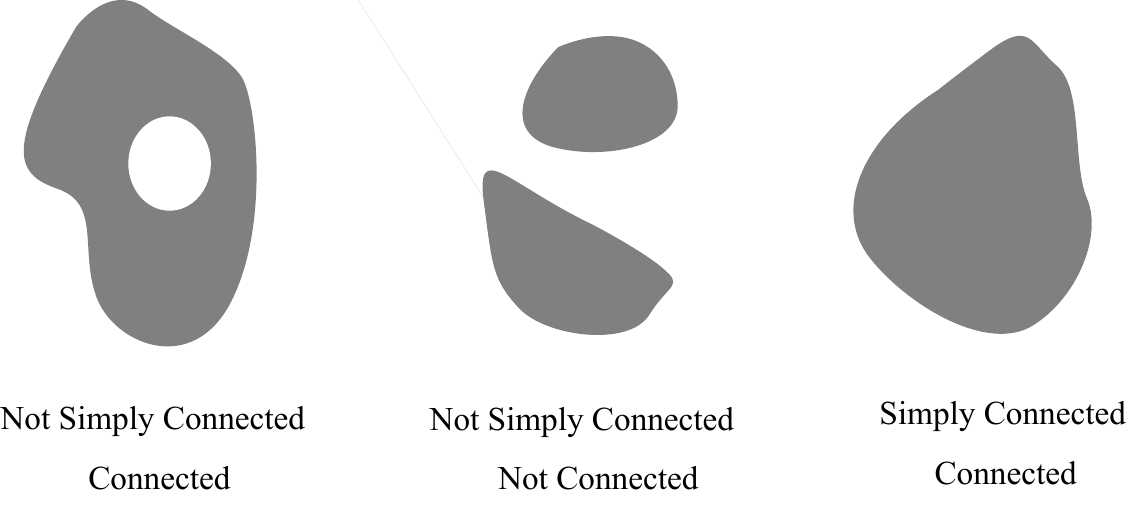}
    \caption{Illustration of connected and simply connected sets}
    \label{fig:connected_sets}
\end{figure}

\textbf{Connected set.} A set $\cC$ is connected (in $\cX$), if and only if there does not exist any disjoint non-empty open sets $\cA, \cB \subset \cX$ such that $\cA \cap \cC \neq \phi$, $\cB \cap  \cC \neq \phi$, and $\cC \subset \cA \cup \cB$ (see Fig. \ref{fig:connected_sets}).

\textbf{Simply connected set.} A simply connected set is a connected set such that any simple closed curve can be shrunk to a point continuously in the set (see Fig. \ref{fig:connected_sets}).

\begin{figure}[ht]
    \centering
    \includegraphics[width=0.6\linewidth]{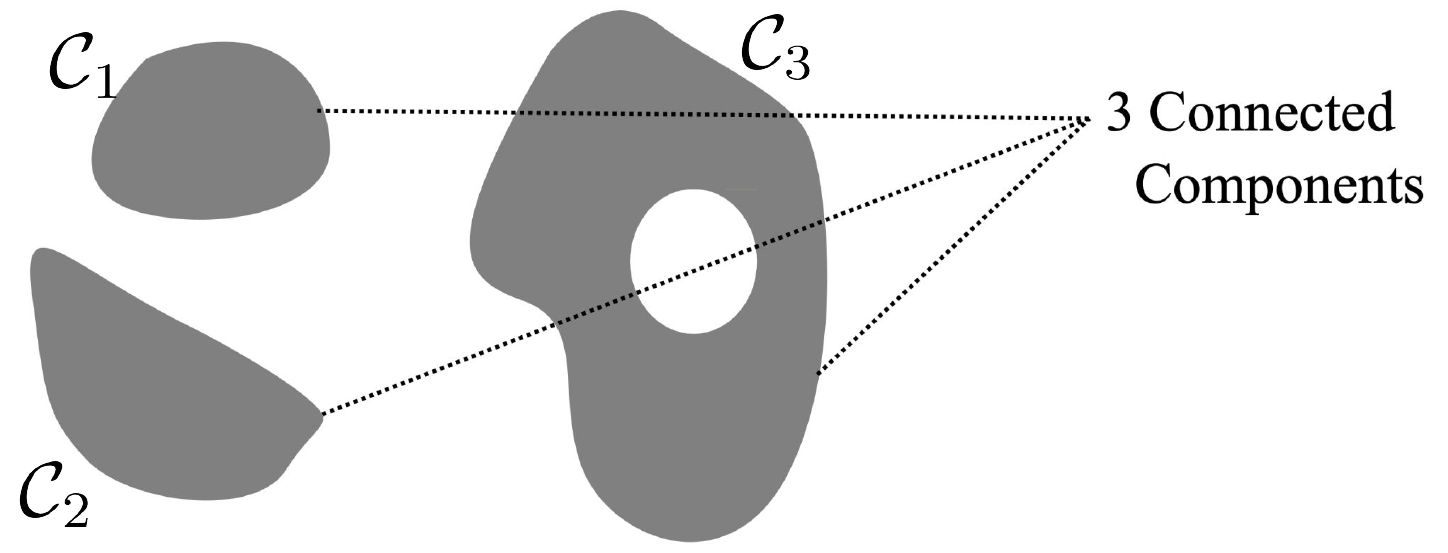}
    \caption{A set $\cA = \cC_1 \cup \cC_2 \cup \cC_3$ with 3 connected components: $\cC_1, \cC_2$, and $\cC_3$.}
    \label{fig:connected_components}
\end{figure}

\textbf{Connected components.} Given a set $\cA$, the maximal connected subsets of $\cA$, such that the subsets are not themselves contained in any other connected subsets of $\cA$, are called connected components of $\cA$. Specifically, a connected set $\cC \subseteq \cA$ is a connected component of $\cA$ if there does not exist any other connected set $\cD \subseteq \cA$, such that $\cC \subset \cD$. In Fig. \ref{fig:connected_components}, $\cA$ denotes the entire shaded regions, and has three connected components $\cC_1$, $\cC_2$, and $\cC_3$. Note that any set can be uniquely written as a disjoint union of its connected components. In Fig. \ref{fig:connected_components}, $\cC_1 \cup \cC_2 \cup \cC_3$ is a unique disjoint union representing $\cA$. 

\textbf{Continuous function.} A function $\m: \cW \to \cZ$, with $\cW \subseteq \bbR^W, \cZ \subseteq \bbR^Z$ is said to be continuous if for any $\w \in \cW$ and $\epsilon >0$, there exists a $\delta>0$ such that 
$$\m \left( (\cN_{\delta}(\w) \cap \cW)\right) \subset \cN_{\epsilon} (\m(\w)) \cap \cZ.$$

\textbf{Continuous and invertible Functions.}
If a function $\m: \cW \to \cZ$ is continuous and invertible, then its inverse $\m^{-1}: \cZ \to \cW$ is also continuous. Some useful properties of continuous and invertible function $\m$ are as follows:
\begin{itemize}
    \item If $\cA \subseteq \cW$ is closed, then $\m(\cA)$ is also closed.
    \item If $\cA \subseteq \cW$ is open, then $\m(\cA)$ is also open.
    \item If $\cA \subseteq \cW$ is connected, then $\m(\cA)$ is also connected.
\end{itemize}

\section{Proof of Lemmas and Facts}
Note that for the ease of reading, the lemmas, facts, and theorems from the main paper are re-stated and highlighted using shaded boxes.
    \renewcommand\theproposition{\ref{prop:mpa_existence}}
    \begin{mdframed}[backgroundcolor=gray!10,topline=false, rightline=false, leftline=false, bottomline=false]
        \begin{proposition}
              Suppose that $\bP_{\x}$ admits a continuous PDF, $p(\x)$ and $p(\x)>0, \forall \x \in \cX$. Assume that $\cX$ is simply connected. Then, there exists a continuous non-trivial (non-identity) $\h(\cdot)$ such that $\h_{\# \bP_{\x}} = \bP_{\x}$. 
        \end{proposition}
    \end{mdframed}
    \setcounter{proposition}{\value{proposition}-1}
    \renewcommand\theproposition{\arabic{proposition}}

    \begin{proof}
        We want to show that there exists a continuous $\h: \cX \to \cX$ such that
        $$ \h_{\# \bP_{\x}} = \bP_{\x}.$$
        To this end, we will construct such MPA by reducing the problem of finding an MPA of $p(\x)$ to finding an MPA of the uniform distribution. Note that one can always construct a continuous invertible function $\d: \cX \to (0,1)^{D_x}$, such that the function maps any continuous distribution with a simply connected support to the uniform distribution. This mapping can be found via the so-called {\it Darmois construction} \citep{darmois1951analyse, hyvarinen1999nonlinear}. 
        Specifically, under the Darmois construction, the $i$th output of $\d(\overline{\x}), \forall \overline{\x} \in \cX$ is given by
        $$ [\d(\overline{\x})]_i := F \left(\overline{x}(i) ~\big|~ \x(1:i-1) = \overline{\x} (1:i-1) \right) , \quad i = 1, \dots, N,$$
        where $F\left(\overline{x}(i) ~\big|~ \cdot \right)$ denotes the conditional CDF of $x(i)$, i.e, 
        $$ F \left(\overline{x}(i) ~\big|~ \x(1:i-1) = \overline{\x} (1:i-1) \right) = \bP_{x(i)|\x(1:i-1) = \overline{\x} (1:i-1)} \left[\{x(i): x(i) \leq \overline{x}(i)\}\right];$$
        see more detailed introduction to the Darmois construction in \citep{hyvarinen1999nonlinear}.

        With the constructed $\bm d$, one can form a continuous mapping $\h: \cX \to \cX$ as follows
        $$ \h = \d^{-1} \circ \h_U \circ \d,$$
        where $\h_U:(0,1)^{D_x} \to (0,1)^{D_x} $ is a continuous MPA on the uniform distribution over $(0,1)^{D_x}$. Since $\d$ is continuous for a continuous distribution, $\h$ is continuous because it is the composition of continuous functions.  
        
        Now, it remains to show that $\h_U$ exists. A simple example of $\h_U$ is reflection around the mean of $d(\x)$, i.e., 
        $${\green \h_U (\z) = -\z + 2\mu},$$
        where $\bm \mu = [\nicefrac{1}{2}, \dots, \nicefrac{1}{2}]^\T \in \bbR^{D_x}$.
        This concludes the proof.
    \end{proof}

        \renewcommand\thelemma{\ref{lemma:fact_sln}}
        \begin{mdframed}[backgroundcolor=gray!10,topline=false, rightline=false, leftline=false, bottomline=false]
        \begin{lemma}
        Assume that an optimal solution of \eqref{eq:split_cyclegan} is $(\widehat{\bm f},\widehat{\bm g},\{ \widehat{\bm d}^{(i)}_x,\widehat{\bm d}^{(i)}_y\})$. Then, under Assumption~\ref{assumption:correspondance}, we have
        $\bP_{\x |u=u_i} = \widehat{\f}_{\# \bP_{\y|u=u_i}}$,  $\bP_{\y |u=u_i}=\widehat{\g}_{\# \bP_{\x|u=u_i}}, ~\forall i \in [I]$, and $\widehat{\f} = \widehat{\g}^{-1}, a.e.$ 
        \end{lemma}
        \end{mdframed}
        \setcounter{lemma}{\value{lemma}-1}
        \renewcommand\thelemma{\arabic{lemma}}
        \begin{proof}
        Fact \ref{lemma:fact_sln} is a direct consequence of \citep[Theorem 1]{goodfellow2014generative}. 
        
        First of all, recall the objective in \eqref{eq:split_cyclegan}:
        \begin{align}
            \min_{\f, \g} \max_{\{\d_{x}^{(i)}, \d_{y}^{(i)}\}} ~ \sum_{i=1}^I \left(\cL_{\rm DSGAN}(\g, \d_{y}^{(i)}, \x, \y) + \cL_{\rm DSGAN}(\f, \d_{x}^{(i)}, \x, \y)\right) + \lambda \cL_{\rm cyc}(\g, \f).
        \end{align}
        The global minimum of $\cL_{\rm DSGAN}(\g, \d_{y}^{(i)}, \x, \y)$ is achieved when \citep[Theorem 1]{goodfellow2014generative}
        \begin{align*}
            \g_{\# \bP_{\x|u = u_i}} = \bP_{\y |u = u_i}.
        \end{align*}
        Similarly, the global minimum of  $\cL_{\rm DSGAN}(\f, \d_{x}^{(i)}, \x, \y)$ is achieved when
        \begin{align*}
            \f_{\# \bP_{\y|u = u_i}} = \bP_{\x |u = u_i}.
        \end{align*}

        Finally, the global minimum of $\cL_{\rm cyc}(\g, \f)$, which is zero, is achieved when
        $$\g = \f^{-1}, a.e.$$

        We know that $\g^\star$ and $\f^\star$ can achieve global minimums of all loss terms simultaneously.
        Hence the solution of \eqref{eq:split_cyclegan}, $\widehat{\f}$ and $\widehat{\g}$, should satisfy

        $$\bP_{\x |u=u_i} = \widehat{\f}_{\# \bP_{\y|u=u_i}},\bP_{\y |u=u_i}=\widehat{\g}_{\# \bP_{\x|u=u_i}}, ~\forall i \in [I], \text{ and } \widehat{\f} = \widehat{\g}^{-1}, a.e.$$       
    \end{proof}

    \renewcommand\theFact{\ref{fact:x_y_symmetry}}
    \begin{mdframed}[backgroundcolor=gray!10,topline=false, rightline=false, leftline=false, bottomline=false]
    \begin{Fact}
        Suppose that Assumption~\ref{assumption:correspondance} holds.
        Then, there exists an admissible MPA of $\{\bP_{\x|u=u_i}\}_{i=1}^I$ if and only if there exists an admissible MPA of $\{\bP_{\y|u=u_i}\}_{i=1}^I$.
    \end{Fact}
    \end{mdframed}
    \setcounter{Fact}{\value{Fact}-1}
    \renewcommand\theFact{\arabic{fact}}
    \begin{proof}
        Let $\h$ be an admissible MPA of $\{\bP_{\x|u=u_i}\}_{i=1}^I$. Then
        \begin{align*}
         \h_{\# \bP_{\x | u = u_i}} ~&=~ \bP_{\x|u = u_i}, \forall i \in [I]. \\
        \iff \g^\star \circ \h_{\# \bP_{\x | u = u_i}} ~&=~ \g^\star_{\# \bP_{\x | u = u_i}}, \forall i \in [I]. \\
        \iff \g^\star \circ \h \circ \f^\star_{\# \bP_{\y | u = u_i}} ~&=~ \bP_{\y|u = u_i}, \forall i \in [I]. \\
        \end{align*}
        This implies that $\g^\star \circ \h \circ \f^\star$ is an admissible MPA of $\{\bP_{\y|u=u_i}\}_{i=1}^I$ if and only if $\h$ is an admissible MPA of $\{\bP_{\x|u=u_i}\}_{i=1}^I$.
        
        Hence, there exists an admissible MPA of $\{\bP_{\x|u=u_i}\}_{i=1}^I$ if and only if there exists an admissible MPA of $\{\bP_{\y|u=u_i}\}_{i=1}^I$.
    \end{proof}
    
\section{Proof of Theorems}
\subsection{Proof of Theorem \ref{thm:identifiability}}\label{app:proof_identifiability}

    \renewcommand\thetheorem{\ref{thm:identifiability}}
    \begin{mdframed}[backgroundcolor=gray!10,topline=false, rightline=false, leftline=false, bottomline=false]
        \begin{theorem}
        Suppose that Assumption~\ref{assumption:correspondance} holds. 
        Let ${\sf E}_{i,j}$ denote the event that the set $\{\bP_{\x|u=u_i},\bP_{\x|u=u_j}\}$ does not satisfy the SDC. Assume that ${\rm Pr}[{\sf E}_{i,j}] \leq \rho$ for any $i \neq j$, where $i,j \in [I]$.
        Let $(\widehat{\f},\widehat{\g})$ be from an optimal solution of the \texttt{DIMENSION} loss \eqref{eq:split_cyclegan}. Then, there is no admissible MPA of $\{\bP_{\x|u=u_i}\}_{i=1}^I$ of the solution, i.e., 
        $
        \widehat{\f} = \f^\star,~a.e.,\quad \widehat{\g} = \g^\star,~a.e.,
        $
        with a probability of at least $1-\rho^{I\choose 2}$.    
        \end{theorem}
    \end{mdframed}
    \setcounter{theorem}{\value{theorem}-1}
    \renewcommand\theFact{\arabic{theorem}}

Theorem \ref{thm:identifiability} is a direct consequence of following lemma:

\begin{Alemma}\label{lemma:sdc_identifiability}
    Suppose that Assumption \ref{assumption:correspondance} holds.
    % Assume that $\cU = \{u_1, \dots, u_I\}$ satisfy SDC. 
    Assume that  $\{\bP_{\x|u=u_i}\}_{i=1}^I$ are sufficiently diverse.
    Then, $\widehat{\g} = \g^\star$ and $\widehat{\f} = \f^\star$, a.e.
\end{Alemma}

\begin{proof}[Proof of Lemma \ref{lemma:sdc_identifiability}]

First, we show that no non-trivial continuous admissible MPA exists for $\{\bP_{\x|u=u_i}\}_{i=1}^I$, i.e., if a continuous $\h$ satisfies
\begin{align}\label{eq:proof_equal_prob}
\h_{\# \bP_{\x | u = u_i}} ~=~ \bP_{\x|u = u_i} \forall i \in [I],
\end{align}
then $\h = \mathbb{I}$, a.e.

Eq. \eqref{eq:proof_equal_prob}, by the definition of push-forward measure, implies that
\begin{align}\label{eq:measure_equal}
\implies \bP_{\x|u_i}[\h(\cA)] &= \bP_{\x|u_i}[\cA], \forall i \in [I].
\end{align}
For the sake of contradiction assume that $\h$ satisfies \eqref{eq:proof_equal_prob}, however, $\h \neq \mathbb{I}$ on a set of strictly positive measure. This means that there exists a $\overline{\x} \in \cX$ such that
$$\h(\overline{\x}) \neq \overline{\x}.$$
Now, let us define an open set around $\overline{\x}$ denoted by $\cD$ such that
$$ \cD = \cN_d(\overline{\x}) \cap \cX.$$
Because of the continuity and invertibility of $\h$, $\h(\cD) \subseteq \cX$ is also an open set and
$$\h(\overline{\x}) \in \h(\cD).$$
Now, one can select $d$ to be small enough (because of the continuity of $\h$) such that $\cD \cap \h(\cD) = \phi$ and $\cD$ is a connected set. $\cD$ being a connected set implies that $\h(\cD)$ is also connected.

The above is a contradiction to Assumption \ref{assump:distinct_distr} since $\cD$ and $\h(\cD)$ are two disjoint, open and connected sets which satisfy
$$\bP_{\x|u_i}[\h(\cD)] = \bP_{\x|u_i}[\cD], \forall i \in [I].$$

Hence, any $\h$ that satisfy $\h_{\# \bP_{\x | u = u_i}} ~=~ \bP_{\x|u = u_i}$ is such that $\h = \mathbb{I}$, a.e.

Finally, We want to show that $\widehat{\g} = \g^\star$, a.e. 
Lemma \ref{lemma:fact_sln} implies that 
\begin{align}
    \widehat{\g}_{\# \bP_{\x | u = u_i}} ~& =~ \bP_{\y|u = u_i}, \forall i \in [I] \nonumber \\
    \implies  \widehat{\g}_{\# \bP_{\x | u = u_i}} ~& =~ \g^\star _{\# \bP_{\x | u = u_i}}, \forall i \in [I] \nonumber \\
    \stackrel{(a)}{\implies} {\g^\star}^{-1} \circ \widehat{\g}_{\# \bP_{\x | u = u_i}} ~& =~ \bP_{\x|u = u_i}, \forall i \in [I] \label{eq:proof_final_result}
\end{align}
where (a) is obtained by applying ${\g^\star}^{-1}$ on both sides, which is allowed because applying the same function preserves the equivalence of the distributions. 

Eq.~\eqref{eq:proof_final_result} implies that $\g^\star \circ \widehat{\g}$ is a continuous admissible MPA of $\{\bP_{\x|u=u_i}\}_{i=1}^I$, which means that the following has to hold: 
$${\g^\star}^{-1} \circ \widehat{\g} = \mathbb{I}, a.e.$$
Therefore, we always have $ \widehat{\g} = \g^\star$, a.e. By role symmetry of $\f^\star$ and $\g^\star$ (also see Fact \ref{fact:x_y_symmetry}), we also have $\widehat{\f} = \f^\star$, a.e.
\end{proof}

\begin{proof}[Proof of Theorem \ref{thm:identifiability}]
    Using the assumption that ${\rm Pr}[{\sf E}_{i,j}] \leq \rho$, the probability that $\{\bP_{\x|u=u_i}\}_{i=1}^I$ are not sufficiently diverse can be bounded as follows:
    \begin{align*}
        & {\rm Pr}[\{\bP_{\x|u=u_i}\}_{i=1}^I \text{ are not sufficiently diverse} ] \\
        & \stackrel{(a)}{\leq} {\rm Pr}\left[ \bigcap_{ i,j \in [I], i < j} {\sf E}_{i,j}\right] \\
        & \stackrel{(b)}{=} \bigcap_{ i,j \in [I], i < j}{\rm Pr}[ {\sf E}_{i,j}] \\
        & \leq \rho^{I \choose 2},
    \end{align*}
    where the $(a)$ holds since $\{\bP_{\x|u=u_i}\}_{i=1}^I$ not being sufficiently diverse implies the existence of open connected sets $\cA$ and $\cB$ such that
    \begin{align*}
    \bP_{\x|u_i}[\cA] &= \bP_{\x|u_i} [\cB], \forall i \in [I] \\
   \implies \bP_{\x|u_i}[\cA] &= \bP_{\x|u_i} [\cB], \forall i \in \{i,j\} \subset [I].
    \end{align*}
    Finally, $(b)$ is due to the independence of the events ${\sf E}_{i,j}$ and ${\sf E}_{i,j'}$ for $j \neq j'$.

    Hence, $\{\bP_{\x|u=u_i}\}_{i=1}^I$ are sufficiently diverse with probability at least $1- \rho^{I \choose 2}$, which implies that $\widehat{\f} = \f^\star$ and $\widehat{\g} = \g^\star$ with probability at least $1- \rho^{I \choose 2}$.
\end{proof}

\subsection{Proof of Theorem \ref{thm:lipschitz}}\label{app:proof_lipschitz}

\renewcommand\thetheorem{\ref{thm:lipschitz}}
\begin{mdframed}[backgroundcolor=gray!10,topline=false, rightline=false, leftline=false, bottomline=false]
\begin{theorem}[Robust Identifiability]
   Suppose that Assumption \ref{assumption:correspondance} holds with $\bm g^\star$ being $L$-Lipschitz continuous, and that any pair of $(\bP_{\x|u_i},\bP_{\x|u_j})$  satisfies the $r$-SDC (cf. Definition~\ref{assump:violation}) with probability at least $1-\gamma$, i.e., ${\rm Pr}[M_{i,j} \geq r] \leq \gamma$ for any $i \neq j$, where $(i, j) \in [I]\times [J]$.
    Let $\widehat{\g}$ be from any optimal solution of the \texttt{DIMENSION} loss in \eqref{eq:split_cyclegan}. Then, we have
    $\|\widehat{\g}(\x) - \g^\star(\x) \|_2 \leq 2rL, \quad \forall \x \in \cX,$
    with a probability of at least $1 - \gamma^{I \choose 2}$.
    The same holds for $\widehat{\f}$.
\end{theorem}
\end{mdframed}
\setcounter{theorem}{\value{theorem}-1}
\renewcommand\theFact{\arabic{theorem}}

% \begin{theorem}
%     Let ${\rm dia}(\cA) = {\rm sup}_{x, y \in \cA} \|x - y\|_2$ denote the diameter of a set. Let $\cV = \{(\cA_1, \cA_2) | (\cA_1, \cA_2) \text{ violates Assumption \ref{assump:distinct_distr}}\}$. Let 
%     $$D = \max_{(\cA_1, \cA_2) \in \cV} {\rm max}\{{\rm dia}(\cA_1), {\rm dia}(\cA_2)\}. $$ Let $\g$ be $L$-Lipschitz continuous. Any solution to \eqref{eq:auxiliary_cyclegan}, $\widehat{\f}$, satisfies 
%     $$ \|\widehat{\f}(\x) - \x^{(2)} \|_2 \leq 2rL. $$
% \end{theorem}

First, consider the following lemma.
\begin{Alemma}\label{lemma:connected_component}
    Given any continuous admissible MPA $\h$ of $\{\bP_{\x|u=u_i}\}_{i=1}^I$, let $\cE_{\h}$ be a set defined as 
    $$\cE_{\h} = \{ \x  ~|~ \h(\x) \neq \x,~\forall \x \in \cX\}.$$
    Then, any connected component $\cC \subseteq {\rm cl}(\cE_{\h})$ satisfies 
    $$ \x \in \cC \implies \h(\x) \in \cC.$$
\end{Alemma}
{Lemma \ref{lemma:connected_component} states an interesting property of a subset of $\cX$ (namely, $\cE_{\h}$) that is ``modified'' by the continuous MPA $\h$.
Here, ``modification'' means that any point in the subset will land on a different point after the $\h$-transformation.
The lemma shows that the source point from $\cE_{\h}$ and its $\h$-transformation both reside in the same connected component, namely, $\cC$. }
This will be useful in proving Theorem \ref{thm:lipschitz}.
\begin{figure}[h]
    \centering
    \includegraphics[width=0.5\linewidth]{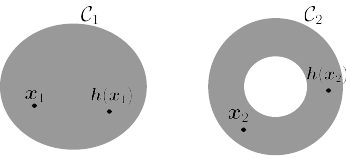}
    \caption{Illustration of Lemma \ref{lemma:connected_component}. $\cC_1, \cC_2$ are the two connected components of  ${\rm cl}(\cE_{\h})$. In this case, ${\rm cl}(\cE_{\h}) = \cC_1 \cup \cC_2$. Points $\x_1$ and $\x_2$ inside $\cC_1$ and $\cC_2$ stay inside the same connected component after transformation by $\h$.}
    \label{fig:connected_component_lemma}
\end{figure}

\textbf{Proof Idea:} The main idea behind the proof of Lemma \ref{lemma:connected_component} is to first note that any point outside of ${\rm cl}(\cE_{\h})$ is stationary under the transformation $\h$ (i.e., $\h(\x) = \x$). Next, if there was a point $\overline{\x}$ from a connected component $\cC_1 \in {\rm conn}({\rm cl}(\cE_h))$ was such that $\h(\overline{\x})$ was not in $\cC_1$, then it should be either in $\cX \backslash {\rm cl}(\cE_{\h})$ or in ${\rm cl}(\cE_{\h}) \backslash \cC_1$. However, since $\h$ is invertible, $\h$ cannot map a point from $\cC_1$ to a $\cX \backslash {\rm cl}(\cE_{\h})$. Therefore $\h(\overline{\x})$ should lie in $\cE_h \backslash \cC_1$. But this will make the function $\h$ discontinuous. Hence, $\h(\overline{\x})$ should be in $\cC_1$.

\begin{proof}[Proof of Lemma \ref{lemma:connected_component}]
    Let ${\rm conn}({\rm cl}(\cE_{\h}))$ denote the set of connected components of ${\rm cl}(\cE_{\h})$. 
    Suppose that there exists $\overline{\x}\in \cC_1$ and $\cC_1 \in {\rm conn}({\rm cl}(\cE_{\h}))$ such that $\h(\overline{\x}) \not \in \cC_1$. First, 
    \begin{align*}
    \h(\widetilde{\x}) = \widetilde{\x},  ~\forall \widetilde{\x} \in \cX \backslash {\rm cl}(\cE_h) & \stackrel{(a)}{\implies} \h(\overline{\x}) \neq \widetilde{\x},  ~\forall \widetilde{\x} \in \cX \backslash {\rm cl}(\cE_h) \\ 
    & \implies \h(\overline{\x}) \in \cC_2 , \text{ for some } \cC_2 \in {\rm conn}({\rm cl}(\cE_{\h})) \backslash \cC_1,
    \end{align*}
    where $(a)$ is due to the invertibility of $\h$.
    
    Because of the continuity of $\h$, the set $\h(\cC_1)$ is a closed connected set containing $\h(\overline{\x})$. However, $\h(\cC_1) \cap (\cX \backslash {\rm cl}(\cE_{\h})) = \phi$. This means that
    \begin{align}\label{eq:proof_subset}
    \h(\cC_1) \subseteq \cC_2,
    \end{align}
    otherwise $\h(\cC_1)$ would be disconnected. 
    
    Note that $\cC_1$ and $\cC_2$ are closed, connected and disjoint sets (by the property of connected components).
    Therefore, one can define $\epsilon$ as follows:
    \begin{align}\label{proof:dist_c1_c2} 
    \epsilon :={\rm dist}(\cC_1, \cC_2) > 0.
    \end{align}
    Now, take any point $\x_b \in {\rm bd}(\cC_1)$, where ${\rm bd}(\cC_1)$ denotes the boundary of $\cC_1$. Due to the continuity of $\h$, there exists a $\delta>0$ such that 
    \begin{align}\label{eq:h_continuity}
        \h(\cN_{\delta}(\x_b)) \subseteq \cN_{\epsilon/4}(\h(\x_b)).
    \end{align}
    However, take any point $\z \in \cN_{\delta}(\x_b) \backslash {\rm cl}(\cE_{\h})$ with $\|\z - \x_b\|_2 < \epsilon/4$. Such a point exists because any neighborhood of a point on the boundary of a closed set has a non-empty intersection with the complement of the closed set. Therefore, we have
    \begin{align}\label{eq:hzz}
    \h(\z) = \z \text{ because } \z \not \in \cE_h.     
    \end{align}

    Since \eqref{eq:proof_subset} implies that $\h(\x_b) \in \cC_2$,  
    $$\|\h(\x_b)-\x_b\|_2 \geq \epsilon \quad \text{ and } \quad  {\rm dist}(\x_b, \cN_{\epsilon/4}(\h(\x_b))) \geq \frac{3\epsilon}{4}.$$
    Therefore 
    \begin{align}
        {\rm dist}(\x_b, \cN_{\epsilon/4}(\h(\x_b))) & \leq \|\x_b - \z\|_2 + {\rm dist}(\z, \cN_{\epsilon/4}(\h(\x_b)))  \nonumber \\
        \implies \frac{3\epsilon}{4} & \leq \frac{\epsilon}{4} + {\rm dist}(\z, \cN_{\epsilon/4}(\h(\x_b))) \nonumber \\
        \implies {\rm dist}(\z, \cN_{\epsilon/4}(\h(\x_b))) & \geq \frac{\epsilon}{2}  \nonumber\\
        \implies {\rm dist}(\h(\z), \cN_{\epsilon/4}(\h(\x_b))) & \geq \frac{\epsilon}{2}, \label{eq:usehzz}
    \end{align}
    where \eqref{eq:usehzz} is by \eqref{eq:hzz}.
    Note that \eqref{eq:usehzz} is a contradiction to \eqref{eq:h_continuity}. Hence, we have
    $$ \x \in \cC \text{ for any } \cC \in {\rm conn}({\rm cl}(\cE_{\h})) \implies \h(\x) \in \cC.$$
    This concludes the proof.
\end{proof}

\begin{Alemma}\label{lemma:theorem2_helper}
    Let $\g^\star$ be $L$-Lipschitz continuous. Suppose that any pair of $(\bP_{\x|u_i},\bP_{\x|u_j})$  satisfies the $r$-SDC (cf. Definition~\ref{assump:violation}).
    Then
    \begin{align}\label{eq:lemma_thm2_bound}
        \|\widehat{\g}(\x) - \g^\star(\x)\|_2 \leq 2rL.
    \end{align}
\end{Alemma}
 
\textbf{Proof Idea:} The proof is by contradiction. Suppose that under the conditions of Lemma \ref{lemma:theorem2_helper}, Eq.~\eqref{eq:lemma_thm2_bound} does not hold for some $\overline{\x} \in \cX$. Then, there would exist a continuous non-trivial admissible MPA $\overline{\h}$ of $\{\bP_{\x|u=u_i}\}_{i=1}^I$ such that $\|\overline{\h}(\overline{\x}) - \overline{\x}\|_2 > 2r$. However, this would imply, using Lemma \ref{lemma:connected_component}, that one can construct an open, connected, disjoint set pair $(\cA, \cB)$ whose diameters are large, which is a contradiction to $r$-SDC that $M\leq r$.

\begin{proof}[Proof of Lemma \ref{lemma:theorem2_helper}]
    % The proof is via contradiction. The main idea of the proof is to use Lemma \ref{lemma:connected_component} to show that if \eqref{eq:lemma_thm2_bound} did not hold, then one can construct a sets with diameter larger than $D$ which results in violation of the sufficiently diverse condition, and therefore should be contained in $\cV$. 
    
    Suppose that there exists $ \overline{\x} \in \cX$ such that 
    \begin{align}\label{eq:proof_contradiction}
    \|\widehat{\g}(\overline{\x}) - \g^\star(\overline{\x})\|_2 > 2rL.
    \end{align}
    Eq.~\eqref{eq:proof_contradiction} means that $\widehat{\g} \neq \g^\star$. 
    {By Lemma \ref{lemma:fact_sln}, we have that 
    \begin{align*}
    \widehat{\g}_{\# \bP_{\x | u = u_i}} & = \bP_{\y | u = u_i} \\   
    \iff \widehat{\g}_{\# \bP_{\x | u = u_i}} & = \g^\star_{\# \bP_{\x | u = u_i}} \\   
    \iff  {\g^\star}^{-1} \circ \widehat{\g}_{\# \bP_{\x | u = u_i}} & = {\g^\star}^{-1} \circ \g^\star_{\# \bP_{\x | u = u_i}} \\   
    \iff  {\g^\star}^{-1} \circ \widehat{\g}_{\# \bP_{\x | u = u_i}} & = \bP_{\x|u=u_i} \\   
    \end{align*}
    As $\widehat{\g} \neq \g^\star$, the function $\overline{\h}:= {\g^\star}^{-1} \circ \widehat{\g} \neq \mathbb{I} $ is a continuous admissible MPA of $\{\bP_{\x|u=u_i}\}_{i=1}^I$. 
    This implies that 
    \begin{align}\label{eq:ghat}
            \widehat{\g} = \g^\star \circ \overline{\h}.
    \end{align}
    }
    % Hence, there exists a non-trivial continuous admissible MPA $\overline{\h}$ such that $\widehat{\g} = \g^\star \circ \overline{\h}$. 
    %{\blue what does this sentence mean? MPA of what? what is $\bar{\h}$ here? is it the same as the one defined in ``Proof Idea''? Hard to follow why this equality holds. Can you start with the definition of $\widehat{g}$ and show the existence?}{\orange Please check now.}
    
    Using \eqref{eq:ghat}, Eq.~\eqref{eq:proof_contradiction} implies that 
    \begin{align}
        \|\g^\star \circ \overline{\h}(\overline{\x}) - \g^\star(\overline{\x}) \|_2 & > 2rL \nonumber \\
        \implies L \|\overline{\h}(\overline{\x}) - \overline{\x} \|_2 & > 2rL  \nonumber \\
        \implies \|\overline{\h}(\overline{\x}) - \overline{\x} \|_2 & > 2r. \label{eq:hx2r}
    \end{align}
    Note that one can re-express \eqref{eq:hx2r} as 
    \begin{align}\label{eq:2depsilon}
        \|\overline{\h}(\overline{\x}) - \overline{\x} \|_2  = 2r + \epsilon,
    \end{align}
    using a certain $\epsilon> 0$.
    By Lemma \ref{lemma:connected_component}, we know that 
    $$\overline{\x}\in\overline{\cC},\quad \overline{\h}(\overline{\x})\in \overline{\cC},$$ where $\overline{\cC}$ is a connected component of ${\rm cl}(\cE_{\overline{\h}})$.
    
    Now, let $d>0$ and
    $$ \cT_d = \cN_{d}(\overline{\x}) \cap \overline{\cC}.$$
    Let $\cR_d$ denote the connected component of $\cT_d$ that contains $\overline{\x}$.  Note that we need to consider the connected component as $\cT_d$ can be a disconnected set.  An illustration of these sets can be seen in Fig. \ref{fig:illus_thm2_proof}.

From \eqref{eq:2depsilon}, it is easy to see that  ${\rm dia}(\overline{\cC}) \geq 2r+\epsilon$. Then, when $d \leq 2r+\epsilon$, since $\overline{\cC}$ is connected, ${\rm bd}(\cN_d(\overline{\x})) \cap \overline{\cC} \neq \phi$. Note that the connected property $\overline{\cC}$ is necessary for  ${\rm bd}(\cN_d(\overline{\x})) \cap \overline{\cC} \neq \phi$ to hold for any $d \leq 2r+\epsilon$. {One can further select $\w \in {\rm bd}(\cN_d(\overline{\x})) \cap \overline{\cC}$ such that $\w \in {\rm cl}(\cR_d)$, i.e., $\w$ lies in the same connected component of $\cT_d$ as $\overline{\x}$.}

{Note that such a $\w$ has to exist. Suppose that such a $\w$ does not exist. Then, ${\rm cl}(\cR_d) \cap {\rm bd}(\cN_d) = \phi$, which means that $\cR_d$ would be disconnected from $\overline{\cC} \backslash \cN_d$.  By the definition of $\cR_d$, $\cR_d$ is then disconnected from $\cT_d \backslash \cR_d$, which implies that $\overline{\cC}$---that is a union of $\cR_d$, $\cT_d \backslash \cR_d$, and $\overline{\cC} \backslash \cN_d$---is disconnected. This is a contradiction. Hence ${\rm cl}(\cR_d) \cap {\rm bd}(\cN_d) \neq \phi$ holds.}

One can see that 
    $$ \|\w - \overline{\x}\|_2 = d, \text{ and } \w, \x \in {\rm cl}(\cR_d) \implies {\rm dia}(\cR_d) \geq d, \text{ for } d \leq 2r + \epsilon.$$
    Hence, there exists a large enough $d\leq 2r+\epsilon$ such that
    $$0<{\rm dist}(\cR_d, \overline{\h}(\cR_d))<\epsilon/3.$$
    % where 
    % $${\rm dist}(\cA, \cB) = \inf_{\x \in \cA, \y \in \cB} \| \x - \y\|_2.$$
    This implies that 
    \begin{align}\label{eq:maxdiaRd}
    {\rm max}\{{\rm dia}(\cR_d), {\rm dia}(\overline{\h}(\cR_d))\} \geq r + \epsilon/3.      
    \end{align}
   Indeed, suppose that ${\rm max}({\rm dia}(\cR_d), {\rm dia}(\overline{\h}(\cR_d))) < r + \epsilon/3$. Then, since $\overline{\x} \in \cR_d$ and $\overline{\h}(\overline{\x}) \in \overline{\h}(\cR_d)$, 
    \begin{align*}
        \|\overline{\x} - \overline{\h}(\overline{\x})\|_2 & \leq 2 {\rm max}\{{\rm dia}(\cR_d), {\rm dia}(\overline{\h}(\cR_d))\} + {\rm dist}(\cR_d, \overline{\h}(\cR_d)) \\
        2r + \epsilon & < 2r + 2\epsilon/3 + \epsilon/3 \\
        2r + \epsilon & < 2r + \epsilon,
    \end{align*}
    which is a contradiction. Hence, 
    \begin{align}\label{eq:proof_thm2_contradiction}
    {\rm max}\{{\rm dia}(\cR_d), {\rm dia}(\overline{\h}(\cR_d))\} \geq r + \epsilon/3.
    \end{align}
    
    \begin{figure}
        \centering
        \includegraphics[width=0.5\linewidth]{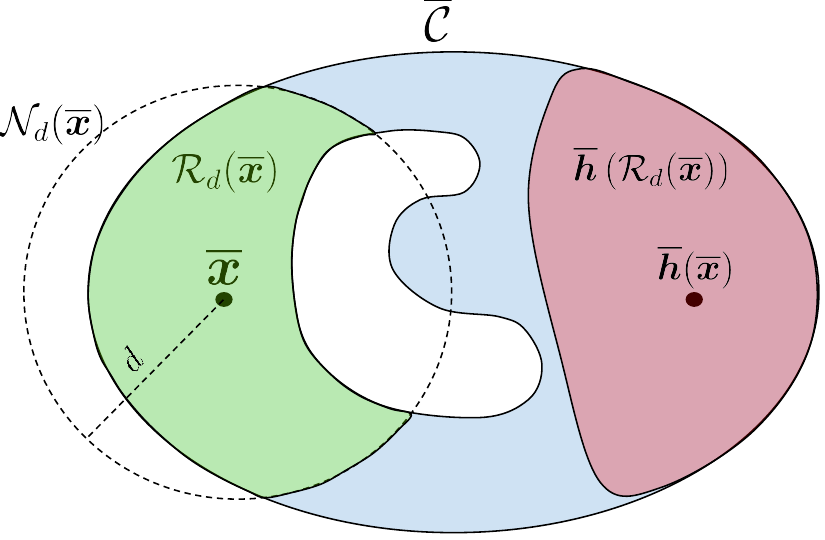}
        \caption{Illustration of the idea in the proof of Lemma \ref{lemma:theorem2_helper}. The green shaded region denote $\cR_d$. Note that ${\cal T}_d = {\cN}_d(\overline{\x}) \cap \overline{\cal C}$ is disconnected in this case. }
        \label{fig:illus_thm2_proof}
    \end{figure}

    Fig.~\ref{fig:illus_thm2_proof} provides a simple illustration of the sets. It follow from the continuity and invertibility of $\overline{\h}$ that
    % $\cR_d(\overline{\x})$ and $\overline{h}(\cR_d(\overline{\x}))$.
    ${\rm dia}(\cR_d) = {\rm dia}({\rm int}(\cR_d))$ and ${\rm dia}(\overline{\h}(\cR_d)) = {\rm dia}({\rm int}(\overline{\h}(\cR_d)))$.
    % (see Fact \ref{fact:fact_interior_equal} after the proof of Theorem \ref{thm:lipschitz}).
    % Next, note that the above holds for arbitrary $\epsilon>0$. 
    
    {By the same argument of reaching \eqref{eq:measure_equal},} $\{{\rm int}(\cR_d), \overline{\h}({\rm int}(\cR_d))\}$ forms a pair of open, connected, disjoint sets such that
    \begin{align}\label{eq:proof_larger_than_d}
        \bP_{\x|u_i}[{\rm int}(\cR_d)] = \bP_{\x|u_i} [\overline{\h}({\rm int}(\cR_d))].
    \end{align}
    Note that \eqref{eq:proof_larger_than_d} and \eqref{eq:proof_thm2_contradiction} constitute a contradiction to the assumption that 
    $$M = \max_{(\cA, \cB) \in \cV} {\rm max}\{{\rm dia}(\cA), {\rm dia}(\cB)\} \leq r$$
    for any open, connected, and disjoint sets $\cA$ and $\cB$\footnote{Note that such requirements are used to ensure that the sets have nonzero measure. The statements can be simplified by replacing the ``open and non-empty'' sets in Definition~\ref{assump:distinct_distr} and Assumption~\ref{assump:violation} with ``measurable sets with positive measures''.} 
    
    %because of \eqref{eq:proof_thm2_contradiction} and \eqref{eq:proof_larger_than_d}.
    Hence, we must have
    $$ \|\widehat{\g}(\x) - \g^\star(\x) \|_2 \leq 2rL, \forall \x \in \cX. $$
    This concludes the proof.
\end{proof}

\begin{proof}[Proof of Theorem \ref{thm:lipschitz}]
    We can bound the probability with which \eqref{eq:lemma_thm2_bound} does not hold as follows:
    \begin{align*}
        & {\rm Pr}[\eqref{eq:lemma_thm2_bound} \text{ does not hold} ] \\
        & = {\rm Pr}[ M > r] \\
        & \stackrel{(a)}{\leq} {\rm Pr}\left[ \bigcap_{ i,j \in [I], i < j} M_{i,j}\right] \\
        & \stackrel{(b)}{=} \bigcap_{ i,j \in [I], i < j}{\rm Pr}[ {M}_{i,j}] \\
        & \leq \gamma^{I \choose 2},
    \end{align*}
where $(a)$ follows because $M > r$ implies that $M_{i,j} > r, \forall i \neq j, \text{ and } i,j \in [I]$ holds, and (b) follows from the independence of the events $M_{i,j} > r$.

Hence, with probability at least $1 - \gamma^{I \choose 2}$,
    $$ \|\widehat{\g}(\x) - \g^\star(\x) \|_2 \leq 2rL, \forall \x \in \cX.$$

The same result follows for $\widehat{\f}$ if $\f^\star$ is $L$-Lipschitz continuous, following the same procedure as above.
\end{proof}

\section{Additional Remark: Relation to Supervised Domain Translation}\label{app:additional_remark}
A remark on objective \eqref{eq:split_cyclegan} is that supervised domain translation can be seen as a special case of \eqref{eq:split_cyclegan}. When paired samples $\{\x_i, \y_i\}_{i=1}^N$ are available, one can view the auxiliary information $u$ as the sample identity. Specifically, $\bP_{\x | u=u_i}$ and $\bP_{\y | u=u_i}$ are Dirac delta distributions peaked at $\x_i$ and $\y_i$, respectively. Matching distributions between $\bP_{\x|u=u_i}$ and $\f_{\# \bP_{\y | \u=u_i}}$ will be equivalent to enforcing $\x_i = \f(\y_i)$. Therefore, the sample loss will be equivalent to minimizing the following objective:
$$ \minimize_{\f, \g} \sum_{i=1}^N \| \x_i - \f(\y_i)\|_2^2 + \| \y_i - \g(\x_i)\|_2^2,$$ 
which is exactly the supervised learning loss. This makes the distribution matching problem boil down to a sample matching problem.

\section{Synthetic Data Experiments}\label{app:synthetic}
In this section, we use controlled generation to validate our identifiability theorems.

{\bf Data Generation.} We generate $\x$ from a Gaussian mixture with $Q$ components. Let $\{\bP_{\x}^{(q)}\}_{q=1}^Q$ denote the $Q$ component distributions of the Gaussian mixture, i.e.,
$$ \bP_{\x}^{(q)} \sim \cN(\bm \mu_q, {\bm \Sigma}),~q=1,\ldots,Q.$$
Here, each ${\bm \mu}_q$ is sampled randomly from the uniform distribution in $\bbR^2$, i.e., ${\rm Unif}\left( [-1,1]^2\right)$. 
We set the covariance to be ${\bm \Sigma} = 0.3^2 \bm Q$. To represent $\g^\star$, we use a three-layer multi-layer perceptron (MLP) with smoothed leaky ReLU, which is defined as $s(x) = \alpha x + (1-\alpha) \log(1+ \exp(x))$, where we set $\alpha$ to $0.2$. To make $\g^\star$ invertible, we generate the neural network weights using the same process as in \citep{hyvarinen1999nonlinear, zimmermann2021contrastive}. Specifically, we use two-hidden units in each layer. We first generate $10,000$ $2\times 2$ matrices, whose elements are sampled randomly from uniform distribution ${\rm Unif}([-1,1])$. 
The matrices' columns are normalized by their respective $\ell_2$ norms. In addition, only the top 25\% well-conditioned matrices in terms of the condition number are used.
%The matrices are normalized with their $\ell_2$-norm, and sorted in an ascending order according to their condition numbers. Then, the matrices ranked smaller than the $2,500$th position are used. 
This way, all the layers of the $\g^\star$ are relatively well-conditioned invertible matrices. Combining with the fact that the activation functions are invertible, such constructed $\g^\star$ in each trial is also invertible.

We use $N = 20,000$ samples in both domains, denoted as $\{\x_n\}_{n=1}^N, \{\y_n\}_{n=1}^N$, to be the training samples. 
In addition, we have 1,000 testing samples.
The data generation process is as follows:
\begin{align*}
    {\bm \mu}_q & \sim {\rm Unif}([-1, 1]^2),~ \forall q \in [Q], \\
    \x_{(q-1)N_q + n} & \sim \bP_{\x}^{(q)}, ~\forall n \in [N_q] ~\forall q \in [Q], \\
    \y_{{(q-1)N_q + n}} & =\g^\star(\x_{(q-1)N_q + n}),
\end{align*}
where $N_q = \lfloor 20000/Q \rfloor$, indicating that the mixture components have equal probability. In our experiments we use $(\x_n,\y_n)$'s association with one of the $Q$ mixture components as our auxiliary variable. Therefore, we have $I=Q$.
In addition, $u$ is uniformly distributed, i.e., ${\rm Pr}(u=u_q) = 1/Q, \forall q \in [Q]$, and 
$$\bP_{\x|u = u_q} = \bP_{\x}^{(q)}.$$

\textbf{Evaluation Metric.}
In the synthetic data, we have access to the ground-truth pairs $(\x_n,\y_n)$. Hence, we measure the translation error (\texttt{TE}) using 
$$
\texttt{TE} = \sum_{n=1}^N \nicefrac{1}{2N}~ (\|\widehat{\g} \left(\x_n\right) - \y_n \|_2^2 + \|\widehat{\f} \left(\y_n\right) - \x_n \|_2^2).
$$

\textbf{Implementation Details.} 
To represent $\g:\mathbb{R}^2 \rightarrow \mathbb{R}^2$ and $\f:\mathbb{R}^2 \rightarrow \mathbb{R}^2$, we use three-layer MLPs, where 256 hidden units are used in each of the 2 hidden layers. We also use leaky ReLU activations with a slope of $0.2$. The discriminator is a five-layer MLP with 128 hidden units in each of the hidden layers. Each layer, except for the last, is followed by layer normalization \citep{ba2016layer} and leaky ReLU activations \citep{maas2013rectifier} with a slope of $0.2$. We use the same architecture for all $I$ discriminators in \texttt{DIMENSION}. In the synthetic-data experiments, we implement the distribution matching module using the least-square GAN loss \citep{mao2017least}.

\textbf{Baseline.} In the sythetic experiments, our purpose is to show the lack of translation identifiability of naive distribution matching. Hence,
we use the CycleGAN loss in \eqref{eq:cyclegan} as a benchmark.

\textbf{Hyperparameter Settings.} 
We use the Adam optimizer with an initial learning rate of $0.0001$ with hyperparameters $\beta_1 = 0.5$ and $\beta_2 = 0.999$ \citep{kingma2015adam}. Note that $\beta_1$ and $\beta_2$ are hyperparameters of Adam that control the exponential decay rates of first and second order moments, respectively. We use a batch size of $1000$ and train the models for $2000$ iterations, where one iteration refers to one step of gradient descent of the translation and discriminator neural networks. We use $\lambda=10$ for \eqref{eq:split_cyclegan}. 

\begin{figure}[t]
    \centering
    \includegraphics[width=0.7\linewidth]{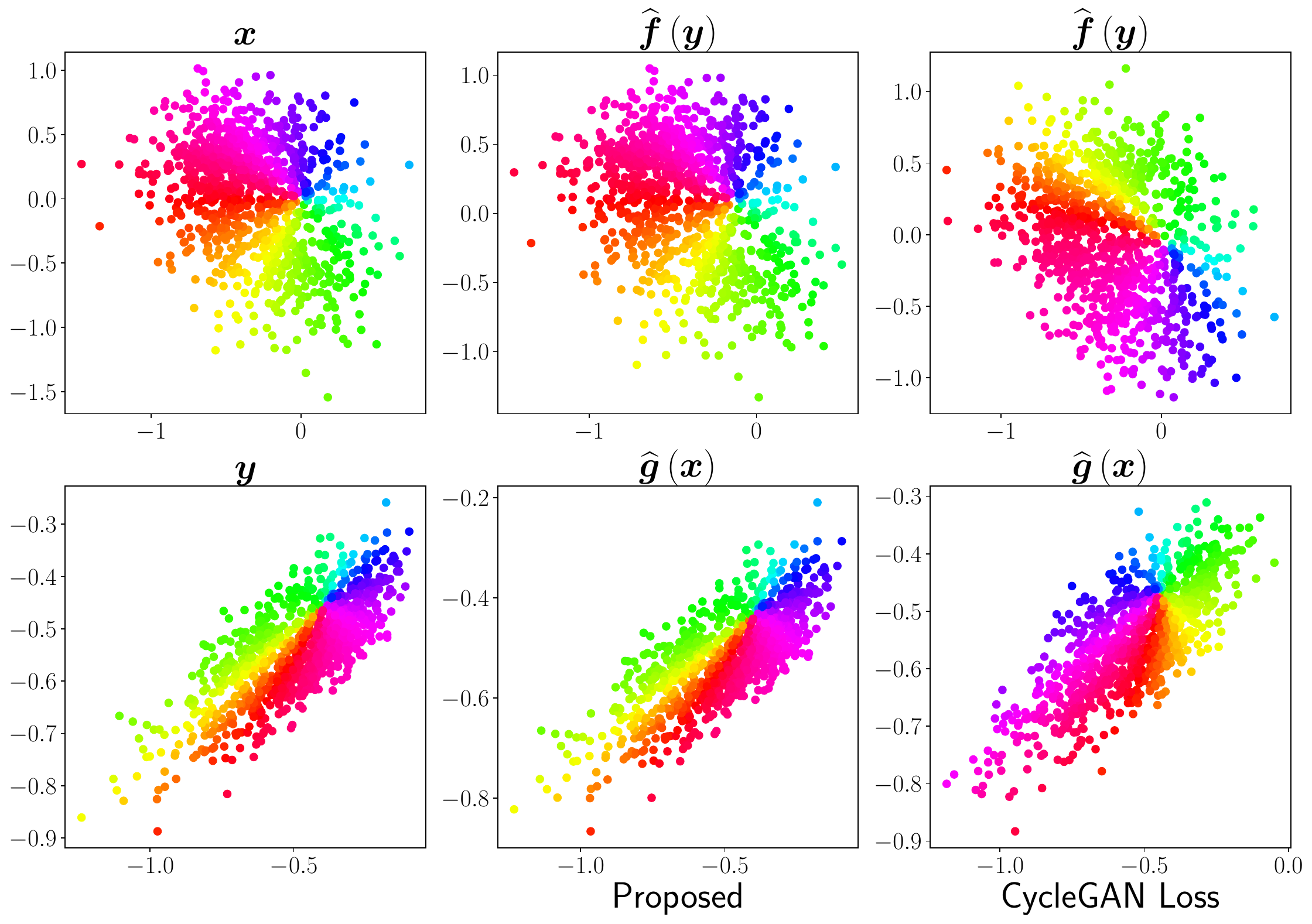}
    \caption{Scatter plots of the source and translated samples. The proposed method uses $I=Q=3$.} 
    \label{fig:complete_synthetic_result}
\end{figure}

\textbf{Results.} Fig. \ref{fig:complete_synthetic_result} shows the scatter plots of the original and translated samples for the 1000 testing samples. Here, we set $I=Q=3$. The original data $\{\x_n\}_{n=1}^N$ and $\{\y_n\}_{n=1}^N$ are plotted on the leftmost column. %Our method uses the information of $u$ and its association of the samples. 
The result of translation using \texttt{DIMENSION} and \texttt{CycleGAN Loss} are presented in the middle and right columns, respectively. In order to qualitatively evaluate the translation performance, we use the same color to plot the paired data points $(\x_n, \y_n)$ and their translations $(\widehat{\f}(\y_n),\widehat{\g}(\x_n))$. The color is determined by the angle of $\x_n$ in polar coordinates. 

As one can see, the supports of $\x$ and $\widehat{\f}(\y)$ (as well as those of $\y$ and $\widehat{\g}(\x)$) are well matched by both methods. 
This implies that both methods can match the distributions fairly well.
However, \texttt{CycleGAN Loss} misaligns the samples (by observing the color). The results given by \texttt{DIMENSION} does not have this misalignment issue.

\begin{figure}[t]
    \centering
    \includegraphics[width=0.4\linewidth]{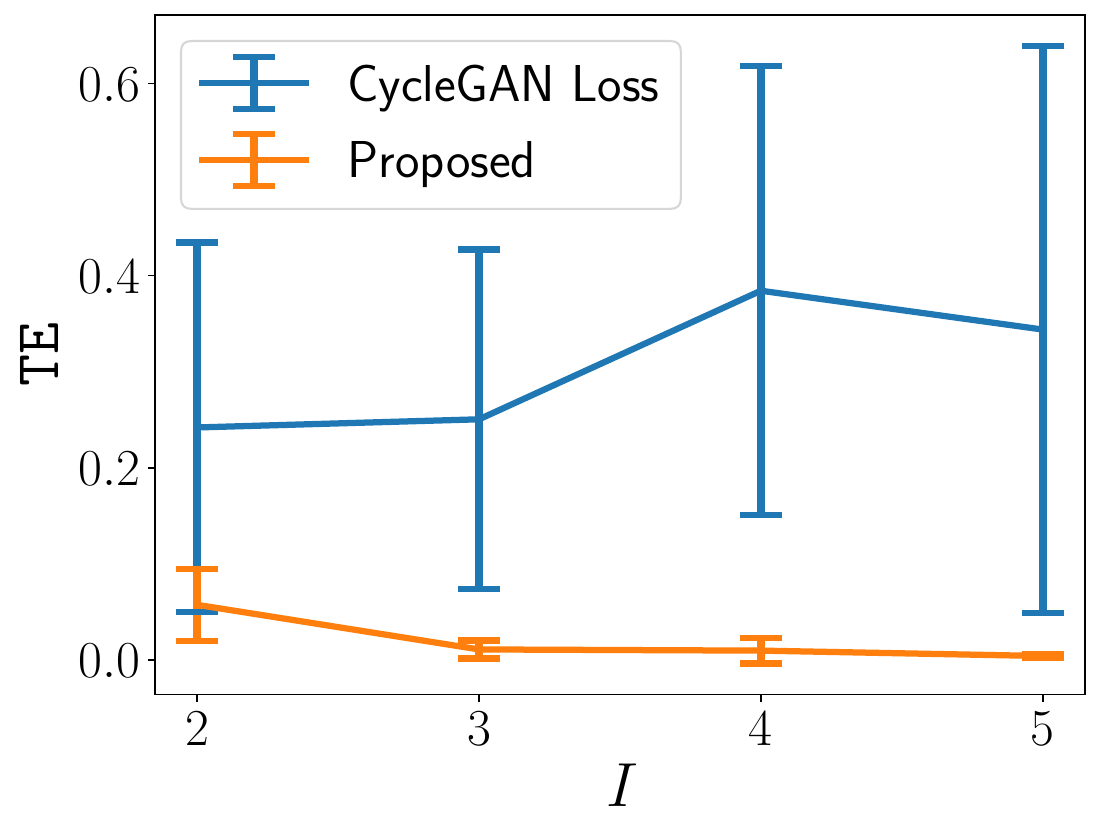}
    \caption{\texttt{TE} under various $I$'s.}
    \label{fig:synthetic_te}
\end{figure}

Fig. \ref{fig:synthetic_te} shows the average \texttt{TE} (over 10 random trials) and the standard deviation attained by \texttt{DIMENSION} and the baseline under different $Q$'s. 
Here, we also set $I=Q$ as before.
One can see that the average \texttt{TE} decreases with the increase in $I$.  
Notably, the {\it variance} of \texttt{TE} also becomes much smaller when $I$ grows from 2 to 5---this shows more stable translation performance when $I$ increases.
The result is consistent with our theorems, which shows that having a larger $I$ has a better chance to avoid MPA.

\section{Real-Data Experiment Setting and Additional Results}
In this section we provide details of the real-data experiment settings.

\subsection{Obtaining $\{u_1, \dots, u_I\}$}\label{app:obtaining_aux}
{\bf MrM and ErS Datasets.} For these two datasets, we use the available category labels as the alphabet of $u$. Specifically, for MrM dataset, we use $u\in\{1,\ldots,10\}$, i.e., the labels of the identity of digits.
For ErS datasets, we use $u \in \{{\rm shoes}, {\rm sandals}, {\rm slippers}, {\rm boots} \}$, which indicates the types of the shoes/edges.

{\bf CB Dataset.} In this dataset, 
we designate the alphabet of $u$ to be $u_1=''\texttt{black hair}''$, $u_2=''\texttt{non-black hair}''$, $u_3=''\texttt{male}''$, $u_4=''\texttt{female}''$.
This information is not fully available in the original CB dataset (to be specific, Bitmoji \citep{bitmojifaces} has the gender attributes available but the hair color is not available).
We use the foundation model, namely, CLIP \citep{radford2021learning}, to acquire the hair color information of each Bitmoji face. 
Specifically, we use the text prompts
``a cartoon of a person whose hair color is mostly black'' and ``a cartoon of a person whose hair color is not black''. The presence of black hair for each image is decided based on cosine distance of the image embedding with the text embeddings of the two prompts.

\subsection{Neural Network Details}\label{app:neural_networks}
We use the nomenclature in Table \ref{tab:nomenclature} to describe the neural network architecture.
For example, 
$$\text{Conv-(C-$N_{\rm in}$ - $N_{\rm out}$, K-$N_k$, S-$N_s$, ZP-$N_p$), LN, LeakyReLU}$$ 
refers to a convolutional layer with $N_{\rm in}$ input channels and $N_{\rm out}$ output channels; K-$N_k$ means that the size of kernel is $N_k$; S-$N_s$ means that the  stride is $N_s$; and ZP-$N_p$ means that the zero padding has a size of $N_p$. The convolutional layer is followed by layer normalization (LN) and then LeakyReLU activations.

\begin{table}[t]
    \centering
    \caption{Nomenclature for neural network components}    \label{tab:nomenclature}
    \resizebox{0.75\linewidth}{!}{
    \def\arraystretch{1.2}%  1 is the default, change whatever you need
    \begin{threeparttable}
    \begin{tabular}{cc}
    \toprule
        Abbreviation & Definition \\ \midrule
         {Conv} & Convolutional Layer  \\ 
         {IN} & Instance normalization \\ 
         {ReLU} & ReLU activation \\ 
         {LeakyReLU} & Leaky-ReLU activation with 0.2 slope \\ 
         Tanh & tanh activation function \\ 
         {UpSample} & Upsample using nearest neighbor with scale factor of $2$ \\ 
         {DownSample} & Downsample using average pooling with a scale factor of 2 \\ 
         K-$N$ & Kernel (filter) of size $N$ \\ 
         S-$N$ & Stride of size $N$ \\ 
         ZP-$N$ & Zero Padding of size $N$ \\ 
         C-$M$-$N$ & $M$ input and $N$ output channels \\ \bottomrule
    \end{tabular}
    \end{threeparttable}
    }
\end{table}

The translation neural networks, $\g$ and $\f$, for images of size $256 \times 256$ follow the architecture outlined in Table \ref{tab:generator}. For images of size $128 \times 128$, a modified architecture is used, where one down-sampling layer (see Layer $\#$6) and one up-sampling layer (Layer $\#11$) in Table \ref{tab:generator} are not included. For images of size $32 \times 32$, three down-sampling layers (indices from $\#4$ to $\#6$) and three up-sampling layers (indices from $\# 11$ to $\# 13$) are not included. 

\begin{table}[t]
    \centering
    \caption{Translation neural network architecture for $\f$ and $\g$.}

    \resizebox{0.55\linewidth}{!}{
    \def\arraystretch{1.2}%  1 is the default, change whatever you need
    \begin{threeparttable}
    \begin{tabular}{cc}
        \toprule
         % Layers Input $\to$ Output Dimension
         Layer Number & Layer Details \\ \midrule
         1  & Conv-(C-3-64, K-1, S-1, ZP-0) \\
         2  & ResBlock-(C-64-128 , DownSample) \\
         3  & ResBlock-(C-128-256, DownSample) \\
         4  & ResBlock-(C-256-512, DownSample) \\
         5  & ResBlock-(C-512-512, DownSample) \\
         6  & ResBlock-(C-512-512, DownSample) \\ \midrule
         7  & ResBlock-(C-512-512, --) \\ 
         8  & ResBlock-(C-512-512, --) \\ 
         9  & ResBlock-(C-512-512, --) \\ 
         10 & ResBlock-(C-512-512, --) \\ \midrule
         11 & ResBlock-(C-512-512, UpSample) \\
         12 & ResBlock-(C-512-512, UpSample) \\
         13 & ResBlock-(C-512-256, UpSample) \\
         14 & ResBlock-(C-256-128, UpSample) \\
         15 & ResBlock-(C-128-64 , UpSample) \\
         16 & Conv-(C-64-3, K-1, S-1, ZP-0) \\ \bottomrule
    \end{tabular}
    \end{threeparttable}
    }
    \label{tab:generator}
\end{table}

ResBlock refers to block of convolutional layers with shortcut connection and optional downsampling. {Specifically, ResBlock-(C-$M$-$N$, \textit{Operation}) is composed of two smaller blocks, namely, 
Process-(C-$M$-$N$, \textit{Operation}) and Shortcut-(C-$M$-$N$, \textit{Operation}).
The Process-(C-$M$-$N$, \textit{Operation}) block has the following layers:
\begin{enumerate}
        \item IN, LeakyReLU, Conv-(C-$M$-$M$, K-3, S-1, ZP-1)
        \item \textit{Operation} 
        \item IN, LeakyReLU, Conv-(C-$M$-$N$, K-3, S-1, ZP-1)
\end{enumerate}
The Shortcut-(C-$M$-$N$, \textit{Operation}) block consists of the following layers:
\begin{enumerate}
    \item Conv-(C-$M$-$N$, K-1,S-1,ZP-0)
    \item \textit{Operation}
\end{enumerate}
Let $\z$ denote the input to the ResBlock and $\w$ the output of the ResBlock. Then the forward pass of ResBlock is expressed as follows:
$$ \w = {\rm ResBlock}(\z) = {\rm Process}(\z) + {\rm Shortcut}(\z).$$
}

\begin{table}[t!]
    \centering
    \caption{Discriminator architecture for $\d_x: \bbR^{256 \times 256 \times 3} \to \bbR^{I}$, $\d_y: \bbR^{256 \times 256 \times 3} \to \bbR^{I}$.}
    \resizebox{0.7\linewidth}{!}{
    \def\arraystretch{1.3}%  1 is the default, change whatever you need
    \begin{threeparttable}
    \begin{tabular}{cc}
    \toprule
     Layer Number & Layer Details \\ \midrule
     1 & Conv-(C-3-64, K-1, S-1, ZP-0) \\
     2 & ResBlock-(C-64-128, DownSample), \\
     3 & ResBlock-(C-128-256, DownSample),  \\
     4 & ResBlock-(C-256-512, DownSample),  \\
     5 & ResBlock-(C-512-512, DownSample),  \\
     6 & ResBlock-(C-512-512, DownSample),  \\ 
     7 & ResBlock-(C-512-512, DownSample), LeakyReLU \\ 
     8 & Conv-(C-512,512, K-4, S=1, ZP=0), LeakyReLU \\
     9 & Reshape-512 \\
     10 & Linear-(512,I) \\ \bottomrule
    \end{tabular}
    \end{threeparttable}
    }
    \label{tab:discriminator}
\end{table}

We use multi-task discriminators \citep{liu2019few} with output dimension of $I$ to represent $\d_x^{(i)}, \d_y^{(i)}, \forall i \in [I]$. Specifically, each of the multi-task discriminators $\d_x$ and $\d_y$ has $I$ output dimensions. The $i$th outputs of $\d_x$ and $\d_y$ correspond to $\d_x^{(i)}$ and $\d_y^{(i)}$, respectively.

\subsection{Hyperparameter Setting}\label{sec:app_hyperparmeters_real}
We use the Adam optimizer with an initial learning rate of $0.0001$ with hyperparameters $\beta_1 = 0.0$ and $\beta_2 = 0.999$ \citep{kingma2015adam}. Note that $\beta_1$ and $\beta_2$ are hyperparameters of Adam that control the exponential decay rates of first and second order moments, respectively. We set our regularization parameter $\lambda = 10$. We use a batch size of $16$. We train the networks for 100,000 iterations. Following standard practice, we add squared $\ell_2$-norm regularization on the network parameters and use a \textit{weight decay} of 0.00001. For the translation tasks with $256 \times 256$ images (CelebA-HQ to Bitmoji Faces), the runtime using a single Tesla V100 GPU is approximately 55 hours. For the translation tasks with $128 \times 128$ images (Edges to Rotated Shoes), the runtime using a single Tesla V100 GPU is approximately 35 hours. 
{In order to stabilize the GAN training dynamics, we add a gradient penalty term. This term penalizes discriminators' large gradients , which is known to help the convergence of the GAN objective \citep{mescheder2018training}. 
We modified the rgeularizer to accommodate our diversified DT loss function.
The modified regularization term is as follows:
$$\cR = \frac{\gamma}{2} {\sf Pr}(u=u_i)\left(\bbE_{\x \sim \bP_{\x|u=u_i}}\|\nabla \d_x^{(i)}\|_2^2 + \bbE_{\y \sim \bP_{\y|u=u_i}}\|\nabla \d_y^{(i)}\|_2^2 \right),$$
where $\nabla \d_x^{(i)}$ denotes the gradient of $\d_x^{(i)}$. We set the value of $\gamma$ to be $1.0$. {We take exponential moving average (EMA) of the parameters during training as the final estimate of the parameters of the trained neural networks. We use a weighting factor of 0.999. This has been observed to improve the performance of GANs \citep{karras2017progressive, yaz2018unusual}.}
}

\subsection{Dataset Details}\label{sec:app_dataset_details}

\textbf{MNIST to Rotated MNIST (MrM).}
We use $60,000$ training samples of the MNIST digits \citep{lecun2010mnist} that have a dimension $28 \times 28$ as the ${\cal X}$-domain. For the ${\cal y}$-domain, each of the $60,000$ digits is rotated by $90$ degrees. The orders of samples are shuffled in both domains to ``break'' the content correspondence. Under this setting, each $\x$ has a ground-truth correspondence $\y$.

\textbf{Edges to Rotated Shoes (ErS).}
Edges2Shoes dataset \citep{isola2017image} consists of $49,825$ training samples. We resize the all images to have $128 \times 128$ pixels. The ${\cal X}$ -domain corresponds to the {\it edges of the shoes}, and the ${\cal Y}$-domain corresponds to the {\it shoes} that are rotated by $90$ degrees. Like in the MrM dataset, the ground-truth correspondence is known to us, which can assist evaluation.

\textbf{CelebA-HQ to Bitmoji Faces (CB)}
We use $29,900$ training samples from CelebA-HQ \citep{karras2017progressive} as the $\x$-domain, and $3,984$ training samples from Bitmoji faces \citep{bitmojifaces} as the $\y$-domain. Note that Bitmoji Faces consists of only $4,084$ samples in total, of which $100$ samples are held out as the test samples. We resize all images in both domains to have $256 \times 256$ pixels.
Unlike the previous two datasets, the ground-truth correspondence is {\it not} known to us in this dataset.

\textbf{Evaluation Details.}
The \texttt{LPIPS} score is computed using 100 test samples. Pre-trained AlexNet\citep{krizhevsky2012imagenet, zhang2018unreasonable} is used in order to compute the \texttt{LPIPS} scores. The \texttt{FID} score is computed using 1000 translated and real samples for each domain. Pre-trained Inception-v3 \citep{szegedy2016rethinking} is used in order to compute the \texttt{FID} scores.

\subsection{Baselines}\label{app:baselines_setting}
We use \texttt{CycleGAN+Id} \citep{zhu2017unpaired}{\footnote{\label{cyclegan}https://github.com/junyanz/pytorch-CycleGAN-and-pix2pix.git}}, 
\texttt{UNIT} \citep{liu2017unit} \footnote{\label{unit}https://github.com/NVlabs/MUNIT.git}, 
\texttt{MUNIT} \citep{huang2018multimodal} \textsuperscript{\ref{unit}}, 
\texttt{U-GAT-IT} \citep{kim2019ugatit} \footnote{https://github.com/znxlwm/UGATIT-pytorch.git}, 
\texttt{StarGAN-v2} \citep{choi2020starganv2} \footnote{https://github.com/clovaai/stargan-v2.git},
\texttt{ZeroDIM} \citep{gabbay2021image} \footnote{https://github.com/avivga/zerodim},
\texttt{OverLORD} \citep{gabbay2021scaling} \footnote{https://github.com/avivga/overlord},
\texttt{Hneg-SRC} \citep{jung2022exploring} \footnote{https://github.com/jcy132/Hneg\_SRC.git},
\texttt{GP-UNIT} \citep{yang2023gp} \footnote{https://github.com/williamyang1991/GP-UNIT.git},
and the plain-vanilla \texttt{CycleGAN Loss} in \eqref{eq:cyclegan} as the baselines.

For \texttt{StarGAN-v2} and \texttt{GP-UNIT}, training is done with their default settings (specifically, the configurations for the `AFHQ' dataset in their papers are used). For \texttt{CycleGAN+Id}, \texttt{UNIT}, \texttt{MUNIT}, \texttt{U-GAT-IT}, and \texttt{Hneg-SRC}, we train the models for 200,000 iterations. We use a batch size of 8 for these methods except for \texttt{U-GAT-IT}, which uses $4$ in order to control the computational load and runtime. These parameters are carefully set for the baselines to our best extent. 
For \texttt{OverLORD} \citep{gabbay2021scaling}, we use the setting used for male to female translation task on CelebA-HQ dataset in their paper. For \texttt{ZeroDIM} \citep{gabbay2021image} , we use the setting used for experiments on FFHQ dataset, which has a similar size as the datasets used in our paper. Note that \texttt{ZeroDIM} also uses the same auxiliary variables as those used in the proposed method.

\subsection{Additional Results}\label{app:additional_results}
In this subsection, we present additional qualitative and quantitative results.

\begin{figure}[t]
    \centering
    \includegraphics[width=\linewidth]{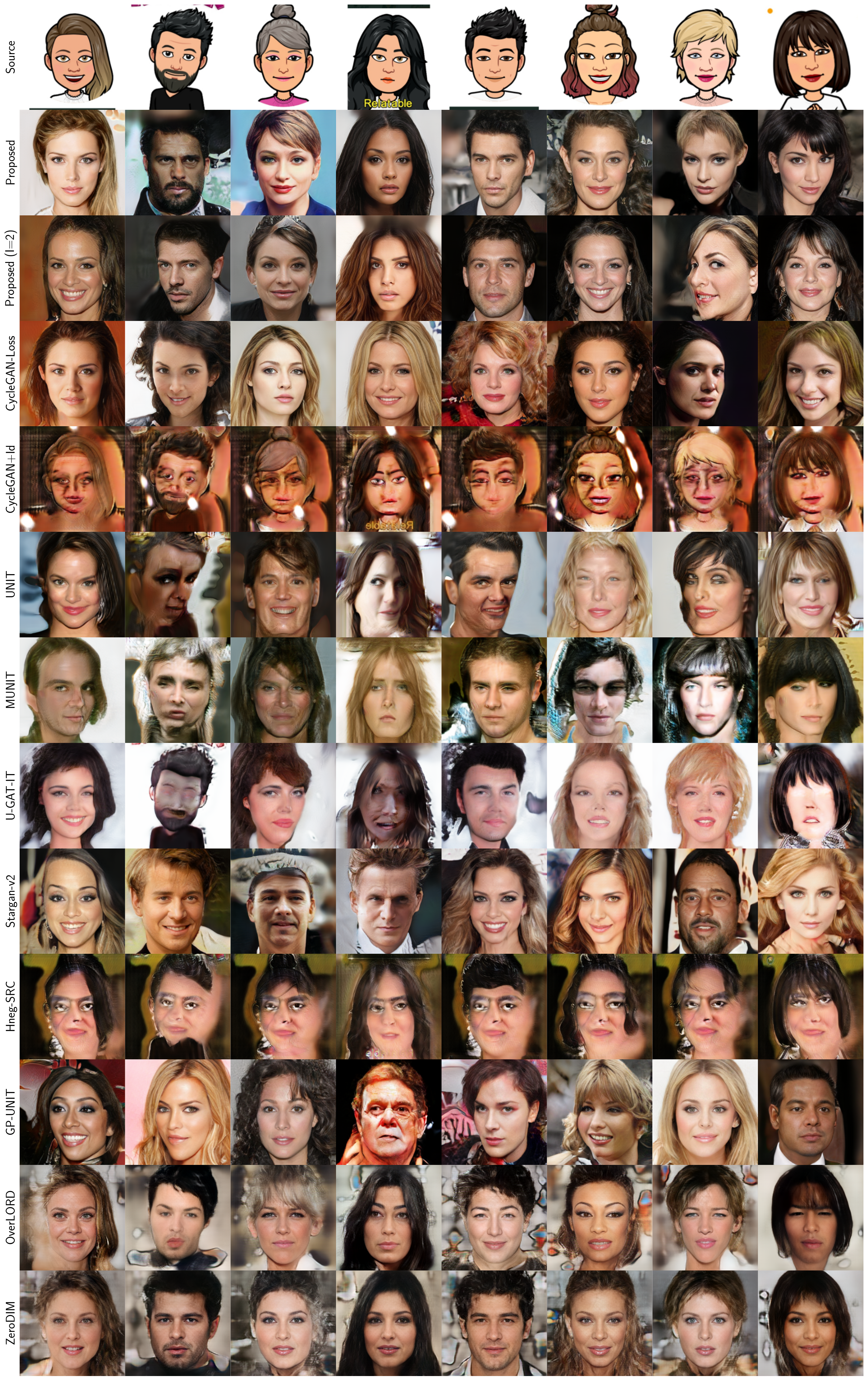}
    \caption{ Translation of Bitmoji to CelebA-HQ.}
    \label{fig:app_bitmoji2celebahq}
\end{figure}

Fig. \ref{fig:app_bitmoji2celebahq} shows the result of translating Bitmoji faces (B) to celebrity proflie photos (C). 
As mentioned in the main text, translating from the B domain to the C domain is a hard task as the learned translation function needs to ``fill in'' a lot of details to make the generated profiles photorealistic.
Visually, one can see that the proposed method (with $I=4$) exhibits much more intuitive content alignment relative to the baselines. In addition, the proposed method using $I=2$ (only using `male' and `female' as the auxiliary variable alphabet) also provides more satisfactory results relative to the baselines. This echos our theoretical claim that the chance of attaining translation identifiability grows quickly when $I$ increases.
It also shows that diversifying the distributions to be matched, even if just one more distribution pair is included, helps improve the final performance.

\begin{figure}[t]
    \centering
    \includegraphics[width=\linewidth]{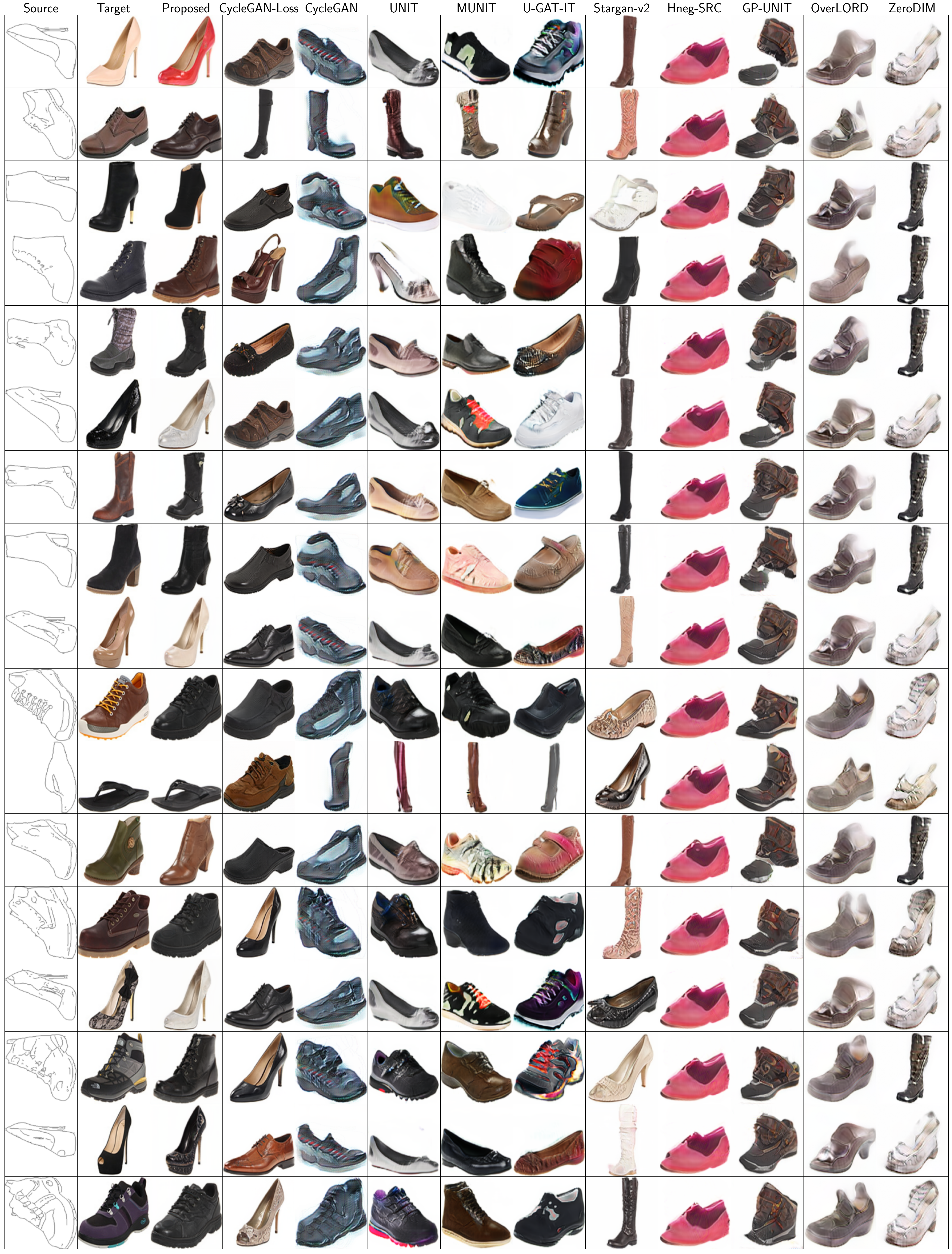}
    \caption{ Translation of edges to rotated shoes. All images rotated by anit-clockwise 90 degrees for visualization. }
    \label{fig:app_edges2shoes}
\end{figure}

Fig. \ref{fig:app_edges2shoes} shows the result of translating edges (E) to rotated shoes (rS). Visually, our method significantly outperforms the baselines in terms of content alignment.
It is interesting to notice that, although ``edges to shoes'' (no rotation) is a well studied dataset, our experiments show that a simple rotation makes most of the existing methods struggle to produce reasonable results.
However, our method is insensitive to this kind of geometric changes. In the literature, the baselines \texttt{U-GAT-IT} \citep{kim2019ugatit} and \texttt{GP-UNIT} \citep{yang2023gp} were shown to be good at handling certain geometric variations. However, one can see that their performance over the ErS dataset is still far from ideal. The result shows the importance of taking transaltion identifiability into account, especially when drastic geometric changes happen across domains.

\begin{figure}[t]
    \centering
    \includegraphics[width=\linewidth]{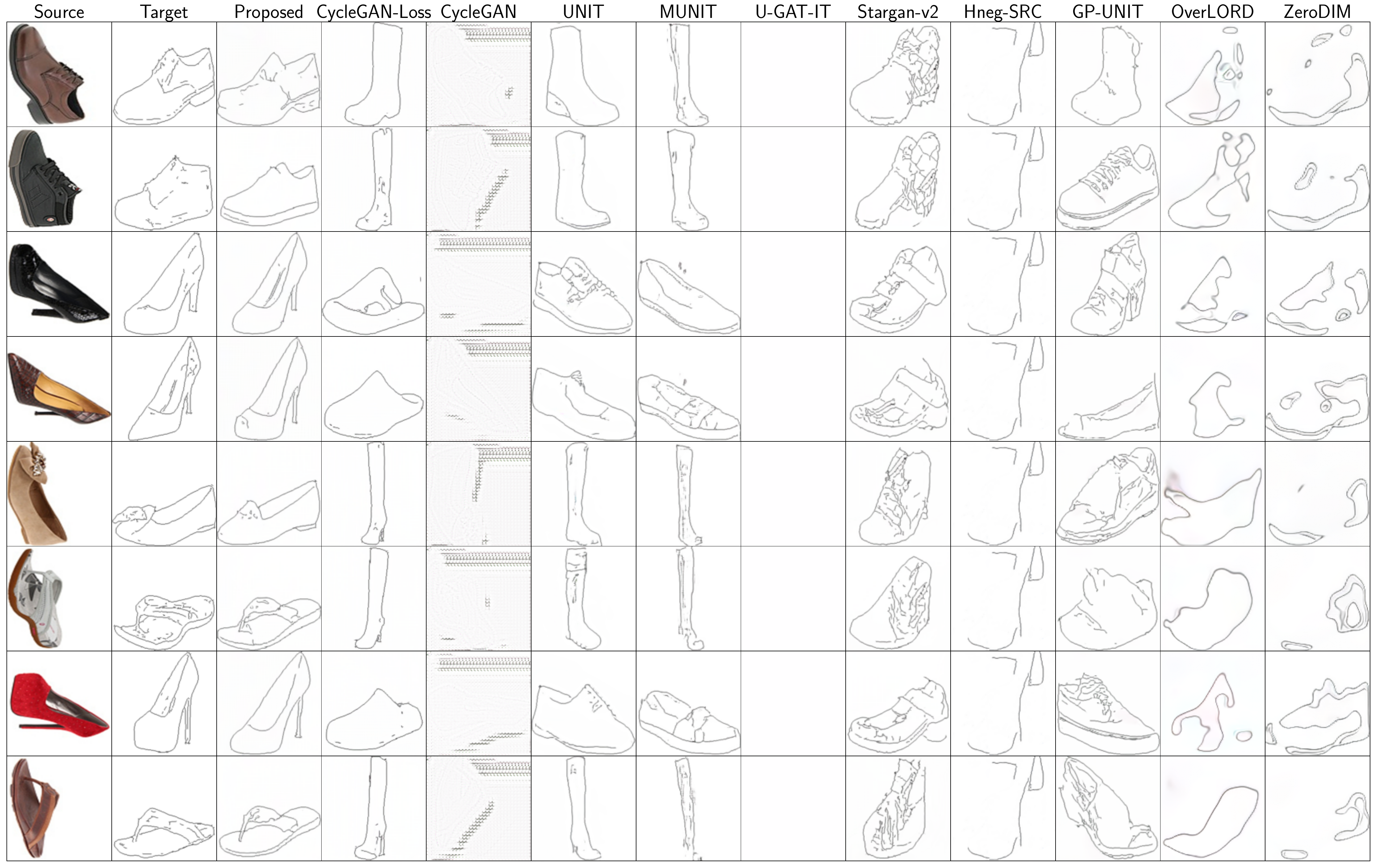}
    \caption{ Translation of rotated Shoes to Edges.}
    \label{fig:app_shoes2edges}
\end{figure}

\begin{figure}[t]
    \centering
    \includegraphics[width=\linewidth]{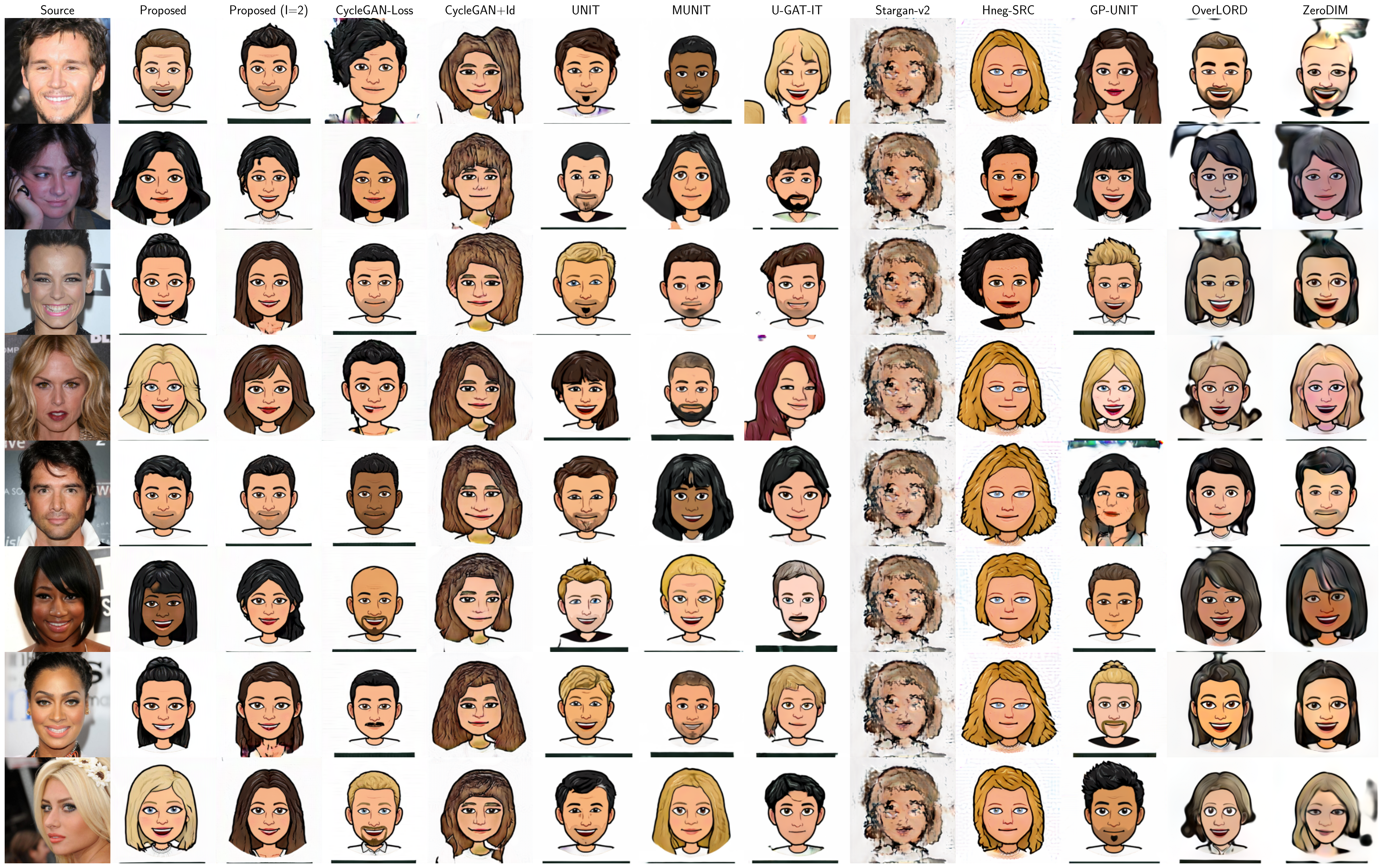}
    \caption{ Translation of CelebA-HQ to Bitmoji.}
    \label{fig:app_celebahq2bitmoji}
\end{figure}

\begin{figure}[t]
    \centering
    \includegraphics[width=0.48\linewidth]{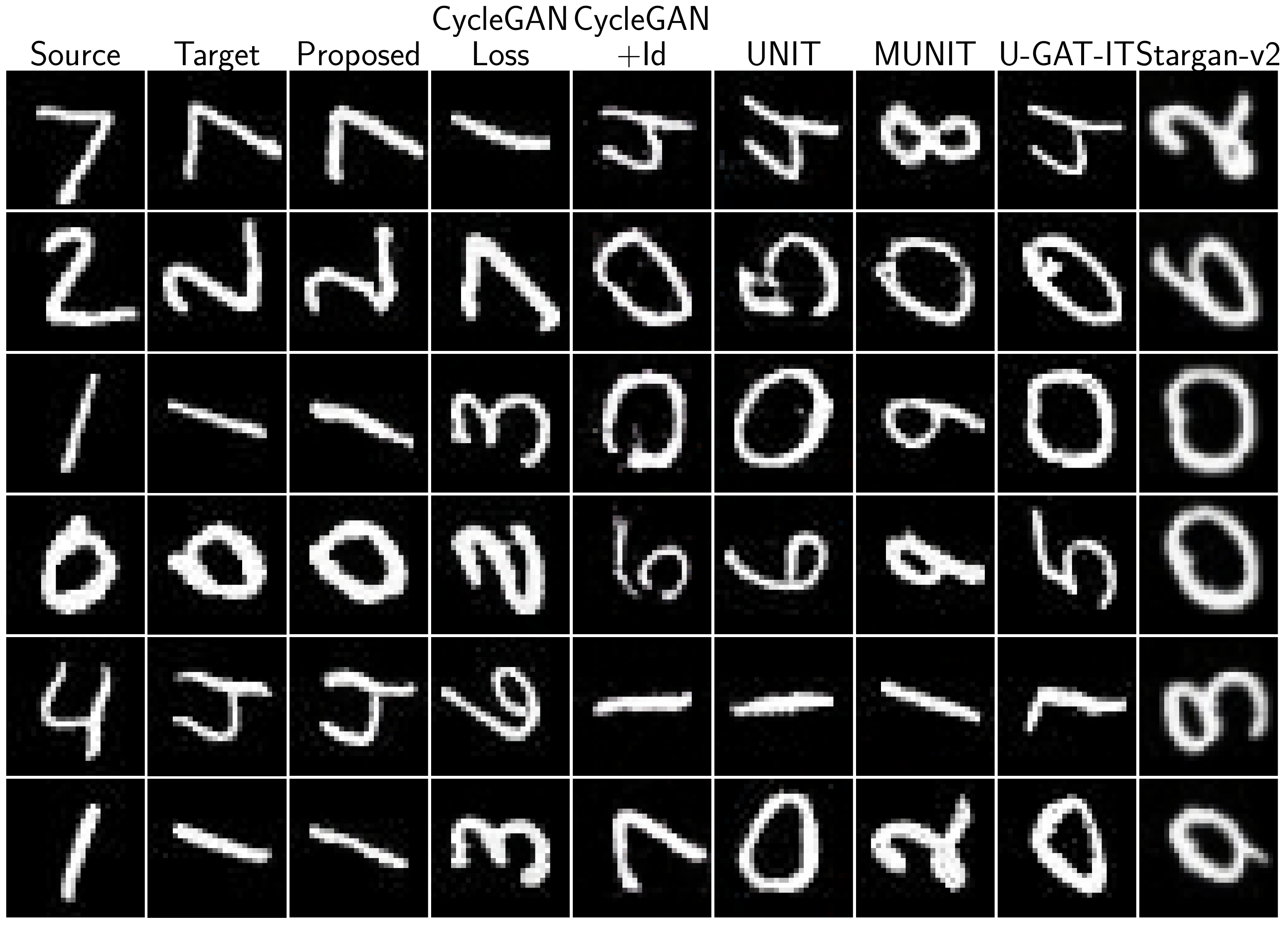}
    \quad
    \includegraphics[width=0.48\linewidth]{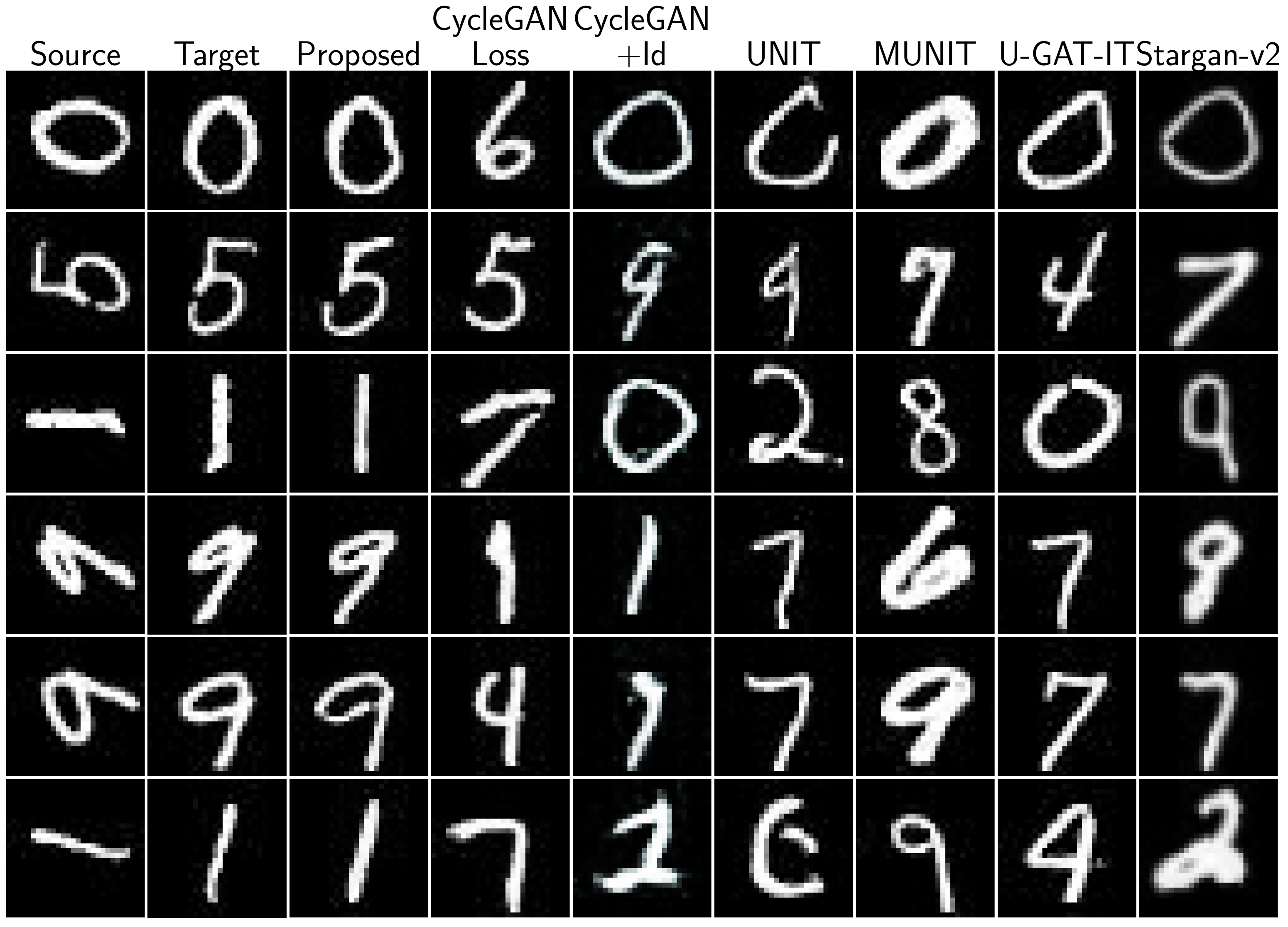}
    \caption{ Translation between MNIST digits and rotated MNIST digits.}
    \label{fig:app_mnist}
\end{figure}

Fig. \ref{fig:app_celebahq2bitmoji} and Fig. \ref{fig:app_shoes2edges} show similar results for the translation of CelebA-HQ (C) to Bitmoji Faces (B) and Rotated Shoes (rS) to Edges (E), respectively. Fig. \ref{fig:app_mnist} shows the translations between MNIST (M) and rotated MNIST digits (rM). 

\clearpage

\begin{figure}
    \centering
    \includegraphics[width=0.48\linewidth]{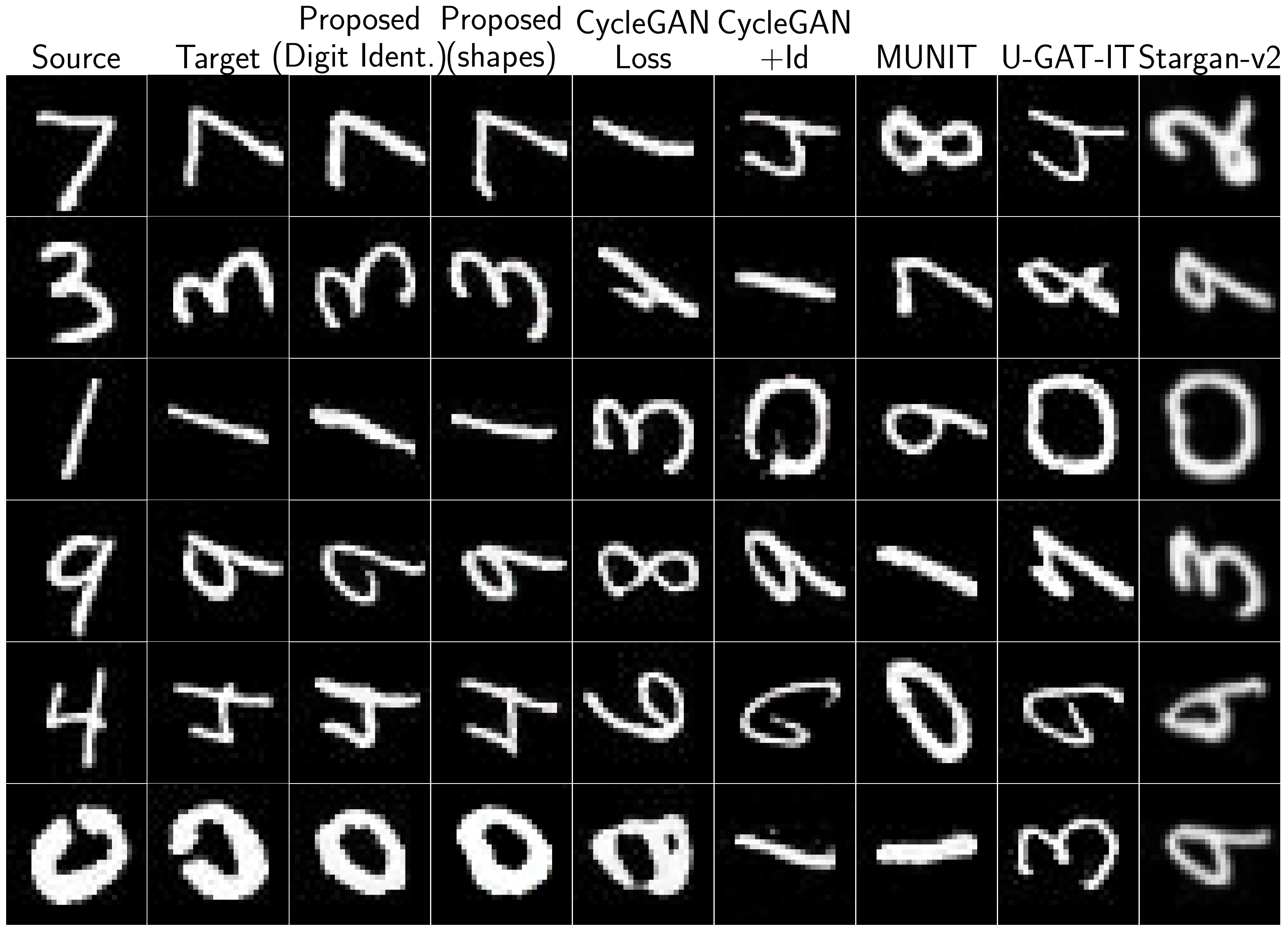}
    \caption{Result of using different auxiliary variable for MNIST digits to rotated MNIST digits task. Using shape attributes incur LPIPS=0.19 $\pm$ 0.08, compared to LPIPS=0.11 $\pm$ 0.08 for the digit identity as auxiliary variable.}
    \label{fig:digit_shape_mnist}
\end{figure}

{There are multiple ways to define auxiliary variables for a given UDT task. However, different choices of auxiliary information can result in different level of translation performance. For example, in the MNIST example, one can alternatively use digit shape as the alphabets of the auxiliary variable. Figure \ref{fig:digit_shape_mnist} shows the result of using different shapes of digits. Here we use the following attributes (with corresponding digits that has those attributes): ``line'' : [1,2,4,5,7],
``circle'' : [0,6,8,9],
``curve'' : [0,2,3,5,6,8,9],
``vertical line'' : [1, 4, 5],
``horizontal line'' : [2,5,4,7],
``curve without loops'' : [2,3,5,6,9],
``only vertical line'' : [1], 
``only horizontal line'' : [2]). One can see that using digit identity as auxiliary information results in a slightly enhanced performance compared to using the digit shape as auxiliary information.}

\section{Improving Existing Methods using Diversified Distribution Matching}\label{app:improving_existing}

We hope to emphasize that the diversified distribution matching (DDM) principle can be combined with many other existing UDT approaches to avoid failure cases.
In this section, we use our diversified distribution matching module to replace the their original ones in existing paradigms and observe the performance.
For the datasets ``Edges vs. Rotated Shoes'' (ErS) and ``CelebA-HQ vs. Bitmoji'' (CB), we select the baselines that are able to generate faithful samples in the target domain based on their \texttt{FID} scores (see Table \ref{tab:lpips}. To be specific, for the ErS dataset, we integrate DDM with \texttt{UNIT} \citep{liu2017unit}. For the CB dataset, we combine DDM with \texttt{GP-UNIT} \citep{yang2023gp}. In both cases, we keep their method-defined regularization terms and other settings unchanged. We refer to the modified methods as \texttt{UNIT-DDM} and \texttt{GP-UNIT-DDM}, respectively.

Our way of combining DDM with these existing approaches is
to replace their discriminators.
To obtain
\texttt{UNIT-DDM} and \texttt{GP-UNIT-DDM}, we modify the discriminator neural networks of \texttt{UNIT} and \texttt{GP-UNIT} into multi-task discriminators. Specifically, for \texttt{UNIT-DDM}, the multi-scale discriminator of \texttt{UNIT} which has one output channel for each scale, is modified to produce $I$ output channels for each scale. Similarly, to obtain \texttt{GP-UNIT-DDM}, the discriminator of \texttt{GP-UNIT} is modified to have $I$ output channels instead of one output channel at the output layer. The $i$th output channel is interpreted as the $i$th discriminator associated with $u_i$. 

\begin{figure}[t]
    \centering
    \includegraphics[width=0.35\linewidth]{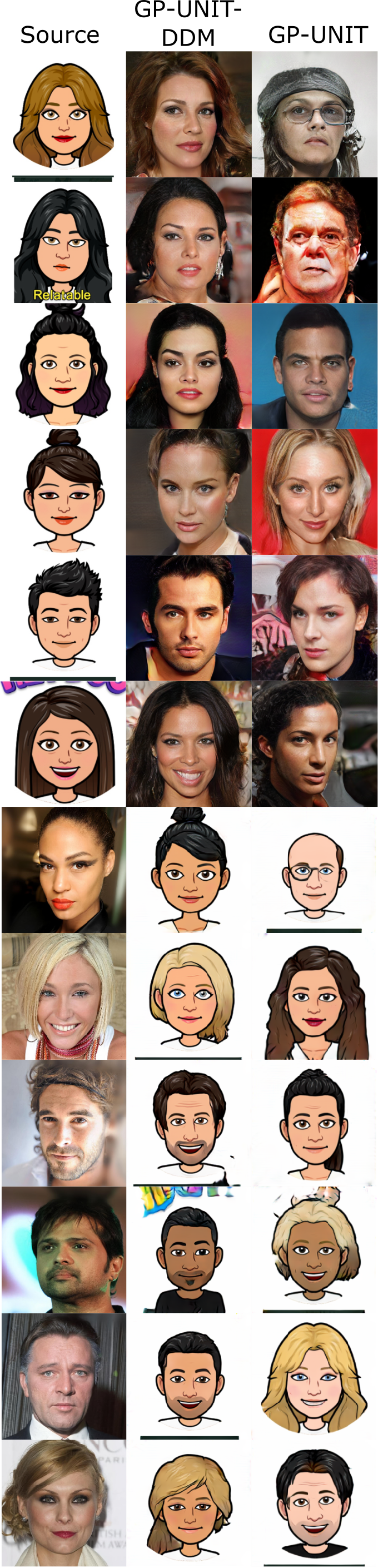}
    \quad 
    \includegraphics[width=0.40\linewidth]{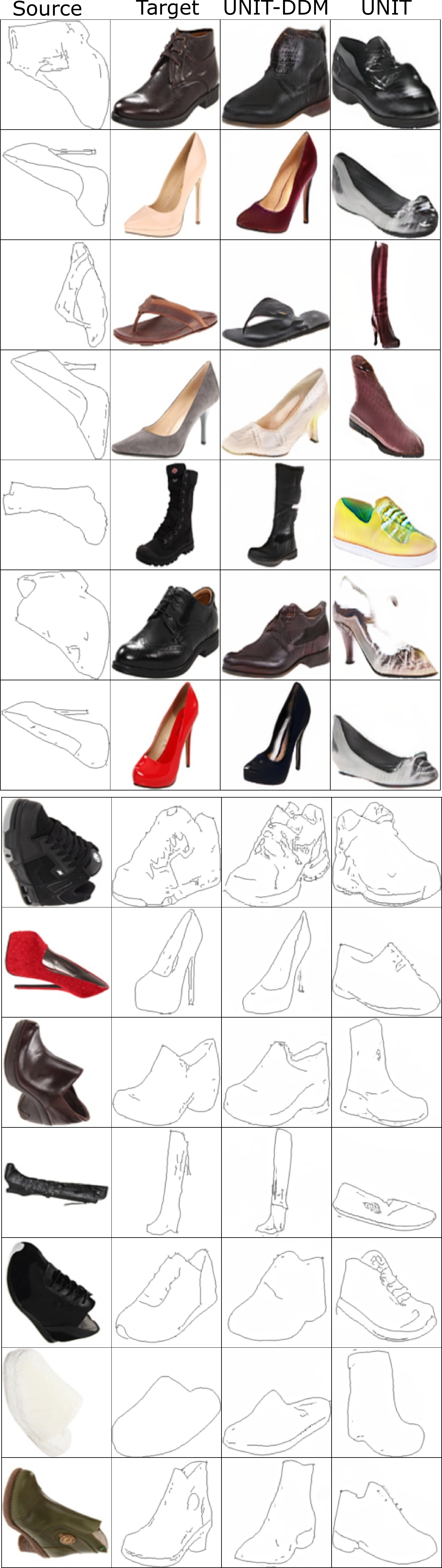}
    \caption{[Left] Result of auxiliary variable-based diverse distribution matching on \texttt{GP-UNIT} on Bitmoji $\rightarrow$ CelebaA-HQ translation task. [Right] Result of Conditional distribution matching on \texttt{UNIT} on Edges $\rightarrow$ Rotated Shoes translation task.}
    \label{fig:existing_method_qual}
\end{figure}

Fig. \ref{fig:existing_method_qual} shows the qualitative results attained by the original versions of \texttt{UNIT} and \texttt{GP-UNIT} as well as their \texttt{DDM}-modified versions. One can see that there is significant improvement in terms of content alignment, without compromising the visual quality---see the \texttt{FID} and \texttt{LPIPS} scores in Table \ref{tab:existing_method_quant}. This attests to the hypothesis that distribution-matching based domain translation frameworks can benefit from the proposed MPA eliminating idea.

\begin{table}[t!]
    \centering
    \caption{The \texttt{FID} and \texttt{LPIPS} scores attained by \texttt{UNIT} and \texttt{GP-UNIT} as well as their \texttt{DDM}-modified versions.}
    \label{tab:existing_method_quant}
    \resizebox{0.8\linewidth}{!}{
    \bgroup
    \def\arraystretch{1.2}%  1 is the default, change whatever you need
    \begin{tabular}{c|cc|cc}
    \toprule
                \multirow{2}{*}{\textbf{Method}} & \multicolumn{2}{c|}{\textbf{\texttt{FID}} ($\downarrow$)} & \multicolumn{2}{c}{\textbf{\texttt{LPIPS}} ($\downarrow$)} \\ \cmidrule{2-5}
                & Edges  & Shoes    & Edges $\to$ Rot. Shoes  & Rot. Shoes $\to$ Edges   \\ \midrule
    \texttt{UNIT}          & 33.95   & 96.28     & $0.49 \pm 0.035$ & $0.58 \pm 0.038$  \\
    \texttt{UNIT-DDM}      & 43.95   & 88.58     & $0.30 \pm 0.075$ & $0.35 \pm 0.092$  \\
\bottomrule
\toprule
    \textbf{Method}         & CelebA-HQ  & Bitmoji  & & \\ \midrule
    \texttt{GP-UNIT}    & 32.40  & 30.30  & &  \\
    \texttt{GP-UNIT-DDM}   & 37.79  & 30.33  & &  \\ \bottomrule
    \end{tabular}
    \egroup
    }
\end{table}
\clearpage

\section{Robustness to Noisy Auxiliary variables.}\label{app:robustness_auxiliary}

It is of interest to know whether using noisy or wrong auxiliary variables would heavily affect the performance of \texttt{DIMENSION}. To this end, we assign random $u_i$'s to a fraction of the training samples in the ``MNIST vs. Rotated MNIST'' dataset.

\begin{figure}[t]
    \centering
    \includegraphics[width=0.48\linewidth]{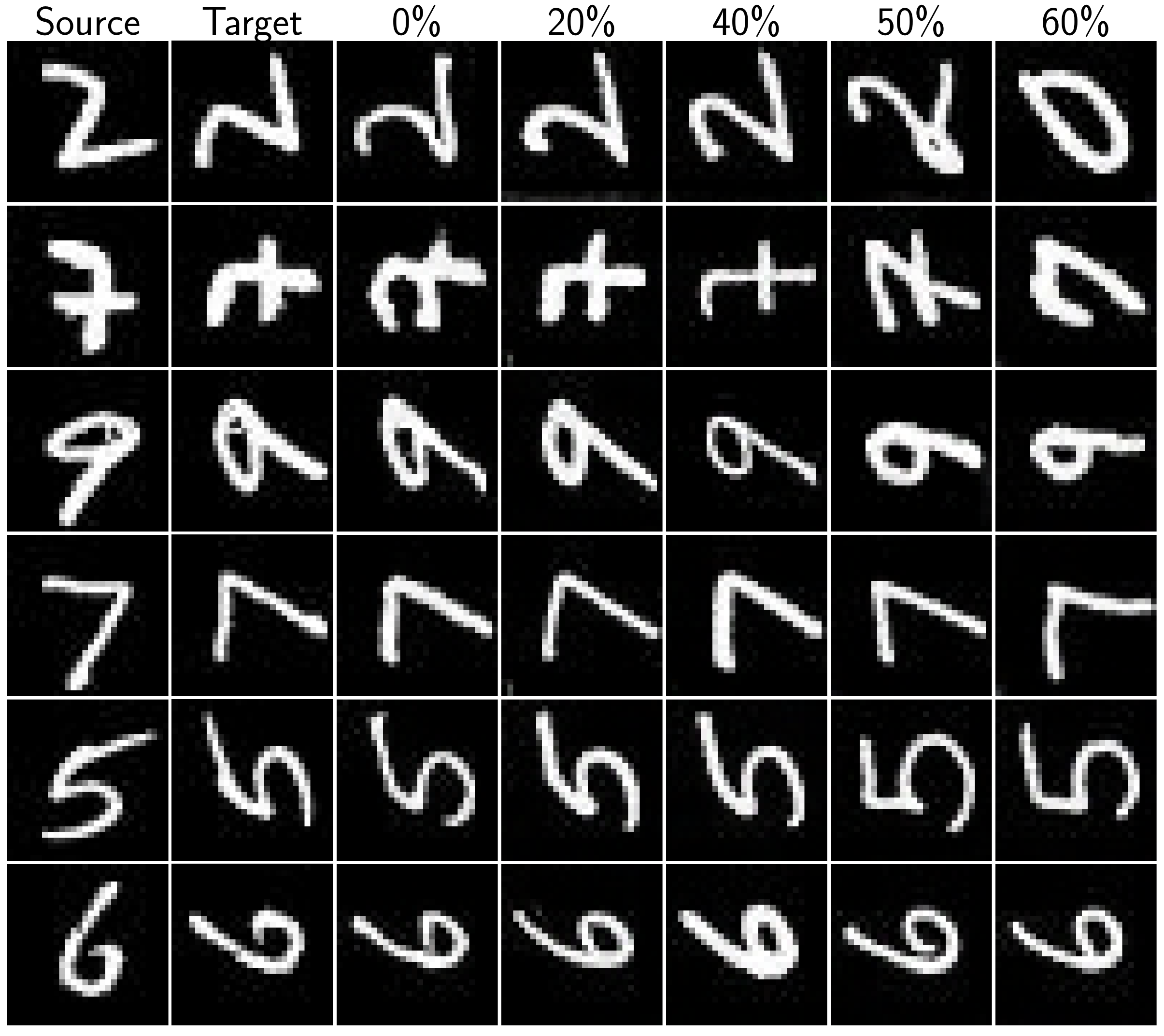}
    \quad
    \includegraphics[width=0.48\linewidth]{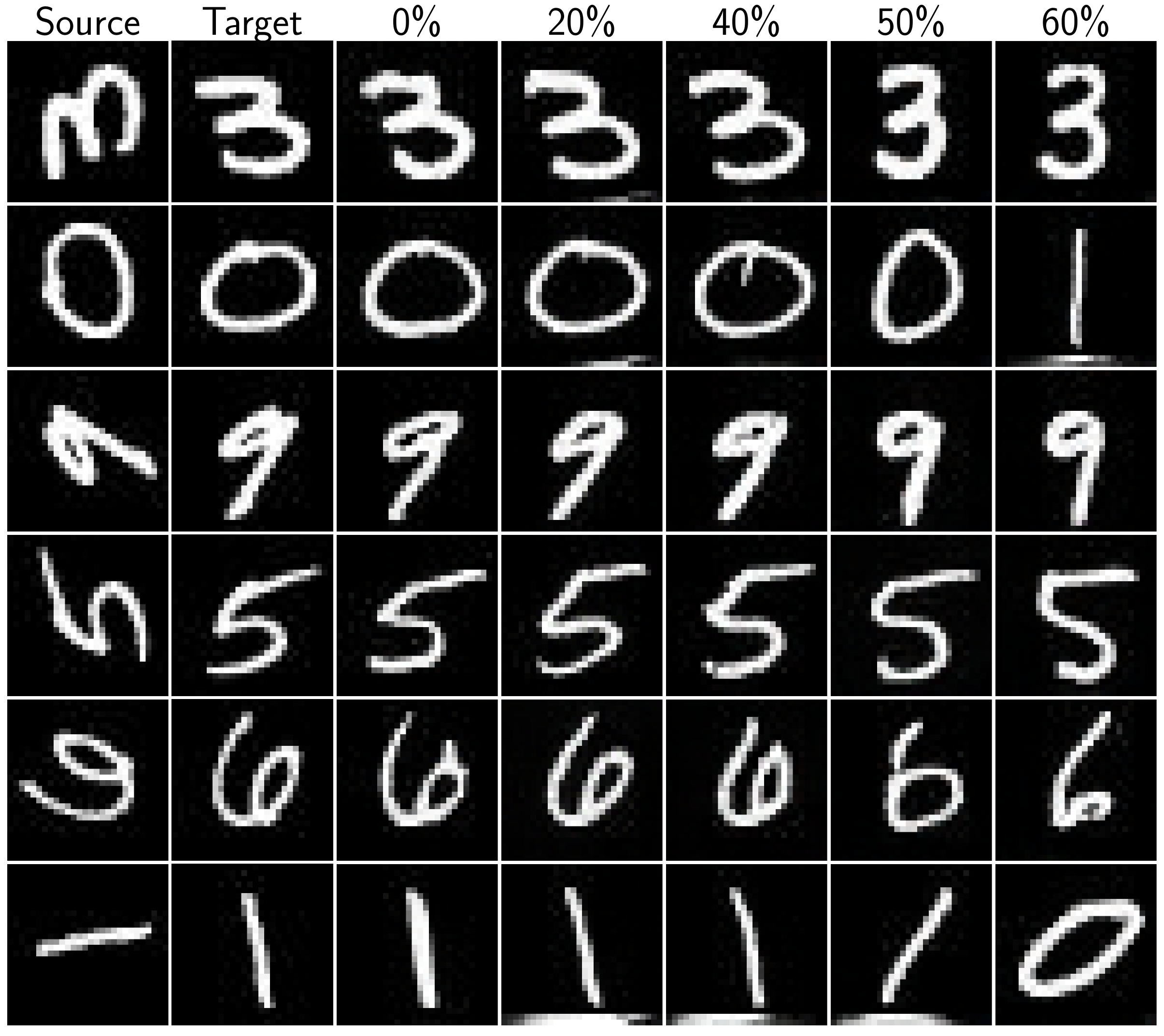}
    \caption{Result of \texttt{DIMENSION} under random $u_i$ assignments to various fractions of training data. }
    \label{fig:app_mnist_mislabel}
\end{figure}

Table \ref{tab:lpips_mnist_mislabel} and Fig. \ref{fig:app_mnist_mislabel} show the \texttt{LPIPS} scores and qualitative results attained by \texttt{DIMENSION}, respectively, under different fractions of random (and highly possibly wrong) auxiliary variables. Notably, there is almost no performance degradation of \texttt{DIMENSION} even when $40\%$ of the assigned $u_i$'s are random. 
This shows the method's robustness to wrong/noisy auxiliary variables.

\begin{table}[t!]
\centering
\caption{\texttt{LPIPS} score attained by \texttt{DIMENSION} using random $u_i$ assignments.}
\label{tab:lpips_mnist_mislabel}
\resizebox{0.7\linewidth}{!}{
\bgroup
\def\arraystretch{1.2}%  1 is the default, change whatever you need
    \begin{tabular}{ccc}
        \toprule
        random $u_i$ proportion & MNIST $\to$ Rot. MNIST & Rot. MNIST $\to$ MNIST\\ \midrule
         0\% & $0.11 \pm 0.082$ &  $0.09 \pm 0.047$ \\
        20\% & $0.09 \pm 0.050$ &  $0.08 \pm 0.040$ \\
        40\% & $0.10 \pm 0.049$ &  $0.13 \pm 0.064$ \\
        50\% & $0.19 \pm 0.080$ &  $0.19 \pm 0.083$ \\
        60\% & $0.25 \pm 0.124$ &  $0.21 \pm 0.086$ \\ \bottomrule
    \end{tabular}
\egroup
}
\end{table}

% \clearpage

% {\red {\bf Move these figures back to close to their text?}}

\clearpage

\clearpage

\end{document}